\documentclass[twoside]{article} 
\usepackage[accepted]{aistats2026}
\usepackage[round]{natbib}

\usepackage{listings}
\usepackage[bookmarksnumbered=true, bookmarksopen=true, bookmarksopenlevel=1,
colorlinks=true,
linkcolor={red!50!black},
citecolor={blue!50!black},
urlcolor={blue!80!black},
pdfstartview=Fit, pdfpagemode=UseOutlines]{hyperref}
\usepackage{url}            
\usepackage{booktabs}       
\usepackage{amsfonts}       
\usepackage{nicefrac}       

\usepackage[textsize=tiny]{todonotes}

\usepackage{subfigure}
\usepackage{wrapfig}
\usepackage{bm}
\usepackage{multicol}
\usepackage{multirow}

\usepackage{amsmath}
\usepackage{amssymb}
\usepackage{mathtools}
\usepackage{amsthm}

\usepackage[capitalize,noabbrev]{cleveref}
\newcommand{\crefp}[1]{(\cref{#1})}
\Crefname{equation}{Eq.}{Eqs.}
\theoremstyle{plain}
\newtheorem{theorem}{Theorem}[section]

\theoremstyle{definition}

\newtheorem{assumption}[theorem]{Assumption}

\makeatletter
\def\th@remark{%
	\thm@headfont{\bfseries}%
	\normalfont 
	\thm@preskip\topsep \divide\thm@preskip\tw@
	\thm@postskip\thm@preskip
}
\makeatother
\theoremstyle{remark}
\newtheorem{remark}[theorem]{Remark}

\usepackage{algorithm}
\usepackage{algpseudocode}
\usepackage{caption}

\usepackage{minitoc}

\begin{document}
\doparttoc 
\faketableofcontents 


\runningtitle{Amortized Safe Active Learning for Real-Time Data Acquisition}
\runningauthor{
Cen-You Li, Marc Toussaint, Barbara Rakitsch$^{\star}$, Christoph Zimmer$^{\star}$
}

\twocolumn[

\aistatstitle{Amortized Safe Active Learning for Real-Time Data Acquisition: Pretrained Neural Policies From Simulated Nonparametric Functions}

\aistatsauthor{
Cen-You Li$^{1,2,\diamond}$
\And
Marc Toussaint$^{1,3}$
\And
Barbara Rakitsch$^{4,\star}$
\And
Christoph Zimmer$^{4,5,\star}$
}

\aistatsaddress{
$^1$Technical University of Berlin, Germany
\quad $^2$University of Helsinki, Finland
\\ $^3$Robotics Institute Germany
\quad $^4$Bosch Center for Artificial Intelligence, Germany
\\ $^5$Baden-W{\"u}rttemberg Cooperative State University, Mannheim, Germany
\\ $^{\diamond}$Correspondence: \texttt{cen-you.li@helsinki.fi}
\quad $^\star$Equal contribution. 
} ]

\begin{abstract}
Safe active learning (AL) is a sequential scheme for learning unknown systems while respecting safety constraints during data acquisition.
Existing methods often rely on Gaussian processes (GPs) to model the task and safety constraints, requiring repeated GP updates and constrained acquisition optimization--incurring significant computations which are challenging for real-time decision-making.
We propose amortized AL for regression and amortized safe AL, replacing expensive online computations with a pretrained neural policy.
Inspired by recent advances in amortized Bayesian experimental design, we leverage GPs as pretraining simulators.
We train our policy prior to the AL deployment on simulated nonparametric functions, using Fourier feature-based GP sampling and a differentiable acquisition objective that is safety-aware in the safe AL setting.
At deployment, our policy selects informative and (if desired) safe queries via a single forward pass, eliminating GP inference and acquisition optimization.
This leads to magnitudes of speed improvements while preserving learning quality. Our framework is modular and, without the safety component, yields fast unconstrained AL for time-sensitive tasks.
\end{abstract}

\section{Introduction}\label{section-introduction}

\begin{figure}[t]
\begin{center}
\centerline{\includegraphics[width=0.9\linewidth]{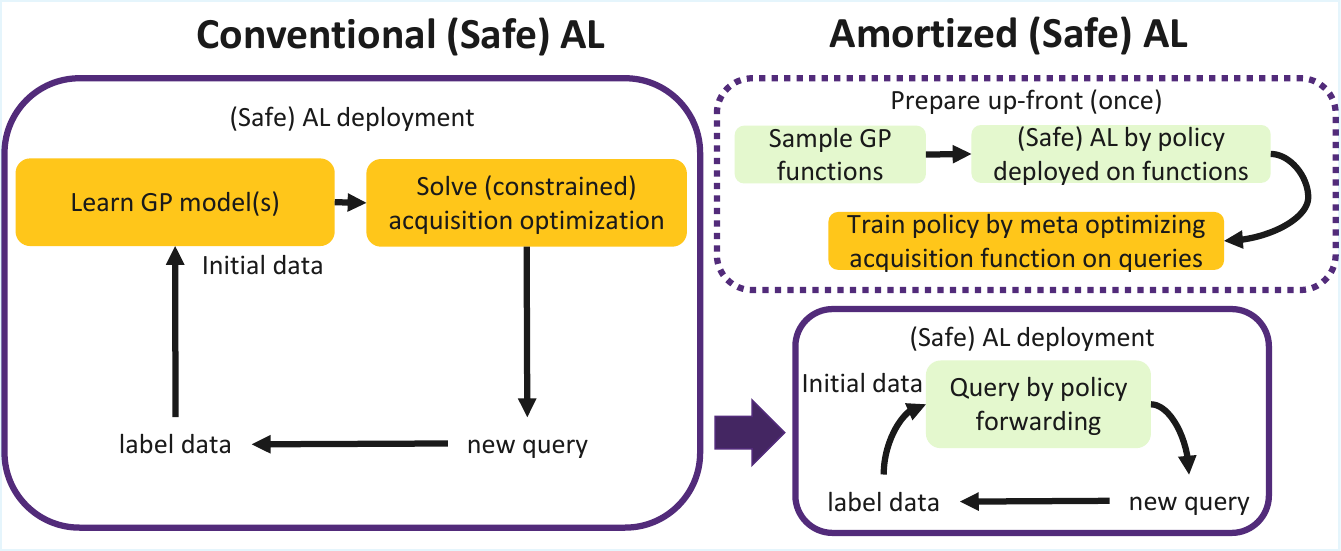}}
\captionof{figure}{Conventional (safe) AL relies on computationally expensive (orange) GP fitting and (constrained) acquisition.
Our amortized approach meta trains a safe learner up-front on synthetic data, allowing fast, real-time (green) deployment.
}\label{figure1}
\end{center}
\vskip -0.3in \end{figure}

Active learning (AL) is a sequential design of experiments, aiming to learn a task with reduced data labeling effort~\citep{settles2010_al,KumarGupta2020,tharwat2023_al}.
Each label is queried by optimizing an acquisition function, 
a function leveraging the current knowledge (typically model-based) to estimate the expected information gained from accessing new labels.
AL is often discussed together with Bayesian optimization (BO), which aims to search global optima with limited evaluations~\citep{Srinivas_2012,brochu2010tutorial}.
The major difference is the acquisition function, where BO focuses only on candidate optima, while AL explores the complete space.

In many engineering~\citep{ZimmerNEURIPS2018_b197ffde,Berkenkamp_2016} or chemical design problems~\citep{griffiths_constrained_2020}, data evaluations can trigger safety concerns.
The safety constraints, often real values, cannot be directly mapped to the input space, motivating the development of safe AL~\citep{Schreiter2015,ZimmerNEURIPS2018_b197ffde,cyli2022} and safe BO~\citep{sui15safeopt,yanan_sui_stagewise_2018,berkenkamp2020bayesian}.
These approaches introduce additional model(s) to quantify safety conditions and constrain the acquisition optimization (\cref{figure1} left).
Gaussian processes (GPs,~\citealt{GPbook}) are widely used in this context due to their well-calibrated uncertainty estimates, suited for modeling safety confidence. 

Safe learning methods are prominent, capable of learning functions adaptively and safely, needing no parametric structure.
However, computation is heavy:
(i) GPs scale cubically with the size of dataset and are updated repeatedly;
(ii) each query solves an acquisition optimization. 
The cost poses a particular challenge for systems requiring real-time responses~\citep{nguyen_tuong10a,andersson2017,Lederer2021_icml_gprealtimecontrol}.
To alleviate this, efficient GP approximations have been explored~\citep{pmlr-v5-titsias09a,pmlr-v38-hensman15,Bitzer2023amortized_gp}, including approaches designed for incremental data~\citep{pmlr-v206-moss23a,pescadorbarrios2024howbigbigenough}.

In this paper, we focus on AL for regressions, particularly under safety constraints~\citep{Schreiter2015,ZimmerNEURIPS2018_b197ffde}, which are also relevant to safe BO, where some recent approaches separate safe space exploration from BO phase~\citep{yanan_sui_stagewise_2018,alessandro2022safe}.
We aim to amortize the online data querying process.
Inspired by recent amortized Bayesian experimental design (BED) literature~\citep{foster_dad_2021,ivanova_idad_2021}, we propose to learn an AL policy offline using synthetic functions.
The policy is a neural network (NN) which suggests a new query via a simple forward pass (\cref{figure1} right) hereby replacing both the GP modeling and the constrained optimization at deployment.
Our paper first amortizes unconstrained AL on regressions, and then introduces safety awareness to the framework.
Note that we use the term \textit{model} to refer to the task-specific model being actively learned, while the NN policy guides the data collections, feeding the data to the model.

In a nutshell, our approach
(i) takes GPs as distributions of general nonparametric functions,
(ii) utilizes a scalable Fourier feature technique~\citep{Rahimi2007_rff,Wilson2020_icml_rff_gp_post} to generate functions in large scale,
(iii) solves and meta learns AL decisions on those functions, and then
(iv) zero-shot generalizes to real-world problems.

\paragraph{Contributions}
Our contributions are:

\begin{itemize}
    \item We propose NN policies for safe AL and, as an intermediate contribution, for unconstrained AL, that suggest new queries based on recorded data, hereby completely replacing the costly GP modeling and acquisition optimization at deployment;
    \item This leads to a tremendous speed gain, allowing for real-time data acquisition with modeling performance comparable to traditional AL methods, while maintaining safety when required;
    \item Our policy is trained up-front exclusively on synthetic data of a broad class of functions sampled via GP Fourier features, enabling generalization across systems;
    \item To train our safe AL policy, we introduce a closed-form, differentiable, safety-aware objective, which may itself be a novel acquisition criterion.
\end{itemize}

\paragraph{Related Works}\label{section-introduction-related_works}


Safe learning often employs GPs.
\citet{Gelbart2014ConstrBO} proposed constrained BO, discounting acquisition scores by a GP constraint confidence. 
\citet{Schreiter2015,sui15safeopt} directly constrained the acquisition optimization, leading to safe AL and safe BO with probabilistic safety guarantees.
\citet{yanan_sui_stagewise_2018} introduced a stagewise safe BO that separates safe space exploration from BO phase, enabling AL methods to be applied specifically for safety~\citep{alessandro2022safe}.
Safe AL and safe BO were extended to systems with multiple safety constraints~\citep{berkenkamp2020bayesian}, time-series modeling~\citep{ZimmerNEURIPS2018_b197ffde}, multi-task learning~\citep{cyli2022} and transfer learning~\citep{cyli2024}.
High-dimensional safe BO has also been explored via approximations and multi-stage procedures~\citep{Kirschner2019lineBO,alessandro2022safe}. 
Safe learning methods employ significant computations, impeding their deployment in real-time problems.

Meta-learning has been explored to streamline sequential learning.
\citet{rothfuss21pacoh} meta-learned GP priors from existing tasks, simplifying GP modeling in safe BO.
\citet{liu2020amortized_gp, Bitzer2023amortized_gp} pretrained on synthetic data to infer the GP parameters.
Neural processes (NPs,~\citealt{garnelo2018conditional,foong2020meta}), in particular transformer-based NPs (TNPs,~\citealt{nguyen2022transformer}), replace GPs by learning the posterior estimates.
\citet{mueller2022pfn} developed the prior-fitted networks (PFNs), TNPs trained on synthetic data, resulting in recent tabular foundation models (e.g. TabPFN,~\citealt{hollmann2023tabpfn,hollmann2025tabpfn}).
NPs and PFNs demonstrate amortized modeling costs and have been extended for unconstrained BO applications~\citep{mueller2023pfns4bo}. 
These methods require existing meta tasks or do not consider (constrained) acquisition optimization, while our method simply queries end-to-end on constrained AL.

To streamline the entire data selection cycle,~\citet{Chen2017_learn_to_learn} proposed an NN optimizer for BO tasks, which bypassed modeling and optimization by directly inferring the next query from evaluated data.
\citet{chang2025ace} proposed predicting the global optimum for BO by taking a prior over the optimum as input to an NN.
\citet{foster_dad_2021,ivanova_idad_2021} proposed the deep adaptive design (DAD), inferring queries for unconstrained AL of Bayesian \textit{parametric} functions (which requires a known parametric structure, in contrast to our method).
\citet{huang2024amortized} extended such amortized BED to account for model-based decision-making.
DAD trains its NN policies on synthetic data, but the necessity of parametric structures limits training flexibility. 

Our approach uses GPs as generic simulators to pretrain end-to-end policies for general, nonparametric (safe) AL, enabling real-time deployment on novel tasks.
The closest related work is ALINE~\citep{huang2025aline}, a recently proposed framework that jointly amortizes posterior estimation and experimental design.
However, ALINE cannot accommodate constrained learning and relies on a discretized search pool.
In contrast, our method addresses standard and safe AL, and is developed so that the NN proposes queries directly on a continuous space.

\section{Problem Statement}\label{section-problem_statement}

We are interested in a regression task of an unknown function $f: \mathcal{X} \rightarrow \mathbb{R}$, where $\mathcal{X} \subseteq \mathbb{R}^D$ is a $D$-dimensional input space.
We have another unknown safety function $q: \mathcal{X} \rightarrow \mathbb{R}$.
As one normally focuses only on a domain of interest, we assume $\mathcal{X}$ is bounded, w.l.o.g., we may say $\mathcal{X}=[0,1]^D$.

Our observations are noisy:
a labeled data point comprises an input $\bm{x} \in \mathcal{X}$, its corresponding output observation $y(\bm{x})=f(\bm{x}) + \epsilon$, and its safety measurement $z(\bm{x})=q(\bm{x}) + \epsilon_{q}$, where $\epsilon, \epsilon_{q}$ are unknown noise values.
For clarity later, let $\mathcal{Y} \subseteq \mathbb{R}, \mathcal{Z} \subseteq \mathbb{R}$ denote the output space and the safety measurement space, respectively.
$\mathcal{D} \subseteq \mathcal{X}\times\mathcal{Y}\times\mathcal{Z}$ is a dataset, and $space(\mathcal{X} \times \mathcal{Y} \times \mathcal{Z}) \coloneqq \{ \mathcal{D} | \mathcal{D} \subseteq \mathcal{X} \times \mathcal{Y} \times \mathcal{Z} \}$.
We write $y_{\text{subscript}}$, $z_{\text{subscript}}$ as evaluated data at $\bm{x}_{\text{subscript}}$.

We follow a safe AL setting:
a small labeled dataset $\mathcal{D}_0 \coloneqq \{ \bm{x}_{\text{init}, i}, y_{\text{init}, i}, z_{\text{init}, i} \}_{i=1}^{N_{\text{init}}}$ is given, and we have budget to label $T$ more data points $\bm{x}_1,...,\bm{x}_T$.
The expensive evaluations will give us $y_1, z_1,..., y_T, z_T$.
It is safety critical if for any $t\in \{1,...,T\}$, $z_t \geq 0$ is violated.
Safe AL aims to select $\bm{x}_1, ..., \bm{x}_T$, such that $z_1 \geq 0, ..., z_T \geq 0$ with high probability, and that $y_1, ..., y_T$ are informative, i.e. 
$
\{ \bm{x}_{\text{init}, i}, y_{\text{init}, i} \}_{i=1}^{N_{\text{init}}} \cup
\{ \bm{x}_t, y_t \}_{t=1}^T$ helps us construct a good model of $f$.
In this paper, we write
$\bm{x}_{1:T} \coloneqq \{\bm{x}_t\}_{t=1}^T$,
$y_{1:T} \coloneqq \{y_t\}_{t=1}^T$,
$z_{1:T} \coloneqq \{z_t\}_{t=1}^T$,
$\bm{X}_{\text{init}} \coloneqq \{\bm{x}_{\text{init},i} \}_{i=1}^{N_{\text{init}}}$,
$Y_{\text{init}} \coloneqq \{y_{\text{init},i} \}_{i=1}^{N_{\text{init}}}$,
$Z_{\text{init}} \coloneqq \{z_{\text{init},i} \}_{i=1}^{N_{\text{init}}}$.

Conventional safe AL (\cref{figure1} left and~\cite{Schreiter2015,ZimmerNEURIPS2018_b197ffde}) obtains each query at step $t+1$ by solving
\begin{align}\label{eq-safe_al}
\begin{split}
\text{argmax}_{\bm{x}\in\mathcal{X}}
    & \ a(\bm{x} | Y_{\text{init}}, y_{1:t})
    \\ & s.t. \ p(z(\bm{x}) \geq 0 | Z_{\text{init}}, z_{1:t}) \geq 1 - \gamma,
\end{split}
\end{align}
where $p(z(\bm{x}) \geq 0 | Z_{\text{init}}, z_{1:t})$ is the predictive safety distribution, $a$ is an acquisition function, $\gamma \in [0, 1]$ is a probability tolerance of being unsafe, and the space of safe $\bm{x}$ is the safe set.
The estimated safe set adapts to each new safety measurement~\citep{sui15safeopt,yanan_sui_stagewise_2018}.
The acquisition function and safety distribution are computed from GPs.
It is expensive to iteratively update GPs and solve the constrained optimization.

\paragraph{Goal}
We aim to have an AL policy up-front so that, at deployment, each query is produced by a single forward pass--replacing the per-query modeling and constrained acquisition optimization (\cref{figure1} right;~\cref{appendix-experiment_details-deployment_algs_ours}).
Formally, the policy takes as inputs a budget variable and a flexible size of observed data, and returns the next query proposal, i.e. $\phi: \mathbb{N} \times space(\mathcal{X} \times \mathcal{Y} \times \mathcal{Z}) \rightarrow \mathcal{X}$.
We will describe in~\cref{section-method-training_objective-unconstrained_random_sequence} why a budget variable is included into the policy input.
We assume no additional real data are available for the policy training.
Our contributions are twofold: (i) We amortize unconstrained AL ($\phi: \mathbb{N} \times space(\mathcal{X} \times \mathcal{Y}) \rightarrow \mathcal{X}$); (ii) we extend to safe AL by incorporating safety-awareness.
Concretely, our paper investigates
(i) how to simulate AL tasks at large scale for training, and
(ii) how to design an effective training objective.
Next, we state the necessary modeling assumptions on the unknown functions $f$ and $q$.

\paragraph{Assumptions}

We assume the tasks are normalized to zero mean and unit variance.
Furthermore, as existing safe learning methods (e.g.~\citealt{ZimmerNEURIPS2018_b197ffde,berkenkamp2020bayesian}), we assume the unknown functions $f$ and $q$ have GP priors~\citep{GPbook}.
A GP is a distribution over functions, characterized by a mean (e.g., $\mathbb{E}[f]$) 
and a kernel specifying covariance between function values at two inputs, $\bm{x}$ and $\bm{x}'$ (e.g., $\text{Cov}[f(\bm{x}),f(\bm{x}')]$).
A kernel, typically parameterized, encodes the function amplitude and smoothness.
W.l.o.g., the prior mean is usually assumed zero, which holds true when the observation values are normalized.
For the safety function, one could as well assume a zero mean prior, but a well-selected prior mean can sometimes be beneficial.
For example, if the domain of interest $\mathcal{X}$ is chosen such that the center area is safe, then we may assign a prior mean which remains reasonably positive at the center but decreases to negative values at the boundary.
Given GP priors, any finite number of output values (or of the safety values) are jointly Gaussian.
The assumption is formulated below, while closed-form GP distributions are detailed in~\cref{appendix-gp_details}.
\begin{assumption}\label{assump-gp_prior}
	The unknown functions follow GP priors: $f \sim \mathcal{GP}(\bm{0}, k_\theta), q \sim \mathcal{GP}_{\theta_{q}}(\mu_{q}, k_{q})$, with kernels
    $k_\theta, k_{q}: \mathcal{X} \times \mathcal{X} \rightarrow \mathbb{R}$ and a mean $\mu_{q}:\mathcal{X} \rightarrow \mathbb{R}$.
    $k_\theta$ is parameterized by $\theta$; $\mu_{q}$ and $k_{q}$ are jointly parameterized by $\theta_{q}$.
	The output and safety observations at $\bm{x}$ are $y(\bm{x}) = f(\bm{x}) + \epsilon$, $z(\bm{x}) = q(\bm{x}) + \epsilon_{q}$, where $\epsilon \underset{i.i.d.}{\sim} \mathcal{N}(0, \sigma^2), \epsilon_{q} \underset{i.i.d.}{\sim} \mathcal{N}(0, \sigma_{q}^2)$. 
	We assume $k_\theta(\bm{x}, \bm{x}'), k_{q}(\bm{x}, \bm{x}') \leq 1$, as the data are normalized.
\end{assumption}

\section{Pretrain Policy to Replace GP and Constrained Acquisition}\label{section-method}

Our goal here is to train a policy $\phi$ to deploy AL on novel tasks.
In other words, we construct our preparation block illustrated in~\cref{figure1}.
Here we take inspiration from~\citet{Chen2017_learn_to_learn} and DAD~\citep{foster_dad_2021,ivanova_idad_2021}.
The idea is to exploit the GP priors~\crefp{assump-gp_prior} before AL deployments.
We use $p(f), p(q)$ and the Gaussian likelihoods
$p(y | \bm{x}, f)=\mathcal{N}\left(
y | f(\bm{x}), \sigma^2
\right)$,
$p(z | \bm{x}, q)=\mathcal{N}\left(
z | q(\bm{x}), \sigma_{q}^2
\right)$
to construct a simulator.
Notably, we go beyond DAD by considering the safety function $q$ and nonparametric priors on $f$ and $q$.
This allows us to sample functions, simulate policy-based (safe) AL and, crucially, \textit{meta optimize on broad classes of functions}, with an objective encoding an acquisition criterion.

The major challenges are:
(i) our scheme requires differentiable objectives, where applying conventional formulations is difficult;
(ii) sampling $f,q$ is not trivial, especially with a constraint on $q$.
We summarize our approach in~\cref{alg-asal_training}, and give details next.

\subsection{Training Objective}\label{section-method-training_objective}

Assume, we are given a batch of GP functions $f$, each coupled with a safety GP function $q$.
Initial evaluations $\mathcal{D}_0 = \{\bm{X}_{\text{init}}, Y_{\text{init}}, Z_{\text{init}}\}$ are given, noises are denoted by $Y_{\text{init}} = f(\bm{X}_{\text{init}}) + \mathcal{E}_{\text{init}}, Z_{\text{init}} = q(\bm{X}_{\text{init}}) + \mathcal{E}_{q, \text{init}}$.
We run our policy on each $(f,q)$ pair for $T_{\text{sim}} \leq T$ iterations to obtain $\bm{x}_{\phi,1:T_{\text{sim}}}$, evaluations $y_{\phi,1:T_{\text{sim}}}$ and $z_{\phi,1:T_{\text{sim}}}$, $y_{\phi,t}=f(\bm{x}_{\phi,t})+\epsilon_{t}$ and $z_{\phi,t}=q(\bm{x}_{\phi,t})+\epsilon_{q,t}$ for $t=1,...,T_{\text{sim}}$ (L.5-11 of~\cref{alg-asal_training}).
In this section, we introduce our training objectives (L.13 of~\cref{alg-asal_training}).
The details of $f, q$ and $\mathcal{D}_0$ sampling will be described in~\cref{section-method-function_sampling} (L.1,2,4 of~\cref{alg-asal_training}).

\paragraph{Unconstrained AL--Simulated Acquisition}
\label{section-method-training_objective-unconstrained_objective}

Assume for now that safety conditions are ignored.
We are then dealing with an unconstrained AL: we collect $y_{\phi, 1:T_{\text{sim}}}$ informative for $f$.
The GP prior $f \sim \mathcal{GP}(\bm{0}, k_{\theta})$ further turns this into an active GP learning problem.
This means common acquisition functions~\citep{Seo2000GaussianPR, guestrin_mi_al_05, krause08a} are valid to guide our queries to high information.
Note, however, that we now optimize w.r.t. the policy where gradient is propagated from \textit{all queries jointly}.
A good choice here is the entropy~\citep{Seo2000GaussianPR,krause_nonmyopic_al_07}, which
(i) has closed forms,
(ii) is differentiable, and (iii) is one of the gold standard acquisition criteria for GPs.

As opposed to an AL deployment, the sampled $y_{\phi,1:T_{\text{sim}}}$ are available when we optimize the queries $\bm{x}_{\phi, 1:T_{\text{sim}}}$.
For each AL instance, we take the following acquisition
\begin{align}
\begin{split}\label{eq-gp_logprob_objective_inner}
h(y_{\phi,1:T_{\text{sim}}}, Y_{\text{init}}) =
&-\log p(y_{\phi,1:T_{\text{sim}}}, Y_{\text{init}})
\\
\propto
h(y_{\phi,1:T_{\text{sim}}} | Y_{\text{init}})
=&-\log p(y_{\phi,1:T_{\text{sim}}} | Y_{\text{init}}),
\end{split}
\end{align}
where $\mathbb{E}_{ f, T_{\text{sim}}, \mathcal{E}_{\text{init}}, \epsilon_{1:T_{\text{sim}}} }\left[ h(y_{\phi,1:T_{\text{sim}}}, Y_{\text{init}}) \right]$, an average of various instances, represents the policy's entropy defined by~\citet{krause_nonmyopic_al_07} (originally for conventional AL).
$p(\cdot)$ is a GP likelihood~\crefp{appendix-gp_details}.
The proportionality symbol here indicates equivalency which holds by applying Bayes rule and removing the part that has no gradient w.r.t. the policy.
The effect of $T_{\text{sim}}$ will be described shortly after, let us say $T_{\text{sim}}=T$ is fixed here.
Maximizing this acquisition criterion means the policy selects points that are the most distinctive to each other.
We will later see that this choice has the advantage of explainability in conjunction with safety constraints.
In our~\cref{appendix-objective_details}~\crefp{figureS-acq_safeprob}, we illustrate the acquisition value with $T=1$.

The ideas until here are similar to~\citet{Chen2017_learn_to_learn,foster_dad_2021,ivanova_idad_2021}, i.e. turning the acquisition criteria we would have optimized sequentially into trainable objectives in an a priori simulated learning.

We further propose the regularized entropy criterion:
\begin{align}
\begin{split}
\label{eq-gp_logprob_reduction_objective_inner}
I(y_{\phi,1:T_{\text{sim}}} | Y_{\text{init}})=
&-\log p(y_{\phi,1:T_{\text{sim}}} | Y_{\text{init}})\\
&+\log p( y_{\phi,1:T_{\text{sim}}} | Y_{\text{init}}, Y_{\text{grid}} ),
\end{split}
\end{align}
where $Y_{\text{grid}}$ are noisy evaluations at randomly sampled $\bm{X}_{\text{grid}} \subseteq \mathcal{X}$ (with $|\bm{X}_{\text{grid}}| \gg T $). 
This $I(\cdot)$ is adapted from the mutual information AL criterion~\citep{guestrin_mi_al_05,krause08a}.
It encourages the policy to look inside the input space and avoids over-emphasizing the border of $\mathcal{X}$, a problem of entropy.
We put details into~\cref{appendix-objective_details}.


\paragraph{Unconstrained AL Objective, beyond Fixed Length \& Fixed Priors}
\label{section-method-training_objective-unconstrained_random_sequence}

One remark of the previous objectives is that they are nonmyopic\footnote{Nonmyopic exploration allocates multiple points jointly, e.g. query $1/3, 2/3$ in $[0, 1]$ instead of $1/2, 1/4$.\label{footnote-nonmyopic}}~\citep{krause_nonmyopic_al_07, foster_dad_2021}, assuming a pre-defined budget $T$.
The resulting queries can be suboptimal if we deploy for fewer steps.
In practice, it is often unrealistic to know the precise number of queries $T$ in advance, especially if we wish to deploy the trained policy on multiple problems.
To this end, we assign random AL budget to each $f$, i.e. $T_{\text{sim}} \sim \text{Uniform}[1, T]$.
Then, the exploration scores are normalized by the sequence length (AL sims indicate random $f, T_{\text{sim}}, \mathcal{E}_{\text{init}}, \epsilon_{1:T_{\text{sim}}}$): 
\begin{align}
\label{eq-gp_logprob_objective}
\mathcal{H}(\phi) =
\mathbb{E}_{
\theta, \sigma^2
}
\mathbb{E}_{
\text{AL sims}
}
\left[
\frac{h(y_{\phi,1:T_{\text{sim}}} | Y_{\text{init}})}{N_{\text{init}}+T_{\text{sim}}}
\right],
\end{align}
\begin{align}
\label{eq-gp_logprob_reduction_objective}
\mathcal{I}(\phi)=
\mathbb{E}_{
\theta, \sigma^2
}
\mathbb{E}_{
\text{AL sims}
}
\left[
\frac{
I(y_{\phi,1:T_{\text{sim}} | Y_{\text{init}})
}}{
N_{\text{init}}+T_{\text{sim}}
}
\right]
.
\end{align}

Crucially, the NN needs a budget variable as input to encode the number of queries. 
Without this budget variable, $T_{\text{sim}}=T$ must be fixed.
Due to the space limit, the NN structure is described in~\cref{appendix-nn}.
Eqs.~\eqref{eq-gp_logprob_objective},~\eqref{eq-gp_logprob_reduction_objective} generalize over diverse functions by sampling the GP hyperparameters $\theta, \sigma^2$.


\paragraph{Safe AL Objective}
\label{section-method-training_objective-safe_objective}

\begin{algorithm}[t]
\captionof{algorithm}{Safe AL Policy Training}
	\label{alg-asal_training}
	\begin{algorithmic}[1]
		\Require  \cref{assump-gp_prior}, $T$, $N_{\text{init}}$
		\State draw a batch of $(\theta, \sigma^2, \theta_{q}, \sigma_{q}^2)$
		\State draw a batch of $(f, q)$ pairs~\crefp{alg-initial_sample}
            \For{each $f,q$}
		\State given $\mathcal{D}_0$ per~\cref{alg-initial_sample}
            \State draw $T_{\text{sim}}\sim \text{Uniform}[1, T]$
		\For{$t=1, ..., T_{\text{sim}}$}
		\State $\bm{x}_{\phi,t} = \phi(T_{\text{sim}} - t + 1, \mathcal{D}_{t-1})$
		\State draw $\epsilon_{t} \sim \mathcal{N}(0, \sigma^2)$, $y_{\phi,t}=f(\bm{x}_{\phi,t}) + \epsilon_t$
        \State draw $\epsilon_{q,t} \sim \mathcal{N}(0, \sigma_{q}^2)$, $z_{\phi,t}=q(\bm{x}_{\phi,t}) + \epsilon_{q,t}$
		\State $\mathcal{D}_{t} \gets \mathcal{D}_{t-1} \cup \{ \bm{x}_{\phi,t}, y_{\phi,t}, z_{\phi,t} \}$
		\EndFor
		\EndFor
	\State compute loss per~\cref{eq-safe_gp_logprob_objective}, update $\phi$
	\end{algorithmic}
\end{algorithm}

We have obtained an entropy objective for unconstrained AL.
Now we take $z_{\phi,1:T_{\text{sim}}}$ into consideration and we wish to follow the same intuition to get a safe AL objective.
Recall that a conventional safe AL~\citep{Schreiter2015,ZimmerNEURIPS2018_b197ffde} solves~\cref{eq-safe_al} for each query.
This constrained acquisition criterion has proven its effectiveness, and we next translate it into a differentiable objective compatible with our simulated environment.

If we query with our unconstrained entropy objective step-wise, the corresponding acquisition function is $a(\bm{x})=-\log p(\hat{y}(\bm{x}) | y_{\phi,1:{t}}, Y_{\text{init}})$, where $\hat{y}(\bm{x})$ is the simulated noisy function.\footnote{
$\hat{y}(\bm{x}_{\phi,t})=y_{\phi,t}$. A standard sequential learning would use a random variable $y(\bm{x})$ to forecast the future query, leveraging $a(\bm{x})=\mathbb{H}(y(\bm{x}) | y_{\phi,1:{t}}, Y_{\text{init}})$.
\label{footnote-acquisition_rv}
}
Plugging this $a(\cdot)$ into~\cref{eq-safe_al}, we see that at each step $t+1$, the conventional safe AL objective
optimizes $-\log p(\hat{y}(\bm{x}) | y_{\phi,1:t}, Y_{\text{init}})$, constrained to $p(z(\bm{x}) \geq 0 | z_{\phi, 1:t}, Z_{\text{init}}) \geq 1-\gamma$, $z(\bm{x})$ is a prediction.
In a Lagrange perspective, a constrained optimization may be transformed to a problem where we optimize the main term regularized by a factor of the constraint term~\citep{Nocedal2006,gramacy2015modelingaugmentedlagrangianblackbox}.
The factor, i.e. Lagrange multiplier, typically needs to be optimized as well, but we take the density interpretation to fix it.
We propose to augment the problem and twist the Lagrangian form:

\begin{enumerate}
\item
$p(z(\bm{x}) \geq 0 | z_{\phi, 1:t}, Z_{\text{init}}) \geq 1-\gamma$ is equivalent to $\log p(z(\bm{x}) < 0 | z_{\phi, 1:t}, Z_{\text{init}}) \leq \log(\gamma)$.
\item
We consider a Lagrangian form
\begin{align*}
\begin{split}
\text{argmax}_{\bm{x}}
\{
-&\log p(\hat{y}(\bm{x}) | y_{\phi,1:t}, Y_{\text{init}})\\
- \lambda \log p(z(\bm{x}) &< 0 | z_{\phi, 1:t}, Z_{\text{init}}) + \lambda \log \gamma 
\},
\end{split}
\end{align*}
where $\lambda$ is a Lagrange multiplier.
We fix $\lambda=1$ as the likelihood terms can be interpreted as a joint likelihood of events $\hat{y}$ and unsafe $z$.
Note in addition that $\log \gamma$ is a constant w.r.t. $\bm{x}$, which can be omitted.
This form has thus become
\begin{align*}
\begin{split}
\text{argmax}_{\bm{x}}
\{
-&\log p(\hat{y}(\bm{x}) | y_{\phi,1:t}, Y_{\text{init}})\\
-&\log p(z(\bm{x}) < 0 | z_{\phi, 1:t}, Z_{\text{init}})
\},
\end{split}
\end{align*}
\item
Importantly, we wish to preserve the parameter $\gamma$ as it defines the desired safety level which helps to balance between exploration and safety. 
We thus adjust the previous form into
\begin{align}
\label{eq-acq_minunsafe}
\text{argmax}_{\bm{x}}
\{-&\log p(\hat{y}(\bm{x}) | y_{\phi,1:t}, Y_{\text{init}})
\\ \nonumber
- \log \ &\text{max}(\gamma, p(z(\bm{x}) < 0 | z_{\phi, 1:t}, Z_{\text{init}}) ) \},
\end{align}
where $\gamma$ disregards the exact safety level when the likelihood of being unsafe is small enough.
\item
Similar to an unconstrained AL, we sum up the step-wise acquisition scores to get a joint safe AL objective which is differentiable w.r.t. the policy.
\end{enumerate}

Our safe AL training objective is thus
\begin{align}
\label{eq-safe_gp_logprob_objective}
\mathcal{S}_{\mathcal{H}}(\phi)
&=
\mathbb{E}
\left[
\frac{
S(\mathcal{D}_{T_{\text{sim}}})
}{
N_{\text{init}}+T_{\text{sim}}
}
\right], \text{with }
\\ \nonumber
S(\mathcal{D}_{T_{\text{sim}}})&=
-\log p(y_{\phi,1:T_{\text{sim}}} | Y_{\text{init}})
\\ \nonumber
-\sum_{t=0}^{T_{\text{sim}}-1} &\log \ \text{max}( \gamma, p(z(\bm{x}_{\phi,t+1}) < 0 | z_{\phi, 1:t}, Z_{\text{init}}) )
.
\end{align}
The expectation is over GP hyperparameters and AL instances, and $z_{\phi, 1:0}=\emptyset$.
$z(\bm{x}_{\phi,t+1})$ is a prediction here, because a realization $z_{\phi,t+1}$ cannot have a likelihood of $z_{\phi,t+1} < 0$.

Maximizing $\mathcal{S}_{\mathcal{H}}(\phi)$ corresponds to maximizing the exploration score of $y_{\phi,1:T_{\text{sim}}}$, while minimizing the likelihood of queries appearing unsafe.
In conventional safe AL, $\gamma \in [0, 1]$ (but usually $<0.5$) allows us to specify the level of safety criticality.
In our setting, this parameter balances safety and AL exploration, offering a more intuitive approach than optimizing a Lagrange multiplier.
Setting $\gamma \rightarrow 0$ means that the safety term $\log p(z(\bm{x}_{\phi,t+1}) < 0 | z_{\phi, 1:t}, Z_{\text{init}})$ is optimized towards $-\infty$, causing safety to dominate the loss and overriding exploration (see ablation studies in~\cref{appendix-ablation-safe_AL_gamma}).
We set $\gamma = 0.05$ similar to many conventional safe AL~\citep{cyli2024}, resulting in a safe yet explorative policy.
In~\cref{appendix-objective_details-safe_objectives}, we provide an alternative objective maximizing exploration directly alongside the likelihood of queries being safe, i.e. $\log p(z(\bm{x}_{\phi,t+1}) \geq 0 | z_{\phi, 1:t}, Z_{\text{init}})$.
This loss cannot accommodate safety criticality control via $\gamma$ due to the concavity of $\log(\cdot)$;~\cref{appendix-ablation-safe_al_criteria} (ablation studies) shows that this alternative is less safe than our primary loss.
\cref{appendix-objective_details-safe_objectives} further illustrates the objectives for $T_{\text{sim}}=1$~\crefp{figureS-safe_acq}.
This illustration demonstrates that our main acquisition loss~\crefp{eq-acq_minunsafe,eq-safe_gp_logprob_objective} features a safe plateau, and optimizing this loss tends to select points within this region.

Our objective function $\mathcal{S}_{\mathcal{H}}(\phi)$ can vary with the order of queried data, as the order directly impacts the safety score.
This inherits the conventional property that safety confidence adapts to every new observation.
Note that the~\cref{eq-acq_minunsafe} can itself be a safety-aware yet unconstrained differentiable acquisition criterion for conventional safe AL setups (see ablation studies in~\cref{appendix-ablation-safe_al_criteria}).

\subsection{Function Sampling}\label{section-method-function_sampling}

In the previous section, we introduced an algorithm based on generative GP functions $f, q$, and an initial dataset $\mathcal{D}_0$. These functions were used to simulate safe AL deployments for policy training as summarized in~\cref{alg-asal_training}.
Here, we detail the generative process (sampling steps: lines 1–4, 8, 9), focusing on: (i) sampling the initial dataset $\mathcal{D}_0$ with a suitable prior mean for the safety function $q$, and (ii) efficient GP function sampling for $f$ and $q$. 
The first point is critical for ensuring realistic synthetic data, the second one for efficient and stable training.


\begin{algorithm}[b]
\captionof{algorithm}{Function and Initial Sampling}
\label{alg-initial_sample}
\begin{algorithmic}[1]
\Require \cref{assump-gp_prior}, center $\mathcal{C} \subseteq \mathcal{X}$, $N_{\text{init}}$,
$\mathcal{D}=\emptyset$,
$\mathcal{GP}(0, k_{\theta}), \mathcal{GP}_{\theta_{q}}(\mu_{q}, k_{q}), \sigma^2, \sigma_{q}^2$,
max\_iter$=50$
\State draw $f_{\text{raw}} \sim \mathcal{GP}(0, k_\theta)$, $q_{\text{raw}} \sim \mathcal{GP}(0, k_{q})$
\State $f(\cdot)=f_{\text{raw}}(\cdot) - \mathbb{E}_{\bm{x}\in\mathcal{X}}[f_{\text{raw}}(\bm{x})]$
\State $q(\cdot)=\mu_{q}(\cdot) + q_{\text{raw}}(\cdot) - \mathbb{E}_{\bm{x}\in\mathcal{X}}[q_{\text{raw}}(\bm{x})]$
\For{i=1,..., max\_iter}
\State draw $\bm{X} \sim \text{Uniform}[\mathcal{C}], |\bm{X}|=N_{\text{init}}$
\State draw $\mathcal{E}_{\text{init}} \sim \mathcal{N}(0, \sigma^2 I), Y=f(\bm{X})+\mathcal{E}_{\text{init}}$
\State draw $\mathcal{E}_{q, \text{init}} \sim \mathcal{N}(0, \sigma_{q}^2 I), Z=q(\bm{X})+\mathcal{E}_{q, \text{init}}$
\State take those safe: $\mathcal{D} \gets \mathcal{D} \cup [\bm{X}, Y, Z]_{z \geq 0}$
\State \textbf{if} i$=$max\_iter \textbf{then} $\mathcal{D} \gets \mathcal{D} \cup (\bm{X}, Y, Z)$
\State \textbf{if} $\vert\mathcal{D}\vert \geq N_{\text{init}}$ \textbf{then} take first $N_{\text{init}}$ and break 
\EndFor
\State \textbf{return} $f, q, \mathcal{D}_0=\mathcal{D}$ ($|\mathcal{D}|=N_{\text{init}}$)
\end{algorithmic}
\end{algorithm}

\paragraph{Safety Function $q$ and Initial Data $\mathcal{D}_{0}$}
\label{section-method-function_sampling-q_and_D0}

The first challenge is to ensure that the simulated $q$ and $\mathcal{D}_0$ are both realistic and representative.
In safe exploration problems~\citep{sui15safeopt,ZimmerNEURIPS2018_b197ffde,alessandro2022safe}, the initial data are usually provided by a domain expert at the centric area of a safe set and the algorithms gradually explore towards the safe set border.
Following this principle, our generative algorithm samples initial data $\mathcal{D}_0$ in a pre-defined safe set, and we assume w.l.o.g. that the safe set lies at the center of $\mathcal{X}$, denoted by $\mathcal{C} \subseteq \mathcal{X}$.
This approach ensures that $\mathcal{D}_0$ closely resembles the initial datasets encountered in the deployment stage.


For this, we construct a GP prior $q \sim \mathcal{GP}_{\theta_{q}}(\mu_{q}, k_{q})$ to increase the likelihood of a safe initial dataset $\mathcal{D}_{0}$.
Specifically, we design a prior mean $\mu_q$ that ensures safety around the center, e.g. $\mathcal{C}=[0.4, 0.6]^D$ for $\mathcal{X}=[0,1]^D$.
As a result, $q(\mathcal{C})$ may likely have a safe region to initiate a safe AL simulation, while still allowing for a diverse set of safety functions by introducing variability within the GP.
In~\cref{appendix-training_samplers}, we provide the exact expression for our mean function $\mu_q$, which is designed to allow for a safe set of varying shape and size, while maintaining consistency with the deployment problem's normalization assumptions.
Examples of $q$ are illustrated in~\cref{figureS-q_examples}.

When we sample the initial dataset $\mathcal{D}_{0}$ in \cref{alg-initial_sample}, we sample safe data from $\mathcal{C}$, but once a maximum number of iteration is reached, the training algorithm proceeds regardless of whether all sampled points are safe.

\paragraph{Fourier Feature Functions - Efficient and Decoupled}
\label{section-method-function_sampling-fff}

The final step is to devise an efficient sampling strategy for the noisy GP values $y_{\phi,1:T_{\text{sim}}}$ and $z_{\phi,1:T_{\text{sim}}}$.
This is not trivial because the observations are sampled iteratively, e.g. $\bm{x}_{\phi, t}=\phi(\mathcal{D}_{t-1})$, meaning $y_{\phi,1:t-1}$, $z_{\phi,1:t-1}$ are sampled before we know $\bm{x}_{\phi, t},..., \bm{x}_{\phi, T}$.
One can make standard GP posterior sampling conditioned on preceding samples, e.g. $y_{\phi, t} \sim p(y(\bm{x}_{\phi, t}) | \mathcal{D}_{t-1}, k_\theta, \sigma^2)$.
However, when performed naively, this posterior sampling approach is bound to the GP cubic complexity for each data point acquisition $y_{\phi, t}$, which is prohibitive. 
The runtime can be further reduced by using low-rank updates~\citep{seeger2004low}.
For specific objectives such as for our $\mathcal{H}$ or $\mathcal{S}_{\mathcal{H}}$, the GP posteriors can be then computed as intermediate results with minimal additional cost (see~\cref{eq-acq_minunsafe,eq-safe_gp_logprob_objective}).
However, low-rank updates can lead to additional issues, as they make the computational graph unnecessarily complex for the loss function differentiation.
To overcome these challenges, we propose a decoupled approach, which is scalable by itself and enables flexible integration of any objectives, e.g. potentially cheap approximated objectives, for future work.

The core is a decoupled function sampling technique \citep{Rahimi2007_rff,Wilson2020_icml_rff_gp_post}, where each GP function is approximated by a linear combination of Fourier features.
The function can be evaluated later at any $\bm{x}\in\mathcal{X}$ in linear time (line 8-9 of~\cref{alg-asal_training}).
One requirement, however, is that the kernels $k_{\theta}, k_{q}$ need to have Fourier transforms (e.g. stationary kernels, see Bochner's theorem in~\citealt{GPbook}).
For $q\sim \mathcal{GP}(\mu_{q}, k_{q})$, one may sample $q_{\text{raw}}\sim \mathcal{GP}(0, k_{q})$ and set $q = \mu_{q} + q_{\text{raw}}$.
Note that a sampled function does not necessarily average to zero on $\mathcal{X}$, but we may compute a domain specific average analytically with usually negligible complexity (\cref{appendix-training_samplers-fff}).
This is preferred because we consider normalized deployment problems, and such an optional mean shift improves the performance.


While not explored in this paper, we point out that alternative approaches might exist for efficient sampling, such as linear system solvers~\citep{lin2024improving} or different random features~\citep{tripp2023tanimoto}.

\section{Experiments}\label{section-experiments}

\begin{table}[t]
\scriptsize
\centering
\setlength{\tabcolsep}{3.5pt}
\caption{\textbf{RMSE on standard AL tasks}. Performance on the test set. 
Our method shows competitive results while being orders of magnitude faster (\cref{figure-result_al}).
\cref{tableS-result_al_varied_T} show results of smaller $T$.
}
\label{table-result_al}
\begin{tabular}{l|cccc}
\toprule
& Sinus & Airline & Branin & LGBB\\
$N_{\text{init}}{+}T$ & ($1{+}20$) & ($1{+}20$) & ($1{+}30$) & ($1{+}30$)\\
\midrule
Our AAL    & $0.14{\pm}0.004$ & $0.41{\pm}0.022$ & $0.21{\pm}0.020$ & $0.17{\pm}0.013$ \\
ALINE      & $0.38{\pm}0.034$ & $0.43{\pm}0.011$ & $0.27{\pm}0.013$ & $0.17{\pm}0.009$ \\
DAD        & $0.49{\pm}0.065$ & $0.43{\pm}0.018$ & $0.99{\pm}0.128$ & $0.48{\pm}0.063$ \\
PFN\_AL    & $0.85{\pm}0.088$ & $0.44{\pm}0.041$ & $0.38{\pm}0.019$ & $0.19{\pm}0.008$ \\
TabPFN\_AL & $1.04{\pm}0.033$ & $0.43{\pm}0.023$ & $0.22{\pm}0.008$ & $0.16{\pm}0.010$ \\
AGP\_AL    & $0.14{\pm}0.007$ & $0.48{\pm}0.020$ & $0.23{\pm}0.025$ & $0.22{\pm}0.006$ \\
GP\_AL     & $0.13{\pm}0.009$ & $0.43{\pm}0.038$ & $0.19{\pm}0.011$ & $0.17{\pm}0.010$ \\
SVGP\_AL   & $0.12{\pm}0.004$ & $0.42{\pm}0.012$ & $0.34{\pm}0.003$ & $0.17{\pm}0.005$ \\
MGP\_AL    & $0.12{\pm}0.007$ & $0.39{\pm}0.020$ & $0.17{\pm}0.006$ & $0.15{\pm}0.008$ \\
Random     & $0.33{\pm}0.055$ & $0.41{\pm}0.023$ & $0.28{\pm}0.052$ & $0.21{\pm}0.050$ \\
\bottomrule
\end{tabular}
\vspace{-10pt}
\end{table}

\begin{figure}[t]
\begin{center}
\centerline{\includegraphics[width=\linewidth]{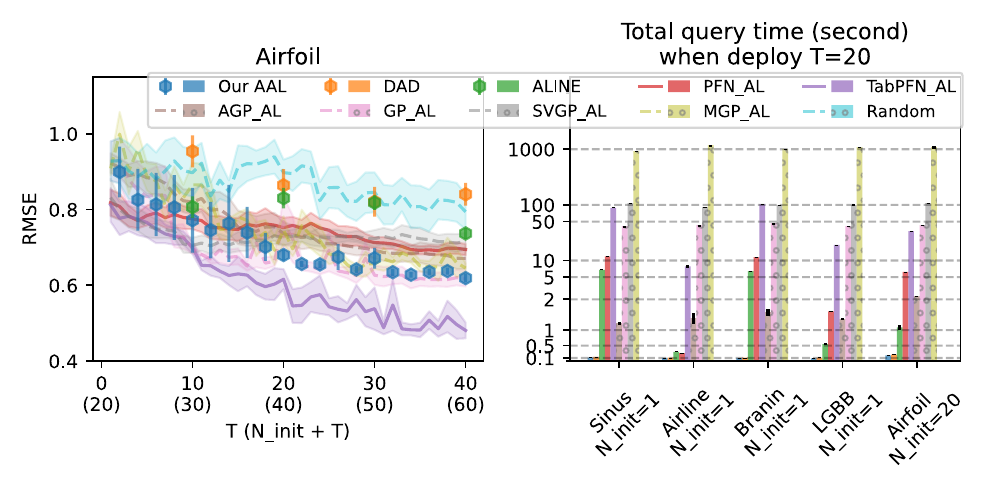}}
\caption{
\textbf{Empirical results on standard AL.}
Left: RMSE on airfoil dataset vs number of queries $T$.
Our trained policy is deployed at $T=2, 4, \dots, 40$.
DAD and ALINE train separate policies for each $T$ (shown for $T=10, 20, 30, 40$).
Right: total query time at $T=20$ across datasets.
Our approach is significantly faster, requiring only a single NN forward pass for each data acquisition.
}\label{figure-result_al}
\end{center}
\vskip -0.3in
\end{figure}

In this section, we empirically evaluate our method. 
We first present unconstrained AL results, demonstrating the effectiveness of our approach in terms of AL exploration.
Next, we incorporate safety constraints and show that our approach remains effective in exploration while maintaining safety requirements.
All experiments confirm that our methods are well suited for AL tasks requiring real-time querying.
Our code is available at \url{https://github.com/cenyou/ASAL}.

\paragraph{Experimental Setup} We prepare the experiments by training our neural network policy utilizing~\cref{alg-asal_training}, corresponding to the up-front preparation block in~\cref{figure1}.
We train one NN policy for each dimensionality $D$ and for each experimental setting (unconstrained, safe). 
The policies are then applied to all benchmark problems of their respective configuration, corresponding to the deployment block in~\cref{figure1}.
Our NN policy returns points on continuous space $\mathcal{X} \subseteq \mathbb{R}^D$.
On benchmark functions, the query $\bm{x}_t = \phi(\cdot)$ is used directly for data acquisition. 
For testing on discrete dataset, we approximate the query by selecting the nearest point from the pool using the $L2$-norm.
Since our high-level goal is to model a regression task, we use the final collected dataset to train a GP model and assess its performance.
All experiments report modeling performance, measured with Root Mean Squared Error (RMSE), and AL deployment time.
For safe AL, we additionally study if the safety threshold is satisfied.
Each AL task is repeated five times with different random seeds and initial data.
We provide more experimental details in~\cref{appendix-experiment_details}, including the training time~\crefp{tableS-training_time}.

\subsection{Amortized Unconstrained AL (AAL)}

We first study our AAL on standard unconstrained AL tasks by training 
on the objective $\mathcal{I}$~\crefp{eq-gp_logprob_reduction_objective} and removing the safety samples from the generative process in~\cref{alg-asal_training} (see~\cref{alg-aal_training} for a cleaner version of unconstrained training).
We compare AAL against (i) ALINE~\citep{huang2025aline}; (ii) DAD: amortized BED~\citep{foster_dad_2021}; (iii) PFN AL: AL where the model is a PFN~\citep{mueller2022pfn} pretrained on our GP data; (iv) TabPFN AL: AL where the model is the TabPFN foundation model~\citep{hollmann2025tabpfn}; (v) AGP AL: AL with amortized GP (AGP,~\citealt{Bitzer2023amortized_gp}), where GP hyperparameters are not trained but inferred; (vi) GP AL: AL with a vanilla GP (trained on Type II maximum likelihood); (vii) SVGP AL: AL with sparse variational GP~\citep{pmlr-v5-titsias09a,hensman2013gaussian,hensman2015mcmc}, where the distribution is approximated with 15 inducing variables (ivs, pseudo data); (viii) MGP AL: AL with mixture of vanilla GPs~
\citep{riis2023mgp_bal}; (ix) Random: random selection criterion. 
We provide details and deployment complexities in~\cref{appendix-experiment_details-deployment_algs_baselines,appendix-experiment_details-gp_modeling}.
We run experiments on 2 benchmark functions and 3 real-world datasets:
Sin function (1D),
Branin function (2D),
Airline dataset (1D),
Langley Glide-Back Booster dataset (LGBB, 2D), and
Airfoil dataset (5D).

We report the mean and standard errors on the test dataset in~\cref{table-result_al,figure-result_al}, training loss values in~\cref{appendix-ablation}.
AL methods generally outperform random sampling, with the exception of the Airline dataset. 
Among the AL methods, we observe that the performance of AGP sometimes and of DAD often deteriorates (DAD is particularly bad when $D \geq 2$).
AGP~\citep{Bitzer2023amortized_gp} infers hyperparameters to approximate Type II maximum likelihood, which might not fully align with the exploration needs of AL tasks.
DAD is not designed for nonparametric modeling (see~\cref{appendix-objective_details-DAD}).
SVGP approximates vanilla GP with 15 ivs. However, training a SVGP requires more iterations than a vanilla GP while the computation reduction of each iteration is not obvious in our data sizes--resulting in a higher time cost.
MGP method provides the strongest modeling results in most of the tasks, but requires multiple GP models and is around 3000x slower than our AAL.
ALINE, PFN and TabPFN have the same deployment complexity up to a constant factor~\crefp{appendix-experiment_details-deploy_complexity,tableS-deploy_complexities}.
ALINE and PFN show comparable learning outcomes and TabPFN learns particularly good on the Airfoil dataset; however, they are slower than our approach in all scenarios due to attention to the pool.
For 20 queries, our AAL spends $< 0.3$ (s), around 10x-3000x faster than AGP's $< 3$ (s), vanilla GP's $< 50$ (s), SVGP's $< 120$ (s), and MGP's $16$ (min).
When compared to other amortized approaches (ALINE, PFN, and TabPFN), AAL is at least 5x faster, except on the small Airline dataset (less than 200 points split into observation, validation, and pool sets) where it is roughly 2.5x faster.
Additionally, our method offers a significant advantage in training time compared to ALINE; because ALINE relies on a discretized search pool, its training time is an order of magnitude longer~\crefp{tableS-training_time}.

In summary, our results clearly highlight the advantages of our method on time-sensitive tasks: 
AAL operates at least an order of magnitude faster, while remaining highly effective in exploration.

\subsection{Amortized Safe AL (ASAL)}

\begin{figure*}[t]
\begin{center}
\centerline{\includegraphics[width=\linewidth]{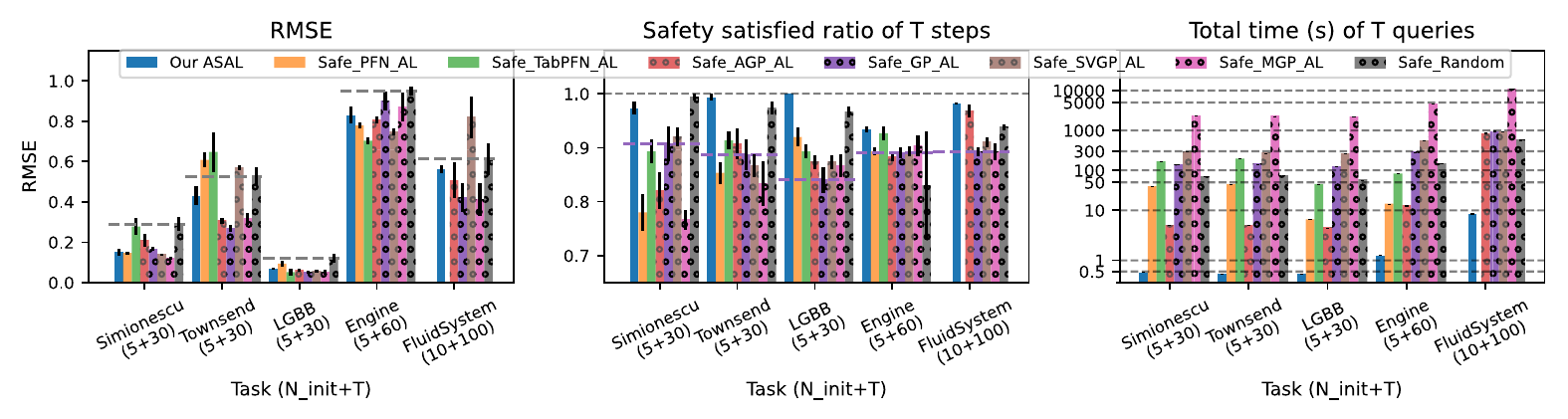}}
\vskip -0.15in
\caption{\textbf{Results on safe AL.}
Left: Our method (blue) achieves competitive RMSE (on safe test data) and outperforms Safe Random across all tasks.
Middle:
The proportion of safe queries out of $T$ queries confirm that our method queries highly safe data.
Right:
Our approach is significantly faster, as we avoid online GP modeling and acquisition optimization.
Safe Random is slow due to the GP-based safe set estimation.
PFN and TabPFN are not feasible on the Fluid System due to the large pool.
}\label{figure-result_sal}
\end{center}
\vskip -0.35in
\end{figure*}

In this section, we study safe AL by training on the 
objective $\mathcal{S}_{\mathcal{H}}$~\crefp{eq-safe_gp_logprob_objective} with~\cref{alg-asal_training}.
We compare our ASAL against conventional model-based safe AL baselines, i.e. solving~\cref{eq-safe_al} for each query decision: 
(i) Safe PFN AL which leverages PFNs for the main $f$ and the safety function $q$ (PFN pretrained on our GP data);
(ii) Safe TabPFN AL where $f$ and $q$ are modeled by the TabPFN foundation model~\citep{hollmann2025tabpfn};
(iii) Safe AGP AL where the models are AGPs~\citep{Bitzer2023amortized_gp};
(iv) Safe GP AL~\citep{Schreiter2015,ZimmerNEURIPS2018_b197ffde,cyli2022} with vanilla GPs (Type II maximum likelihood);
(v) Safe SVGP AL with SVGPs, where the number of ivs are $20$ for the Engine and Fluid System and $15$ for other tasks (see tasks below);
(vi) Safe MGP AL where MGPs~\cite{riis2023mgp_bal} are the models;
(vii) Safe Random, where base acquisition scores are random values and the safety values are estimated by a vanilla GP. 
Except for Safe Random, all the baselines employ constrained entropy as the acquisition criteria.
All methods are deployed with a safety level of $\gamma=0.05$~\crefp{eq-safe_al}.

We deploy safe AL on two constrained benchmark functions and two real-world datasets and a higher-dimensional Fluid System task:
Simionescu function (2D), Townsend function (2D), LGBB dataset (2D), Engine measurements (3D), high-pressure Fluid System (7D).
Our method builds on the standard safe AL criteria~\crefp{eq-safe_al} with a focus on real-time deployment.
Prior work typically focuses on low dimensions ($D \leq 3$ or $4$,~\cref{tableS-benchmark_dimension}) because safe acquisition optimization becomes increasingly burdensome as $D$ grows.
Our experiments on the Fluid System extend beyond the typical dimension. For this task, the training lengthscale of our GP prior is centered on a plausible operating scale for the system but kept sufficiently broad to remain generic (all other training and deployment settings are unchanged).
Note that PFN and TabPFN cannot be deployed on the Fluid System.
We learn the Fluid System actively from a million trajectories; this size of dataset is far beyond what PFNs and TabPFN are designed for. PFNs and TabPFN need to process all candidate queries, consuming memory exceeding the limit of our machine.

Our main results are shown in~\cref{figure-result_sal}, with further analyses presented in~\cref{appendix-ablation}, including an ablation on the effect of $\gamma$ during policy training, and an ablation of conventional safe AL equipped with our novel acquisition criterion~\crefp{eq-acq_minunsafe}.
ASAL achieves comparable model performance to Safe (A)GP AL on all datasets, except for a slight disadvantage on Townsend and the 7D Fluid System, and outperforms Safe Random across the board. 
It ensures safety awareness--the proportion of safe queries exceeds $1-\gamma$ on all tasks. 
The PFN and TabPFN methods have mixed results in terms of performance and safety awareness -- while the methods are comparable to the GP baselines on LGBB and Engine, on Simionescu and Townsend either the performance is bad or the queries are unsafe (Safe GP AL as a reference).
Importantly, ASAL is at least an order of magnitude faster than all PFN-based and GP-based methods (10x-600x), including Safe Random which relies on GPs for safe set estimation.
On the 7D Fluid System, where the complexity of acquisition optimization is significant, the typically fast AGP baseline takes around 8.4 s/query, whereas our ASAL remains below 0.1 s/query, at least 80x faster.
These results highlight the practical advantages of ASAL in real-time data acquisitions.



\section{Conclusion and Discussion}\label{section-conclusion}
We propose AAL and ASAL approaches that leverage GPs to pretrain AL and safe AL policies.
In particular, we introduce a novel safety-aware yet unconstrained, differentiable loss and an efficient sampling scheme via GP Fourier features to simulate data acquisitions.
The trained policy can zero-shot generalize to novel problems,
enabling deployment that (i) queries informative and (if desired) safe data, and (ii) eliminates the need for online GP modeling and (constrained) acquisition optimization, resulting in a significant speed-up that facilitates real-time data acquisition.

Many existing safe learning methods~\citep{sui15safeopt,yanan_sui_stagewise_2018,ZimmerNEURIPS2018_b197ffde,Lederer2021_icml_gprealtimecontrol} offer theoretical safety guarantees, which our method does not provide. However, these guarantees typically rely on the assumption of well-chosen GP hyperparameters a priori--a strong assumption that is rarely met in real-world AL applications.

Our training relies on standard GP-based AL approaches.
Therefore, it inherits limitations from established GP methods and also benefits from progress in GP modeling research.
A future direction here is to extend to higher dimension (10D or higher, which remains challenging in general), for example by incorporating dimension reduction methods or scalable GPs.
Our approach leverages similar idea of amortized BED~\citep{foster_dad_2021,ivanova_idad_2021,huang2024amortized}, where sequential acquisitions are compressed into an offline training objective.
This formulation complicates the integration of multi-stage optimizations, as often required e.g. to incorporate approximate GPs~\citep{pmlr-v5-titsias09a,pmlr-v38-hensman15} (even with known GP hyperparameter samples, approximate posteriors and acquisition criteria would be optimized with distinct objectives) and higher-dimensional AL~\citep{Zhang_etal16moAL_AAAI1611879} and safe learning~\citep{Kirschner2019lineBO,alessandro2022safe}. 


\begin{acknowledgments}

This work was supported by Bosch Center for Artificial Intelligence, which provided financial support and computers, as well as by the German Federal Ministry of Research, Technology and
Space (BMFTR) under the Robotics Institute Germany (RIG).
The Bosch Group is carbon neutral. Administration, manufacturing and research activities no longer leave a carbon footprint.

During the course of this project, Cen-You Li transitioned to a position supported by the Research Council of Finland (Flagship programme: Finnish Center for Artificial Intelligence, FCAI), and Christoph Zimmer transitioned to Baden-W{\"u}rttemberg Cooperative State University.

\end{acknowledgments}

\bibliography{ref}

\section*{Checklist}

\begin{enumerate}

  \item For all models and algorithms presented, check if you include:
  \begin{enumerate}
    \item A clear description of the mathematical setting, assumptions, algorithm, and/or model. [Yes,~\cref{section-problem_statement},~\cref{section-method}]
    \item An analysis of the properties and complexity (time, space, sample size) of any algorithm. [Yes,~\cref{appendix-training_complexity},~\cref{appendix-experiment_details-deploy_complexity}]
    \item (Optional) Anonymized source code, with specification of all dependencies, including external libraries. [Yes, \url{https://github.com/cenyou/ASAL}]
  \end{enumerate}

  \item For any theoretical claim, check if you include:
  \begin{enumerate}
    \item Statements of the full set of assumptions of all theoretical results. [Not Applicable, our work is mainly empirical while assumptions are stated in~\cref{section-problem_statement}]
    \item Complete proofs of all theoretical results. [Not Applicable]
    \item Clear explanations of any assumptions. [Yes]     
  \end{enumerate}

  \item For all figures and tables that present empirical results, check if you include:
  \begin{enumerate}
    \item The code, data, and instructions needed to reproduce the main experimental results (either in the supplemental material or as a URL). [Yes, \url{https://github.com/cenyou/ASAL}; all datasets are publically available (described in~\cref{appendix-experiment_details-datasets})]
    \item All the training details (e.g., data splits, hyperparameters, how they were chosen). [Yes,~\cref{appendix-experiment_details-training_numerical}]
    \item A clear definition of the specific measure or statistics and error bars (e.g., with respect to the random seed after running experiments multiple times). [Yes]
    \item A description of the computing infrastructure used. (e.g., type of GPUs, internal cluster, or cloud provider). [Yes,~\cref{tableS-training_time}~\cref{appendix-experiment_details}]
  \end{enumerate}

  \item If you are using existing assets (e.g., code, data, models) or curating/releasing new assets, check if you include:
  \begin{enumerate}
    \item Citations of the creator If your work uses existing assets. [Yes, we cite existing works we build upon]
    \item The license information of the assets, if applicable. [Yes]
    \item New assets either in the supplemental material or as a URL, if applicable. [Not Applicable]
    \item Information about consent from data providers/curators. [Not Applicable]
    \item Discussion of sensible content if applicable, e.g., personally identifiable information or offensive content. [Not Applicable]
  \end{enumerate}

  \item If you used crowdsourcing or conducted research with human subjects, check if you include:
  \begin{enumerate}
    \item The full text of instructions given to participants and screenshots. [Not Applicable]
    \item Descriptions of potential participant risks, with links to Institutional Review Board (IRB) approvals if applicable. [Not Applicable]
    \item The estimated hourly wage paid to participants and the total amount spent on participant compensation. [Not Applicable]
  \end{enumerate}

\end{enumerate}

\clearpage
\newpage
\appendix
\thispagestyle{empty}
\onecolumn

\aistatstitle{Supplementary Materials}
\setcounter{algorithm}{0}
\setcounter{equation}{0}
\setcounter{table}{0}
\setcounter{figure}{0}
\renewcommand{\thealgorithm}{S.\arabic{algorithm}}
\renewcommand{\theequation}{S.\arabic{equation}}
\renewcommand{\thefigure}{S.\thesection.\arabic{figure}}
\renewcommand{\thetable}{S.\thesection.\arabic{table}}

\part{} 
\parttoc 

\newpage

\section{Gaussian Process: Distribution and Entropy}\label{appendix-gp_details}
\paragraph{GP Distribution}
We first write down the GP predictive distribution.
Details can be seen in~\citet{GPbook}.
We write down a general form with a non-zero prior mean $q \sim \mathcal{GP}(\mu_q, k_q)$.
The GP hyperparameters are omitted for brevity.
One can remove $\mu_q$ to get a zero-mean GP distribution, e.g. for $y$.

Given a set of $N_{\text{observe}}$ data points $\mathcal{D}=\{ \bm{X}, Z \} \subseteq \mathcal{X} \times \mathcal{Z}$, we wish to make inference at points $\bm{X}_{\text{test}} = \{ \bm{x}_{\text{test},1}, ..., \bm{x}_{\text{test},N_{\text{test}}} \}$.
We write $Z_{\text{test}}=(z(\bm{x}_{\text{test},1}), ..., z(\bm{x}_{\text{test},N_{\text{test}}}))$ for brevity.
The joint distribution of $Z$ and predictive $Z_{\text{test}}$ is Gaussian:
\begin{align}\label{eqS-gp_prior_distribution}
\begin{split}
p(Z, Z_{\text{test}})
=\mathcal{N}\left(
\mu_{q}(\bm{X} \cup \bm{X}_{\text{test}}),
k_{q}( \bm{X} \cup \bm{X}_{\text{test}}, \bm{X} \cup \bm{X}_{\text{test}} )
+ \sigma_{q}^2 I_{N_{\text{observe}}+N_{\text{test}}}
\right)
\end{split}
\end{align}
where $k_{q}( \bm{X} \cup \bm{X}_{\text{test}}, \bm{X} \cup \bm{X}_{\text{test}} )$ is a gram matrix with $[k_{q}( \bm{X} \cup \bm{X}_{\text{test}}, \bm{X} \cup \bm{X}_{\text{test}} )]_{i,j}=k_{q} \left( [\bm{X} \cup \bm{X}_{\text{test}}]_i, [\bm{X} \cup \bm{X}_{\text{test}}]_j \right)$.

This leads to the following predictive distribution (or GP posterior distribution)
\begin{align}
\begin{split}\label{eqS-gp_posterior}
p( Z_{\text{test}} | Z)
&=\mathcal{N}\left(
Z_{\text{test}} \vert \mu_{\mathcal{D}}(\bm{X}_{\text{test}}), cov_{\mathcal{D}}(\bm{X}_{\text{test}})
\right),\\
\mu_{\mathcal{D}}(\bm{X}_{\text{test}})&=
k_{q}\left( \bm{X}_{\text{test}}, \bm{X} \right)
\left[k_{q}\left( \bm{X}, \bm{X} \right) + \sigma_q^2 I_{N_{\text{observe}}}\right]^{-1} \left[ Z - \mu_{q}\left( \bm{X} \right) \right] + 
\mu_{q}\left( \bm{X}_{\text{test}} \right),\\
cov_{\mathcal{D}}(\bm{X}_{\text{test}})&=
k_{q}\left( \bm{X}_{\text{test}}, \bm{X}_{\text{test}} \right) + \sigma_q^2 I_{N_{\text{test}}} \\
&-
k_{q}\left( \bm{X}_{\text{test}}, \bm{X} \right)
\left[k_{q}\left( \bm{X}, \bm{X} \right) + \sigma_q^2 I_{N_{\text{observe}}}\right]^{-1}
k_{q}\left( \bm{X}, \bm{X}_{\text{test}} \right).
\end{split}
\end{align}
Elements of the predictive mean vector $\mu_{\mathcal{D}}(\bm{X}_{\text{test}})$ are the noise-free predictive values.

Inverting a $N_{\text{observe}} \times N_{\text{observe}}$ matrix $\left[k_{q}\left( \bm{X}, \bm{X} \right) + \sigma_{q}^2 I_{N_{\text{observe}}}\right]$ has complexity $\mathcal{O}(N_{\text{observe}}^3)$ in time.

When a GP is used to model the probability beyond a threshold, e.g. a predictive safety probability, at a test point $\bm{x}$ this is a cumulative distribution function
\begin{align}
\begin{split}
\label{eqS-gp_posterior_cdf}
p( z(\bm{x}) \geq 0 | Z) =
1 - p( z(\bm{x}) < 0 | Z) =
&
1/2 \left[
1 + erf\left(
\frac{
\mu_{\mathcal{D}}(\bm{x})
}{
\sqrt{2 cov_{\mathcal{D}}(\bm{x})}
}
\right)
\right].
\end{split}
\end{align}

The log probability density function is
\begin{align}
\begin{split}
\label{eqS-gp_posterior_log_prob}
\log p( Z_{\text{test}} | Z)=&
- N_{\text{test}}/2 \log(2 \pi)
- 1/2 \log\det(cov_{\mathcal{D}}(\bm{X}_{\text{test}}))\\
&-{\color{blue}
1/2 (Z_{\text{test}} - \mu_{D}(\bm{X}_{\text{test}}))^T
\left[cov_{\mathcal{D}}(\bm{X}_{\text{test}})\right]^{-1}
(Z_{\text{test}} - \mu_{D}(\bm{X}_{\text{test}}))
},
\end{split}
\end{align}
inverting $cov_{\mathcal{D}}(\bm{X}_{\text{test}})$ or computing the determinant takes $\mathcal{O}(N_{\text{test}}^3)$ in time.

\paragraph{GP Entropy}
If we consider $Z_{\text{test}}$ as a collection of $N_{\text{test}}$ random variables, the entropy is
\begin{align}
\label{eqS-gp_posterior_entropy}
\mathbb{H}( Z_{\text{test}} | Z)=&
\mathbb{E}_{p( Z_{\text{test}} | Z)}[ -\log p( Z_{\text{test}} | Z)]=
N_{\text{test}}/2 \log(2 \pi {\color{blue} e}) + 1/2 \log\det(cov_{\mathcal{D}}(\bm{X}_{\text{test}})).
\end{align}
The difference between $-\log p(\cdot)$ and $\mathbb{H}(\cdot)$ is marked {\color{blue} blue}, and this comparison will be described later when we look into different training objectives.
Note further that if we plug predicted values $Z_{\text{test}}=\mu_{D}(\bm{X}_{\text{test}})$ into~\cref{eqS-gp_posterior_log_prob}, then the blue term becomes zero and this becomes proportional to negative entropy.

\paragraph{GP Training}
When we deploy a conventional (safe) AL, or when we use the collected data after a deployment to model a function, we usually need to select the hyperparameters.
One standard approach is to conduct a Type II maximum likelihood (vanilla GP):
\begin{align*}
\text{argmax}_{\theta_{q}, \sigma^2} \log p(Z | \bm{X}) = \text{argmax}_{\theta_{q}, \sigma^2} \log \mathcal{N}\left(
Z | \mu_{q}(\bm{X}), k_{q}(\bm{X},\bm{X}) + \sigma^2 I_{N_{\text{observe}}}
\right),
\end{align*}
where the free variables are hyperparameters of $\mu_{q}, k_{q}$ and the noise variance $\sigma^2$.
The time complexity is $\mathcal{O}\left( N_{\text{observe}}^3 \right)$ while the exact factor depends on the number of parameters, the optimizer, and the numerical stability.
We keep the prior mean $\mu_q$ for consistent notation even though the only place when we train GPs is during deployment where we use zero mean GPs. 

\section{Policy NN Structure}\label{appendix-nn}
\begin{figure}[t]
	\vskip 0.2in
	\begin{center}
		\centerline{
        \includegraphics[width=0.9\linewidth]{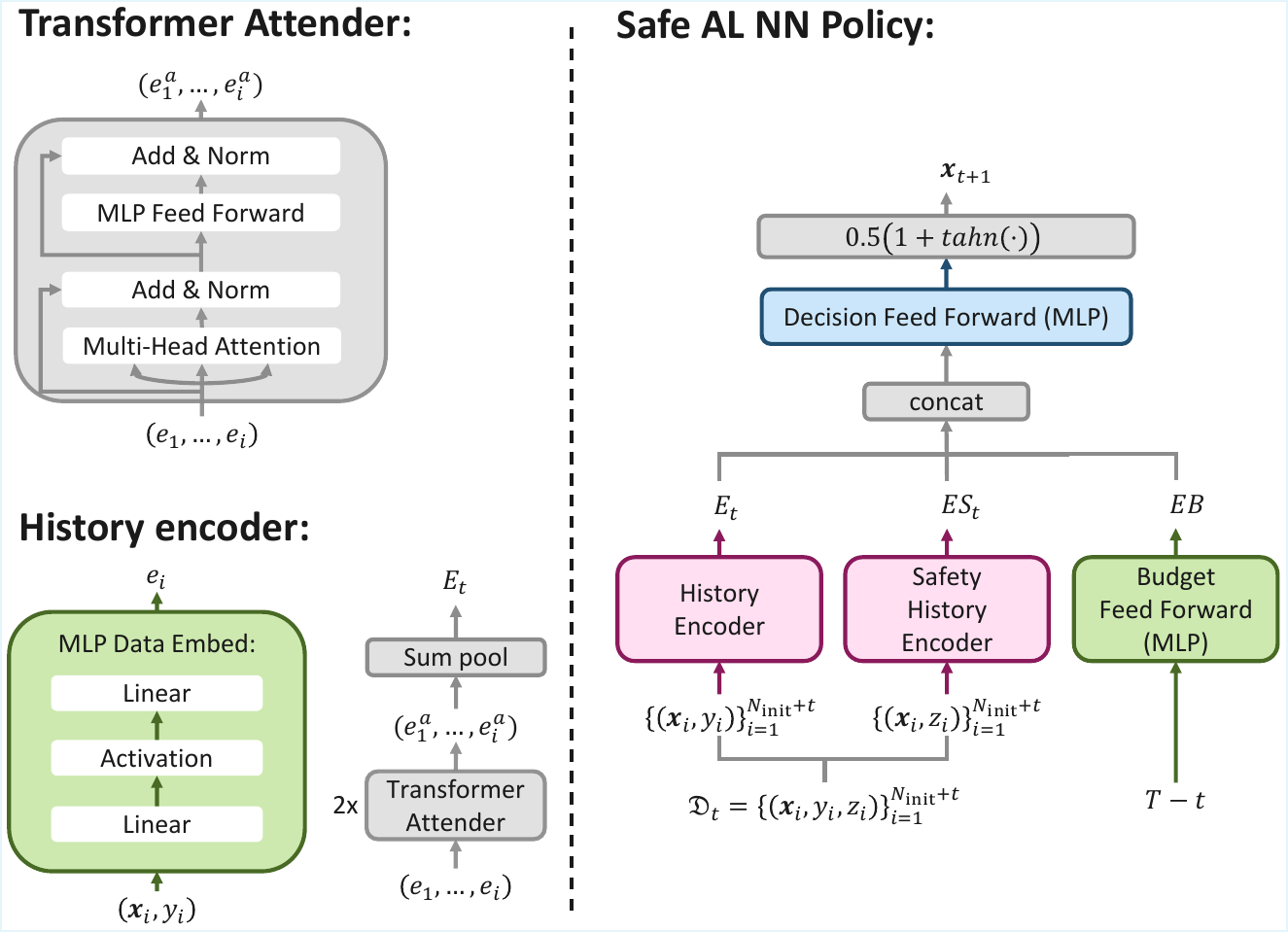
        }
  }
		\caption{
			\textbf{Our NN} $\phi(T-t, \mathcal{D}_{t})=\bm{x}_{t+1}$.
            Each observation pair $(\bm{x}_t, y_t)$ (or $(\bm{x}_t, z_t)$) is first mapped by an MLP to an embedding $e_t$.
            Data of different $t$ use the same MLP Data Embed module, i.e. one history encoder has only one MLP Data Embed.
            A sequence of data embeddings is then stacked and attended by a $2$-layer transformer encoder~\citep{Vaswani2017_attention}, and then a final history embedding is obtained by summing up the attended per-data embeddings $\sum_{t} e_t^a$.
            The sum pool ensures an invariance of input data order.
            Details of the history encoder are described in~\citet{foster_dad_2021,ivanova_idad_2021}.
            The history encoder and safety history encoder are two modules of individual parameters.
            The budget variable is as well mapped by a separate MLP module to a budget embedding $EB$.
            The history embedding $E_t$, safety history embedding $ES_t$, and the budget embedding $EB$ are stacked, mapped by another MLP module, and then refrained by a $tahn$ function to a bounded input domain $[0,1]^D$.
		}
		\label{figureS-nn}
	\end{center}
	\vskip -0.2in
\end{figure}

We build our NN upon the structure described in~\citet{foster_dad_2021,ivanova_idad_2021}.
The final structure is sketched in~\cref{figureS-nn}.
We summarize the comparison between our network and the one developed by~\citet{ivanova_idad_2021}:
\begin{enumerate}
    \item the history encoder ($\{(\bm{x}_i, y_i)\}_{i=1}^t \rightarrow E_t$) and the decision feed forward MLP (originally $E_t \rightarrow \bm{x}_{t+1}$) are taken from~\citet{ivanova_idad_2021};
    \item we add a hyperbolic tangent function as the last layer to ensure the policy output is in our bounded $\mathcal{X}$, which was not needed in the original BED problems;
    \item we add another history encoder to handle the safety data;
    \item we add a budget encoder to handle the budget variable.
\end{enumerate}

Note that the history encoder incorporates the inductive bias that observed data are order-invariant (see~\cite{foster_dad_2021,ivanova_idad_2021} for details).
This can be seen by noticing that a conventional AL computes the acquisition score 
conditioned on the past observations and the order of the past data does not 
matter.
We leverage the same architecture to encode the safety embedding because our safety likelihoods follows the same invariance (future likelihood conditioned on the past, see~\cref{eq-acq_minunsafe}).
We stack the history embedding, the safety history embedding, and the budget 
to reflect that the querying decision is made by balancing the three of them

Our NN structure is highly modularized.
The safety history encoder can be removed to form a budget-aware unconstrained AL policy.
If we intend to work on a policy under fixed budget, i.e. $T_{\text{sim}}$ is fixed to $T$ in~\cref{alg-asal_training}, we can remove the Budget Feed Forward module. 

For the numerical settings, please see~\cref{appendix-experiment_details-training_numerical}.


\section{Training Objectives: Details, Illustrations, More Objectives}\label{appendix-objective_details}

In this section, we provide
\begin{itemize}
    \item an illustration of our entropy objective, another version of our entropy objective and an illustration~\crefp{appendix-objective_details-unconstrained_objectives},
    \item details of our regularized entropy (mutual information) objective, another version of our regularized entropy objective~\crefp{appendix-objective_details-unconstrained_objectives},
    \item details of the DAD baseline in our GP context~\crefp{appendix-objective_details-DAD},
    \item an illustration of our safe AL objective, an additional safe AL objective and an illustration~\crefp{appendix-objective_details-safe_objectives}.
\end{itemize}
This section inherits the notation of~\cref{section-method-training_objective}.
The objectives are computed with simulated (safe) AL instances.

\subsection{Objectives: Unconstrained AL}\label{appendix-objective_details-unconstrained_objectives}

In this section, we provide an illustration of our main unconstrained AL entropy objective, details of our regularized entropy, and additional objectives.

Note first that our unconstrained AL training does not need the safety measurements:
a NN may have no safety history encoder~\crefp{figureS-nn}, and the training algorithm can operate with neither safety measurements $q$ and $z$ nor safety center $\mathcal{C}$.
Therefore, our main safe AL training~\crefp{alg-asal_training} can be reduced to~\cref{alg-aal_training}.

\begin{algorithm}[H]
\captionof{algorithm}{Unconstrained AL Policy Training}
\label{alg-aal_training}
\begin{algorithmic}[1]
\Require \cref{assump-gp_prior}, $T$, $N_{\text{init}}$
\State draw a batch of $(\theta, \sigma^2)$
\State draw a batch of $(f, \mathcal{D}_0)$ per~\cref{alg-unconstrained_initial_sample}
\For{each $(f, \mathcal{D}_0)$}
\State draw $T_{\text{sim}}\sim \text{Uniform}[1, T]$
\For{$t=1, ..., T_{\text{sim}}$}
\State $\bm{x}_{\phi,t} = \phi(T_{\text{sim}} - t + 1, \mathcal{D}_{t-1})$
\State draw $\epsilon_{t} \sim \mathcal{N}(0, \sigma^2), y_{\phi,t}=f(\bm{x}_{\phi,t}) + \epsilon_t$
\State $\mathcal{D}_{t} \gets \mathcal{D}_{t-1} \cup \{ \bm{x}_{\phi,t}, y_{\phi,t} \}$
\EndFor
\EndFor
\State compute AL loss (e.g.~\cref{eq-gp_logprob_reduction_objective}), update $\phi$
\end{algorithmic}
\end{algorithm}
\begin{algorithm}[H]
\captionof{algorithm}{Unconstrained Initial Sampling}
\label{alg-unconstrained_initial_sample}
\begin{algorithmic}[1]
\Require \cref{assump-gp_prior}, $N_{\text{init}}$,
$\mathcal{GP}(0, k_{\theta}), \sigma^2$
\State draw $f_{\text{raw}} \sim \mathcal{GP}(0, k_\theta)$
\State $f(\cdot)=f_{\text{raw}}(\cdot) - \mathbb{E}_{\bm{x}\in\mathcal{X}}[f_{\text{raw}}(\bm{x})]$
\State draw $\bm{X}_{\text{init}} \sim \text{Uniform}[\mathcal{X}], |\bm{X}_{\text{init}}|=N_{\text{init}}$
\State draw $\mathcal{E}_{\text{init}} \sim \mathcal{N}(0, \sigma^2 I)$, $Y_{\text{init}}=f(\bm{X}_{\text{init}})+\mathcal{E}_{\text{init}}$
\State $\mathcal{D}_0 = \{ \bm{X}_{\text{init}}, Y_{\text{init}} \}$
\State \textbf{return} $f, \mathcal{D}_0$
\end{algorithmic}
\end{algorithm}

\subsubsection{Objectives: Unconstrained AL - Entropy}

\begin{figure}[t]
	\vskip 0.2in
	\begin{center}
		\centerline{\includegraphics[width=\linewidth]{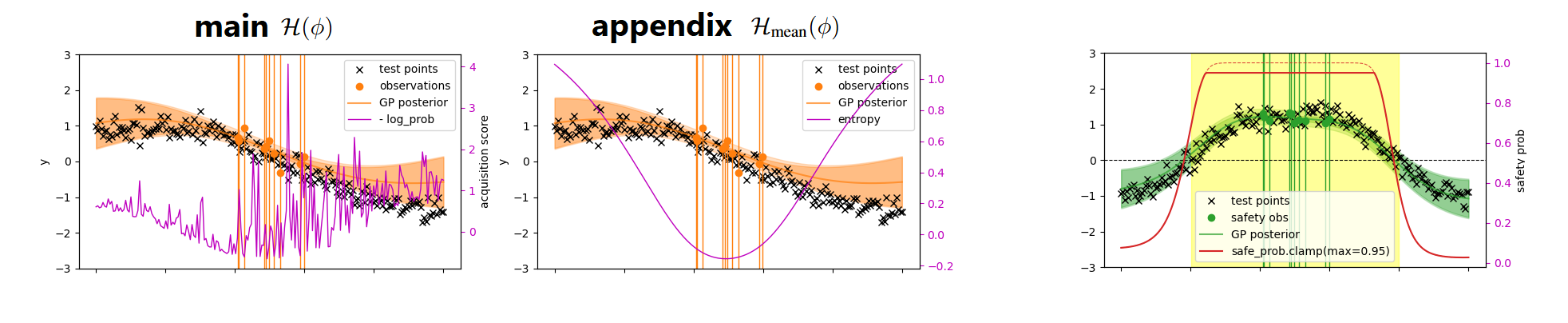}}
		\caption{
\textbf{Simulated unconstrained AL objectives and safety probability.}
$T_{\text{sim}}=1$.
Given observed data ($y$ orange, $z$ green), we plot GP posteriors (orange,~\cref{eqS-gp_posterior}), compute AL objectives (pink) and the safety likelihood $p(z(\cdot)\geq 0 | z_{\text{observed}})$ (red,~\cref{eqS-gp_posterior_cdf}) of the next queries (black).
The yellow region is the true safe area.
The objectives illustrated are meant to be maximized.
Note that $y,z$ are available because our training scheme simulates AL before utilizing the evaluations for objective computations.
A conventional (safe) AL optimizes queries based on inferred future evaluations.
}
		\label{figureS-acq_safeprob}
	\end{center}
	\vskip -0.2in
\end{figure}

Here, we illustrate our main entropy objective, and we introduce another version of the entropy objective. 

In our main paper, we introduce the unconstrained AL meta objective
\begin{align*}
\mathcal{H}(\phi)\propto
&\mathbb{E}_{
\theta, \sigma^2
}
\mathbb{E}_{
f, T_{\text{sim}}, \mathcal{E}_{\text{init}}, \epsilon_{1:T_{\text{sim}}}
}
\left[
\frac{
- \log p(y_{\phi,1}, ..., y_{\phi,T_{\text{sim}}} | Y_{\text{init}})
}{
N_{\text{init}}+T_{\text{sim}}
}
\right]~\text{\crefp{eq-gp_logprob_objective,eq-gp_logprob_objective_inner}}.
\end{align*}

Here, we introduce another objective function which computes
\begin{align}
\label{eq-gp_entropy_objective}
\mathcal{H}_{\text{mean}}(\phi) & = \mathbb{E}_{
\theta, \sigma^2
}
\mathbb{E}_{
f, T_{\text{sim}}, \mathcal{E}_{\text{init}}, \epsilon_{1:T_{\text{sim}}}
}
\left[
\frac{
\mathbb{H}(y(\bm{x}_{\phi,1}), ..., y(\bm{x}_{\phi,T_{\text{sim}}}) | Y_{\text{init}})
}{
N_{\text{init}}+T_{\text{sim}}
}
\right].
\end{align}

Note that entropy $\mathbb{H}$ takes random variables, not the sampled $y_{\phi,1}, ..., y_{\phi,T_{\text{sim}}}$.
Substituting~\cref{eqS-gp_posterior_log_prob,eqS-gp_posterior_entropy} into the objectives, we see that the key difference (marked {\color{blue} blue}) is indeed whether the observation values $y_{\phi,1},...,y_{\phi,T}$ are taken into account.
Our main $\mathcal{H}$ aims for points $y_{\phi,1},...,y_{\phi,T}$ that are the most distinctive, while $\mathcal{H}_{\text{mean}}$ aims for $\bm{x}_{\phi,1},...,\bm{x}_{\phi,T}$ that have the most uncertainty jointly (no actual output values).
We suspect that having $y_{\phi,1},...,y_{\phi,T}$ in the loss (our main $\mathcal{H}$,~\cref{eq-gp_logprob_objective}) may help the policy adapt in an AL deployment.
Please see~\cref{figureS-acq_safeprob} for illustrations.
One can see that $\mathcal{H}$ has the evaluated $y_{\phi,1:T_{\text{sim}}}$ which encodes output values and noises.

The subscript name is "mean" because $\mathbb{H}$ computes the expectation of negative log likelihood.

Note that, if we would obtain each $\bm{x}_{\phi, t+1}$ step-wise, the objectives $\mathcal{H}, \mathcal{H}_{\text{mean}}$ correspond to the following acquisition function: 
\begin{align}\label{eqS-gp_entropy_objective_acquisition_function}
\begin{split}
a_{\mathcal{H}}(\bm{x} | y_{\phi,1:t}, Y_{\text{init}})
&=
- \log p(\hat{y}(\bm{x}) | y_{\phi,1:t}, Y_{\text{init}}),\\
a_{\mathcal{H}_{\text{mean}}}(\bm{x} | y_{\phi,1:t}, Y_{\text{init}})
&= \mathbb{E}_{y(\bm{x}) | y_{\phi,1:t}, Y_{\text{init}}}
\left[
- \log p(y(\bm{x}) | y_{\phi,1:t}, Y_{\text{init}})
\right]\\
&= \mathbb{H}(y(\bm{x}) | y_{\phi,1:t}, Y_{\text{init}}).
\end{split}
\end{align}
Here, $y(\bm{x})$ is a random variable forecasting into the next query, while $\hat{y}(\bm{x})$ is an estimated value.
In a conventional scenario where the acquisition function measures the unqueried future point, one may take the GP predictive mean for $\hat{y}(\bm{x})$ resulting in $a_{\mathcal{H}}(\bm{x} | y_{\phi,1:t}, Y_{\text{init}})
=-1/2 \log e + \mathbb{H}(y(\bm{x}) | y_{\phi,1:t}, Y_{\text{init}})
\propto \mathbb{H}(y(\bm{x}) | y_{\phi,1:t}, Y_{\text{init}})$~\crefp{eqS-gp_posterior,eqS-gp_posterior_log_prob,eqS-gp_posterior_entropy}.
In our simulation, $\hat{y}(\bm{x})$ would be the simulated noisy realization on the entire domain (not just at $\bm{x}_{\phi, t}$ but all $\bm{x} \in \mathcal{X}$) where $\hat{y}(\bm{x}_{\phi,t+1})=y_{\phi,t+1}$.

$\mathbb{H}(y(\bm{x}) | y_{\phi,1:t}, Y_{\text{init}})$ is the standard predictive entropy~\citep{Seo2000GaussianPR,krause_nonmyopic_al_07}.

We refer the readers to~\citet{krause_nonmyopic_al_07,krause08a} for detailed discussions of conventional GP AL methods.
In these papers, the authors discuss sequential step-wise querying and joint batch querying.
Our training scheme can be considered as a special case where the policy performs sequential querying while the objectives compute joint batch querying but with true evaluations.

\subsubsection{Objectives: Unconstrained AL - Regularized Entropy}
Here, we discuss empirical issues of unconstrained entropy and our approach using mutual information objectives~\citep{guestrin_mi_al_05,krause08a}.

Maximizing an entropy objective favors a set of distinctive points, which naturally encourages points at the border, as they are the most scattered.
\citet{guestrin_mi_al_05} argued that exploring the border may be suboptimal because we would rather explore inside the space.
The authors further proposed a mutual information acquisition criterion to tackle this problem, at least in conventional AL settings.
In our training scheme, the objectives $\mathcal{H}(\phi)$ and $\mathcal{H}_{\text{mean}}(\phi)$ sometimes overemphasize the border and completely ignore the inner region of $\mathcal{X}$, which is clearly not desired.
Part of the reasons can be the tanh layer in our NN~\crefp{figureS-nn}, which, although helping us bound the NN outputs in $\mathcal{X}$, has vanishing gradients at the boundary.
To tackle this issue, we wish to derive a mutual information objective suitable for our framework.

In active GP learning,~\citet{guestrin_mi_al_05} define a mutual information criterion as the reduction of entropy in an unexplored region:
$\mathbb{H}(y(\bm{x}_{\phi,1}), ..., y(\bm{x}_{\phi,T}) | Y_{\text{init}})
- \mathbb{H}( y(\bm{x}_{\phi,1}), ..., y(\bm{x}_{\phi,T}) | Y_{\text{init}}, \hat{y}(\mathcal{X} \setminus \bm{X}_\phi) )$, $\hat{y}(\mathcal{X} \setminus \bm{X}_\phi)$ means the output estimates corresponding to $\mathcal{X} \setminus \{ \bm{x}_{\phi,1}, ..., \bm{x}_{\phi,T} \}$.
The original paper considered small discrete $\mathcal{X}$ where $\hat{y}(\mathcal{X} \setminus \{ \bm{x}_{\phi,1}, ..., \bm{x}_{\phi,T} \})$ is a computable set of variables.
In our framework, $\mathcal{X}$ is a continuous space, and thus this is not well-defined.
Even if $\mathcal{X}$ is discrete, conditioning on a large pool (fine discretization) is computationally heavy, i.e. GP cubic complexity $\mathcal{O}\left( |\mathcal{X}|^3 \right)$~\crefp{appendix-gp_details}.
A discrete pool also enforces a classifier-like policy $\phi$, as the policy needs to select points from a pool, which prohibits us from utilizing the current NN structure developed on top of~\citet{foster_dad_2021,ivanova_idad_2021}.

To derive mutual information objectives suitable for our learning framework, note that entropy reduction is a regularized entropy objective.
We propose a simple yet effective approach: compute the regularization term only on a sparse set of $N_{\text{grid}}$ samples $\left( \bm{X}_{\text{grid}}, Y_{\text{grid}} \right) \subseteq \mathcal{X} \times \mathcal{Y}$. 
Then we turn the acquisition functions proposed by~\citet{guestrin_mi_al_05,krause_nonmyopic_al_07} into the following training objectives
\begin{align}
\nonumber
\begin{split}
\mathcal{I}(\phi)=
&\mathbb{E}
\left[
\frac{
-\log p(y_{\phi,1:T_{\text{sim}}} | Y_{\text{init}})
+\log p( y_{\phi,1:T_{\text{sim}}} | Y_{\text{init}}, Y_{\text{grid}} )
}{
N_{\text{init}}+T_{\text{sim}}
}
\right]~\text{\crefp{eq-gp_logprob_reduction_objective,eq-gp_logprob_reduction_objective_inner}},
\end{split}\\
\label{eq-gp_entropy_reduction_objective}
\begin{split}
\mathcal{I}_{\text{mean}}(\phi)= &\mathbb{E} \left[
\frac{
\mathbb{H}(y(\bm{x}_{\phi,1:T_{\text{sim}}}) | Y_{\text{init}})
-\mathbb{H}( y(\bm{x}_{\phi,1:T_{\text{sim}}}) | Y_{\text{init}}, Y_{\text{grid}} )
}{
N_{\text{init}}+T_{\text{sim}}
}
\right].
\end{split}
\end{align}
The expectation is over GPs ($\theta, \sigma^2$) and AL functions ($f, T_{\text{sim}}, \mathcal{E}_{\text{init}}, \epsilon_{1:T_{\text{sim}}}$).
$N_{\text{grid}}$ should be much larger than $T$.
Maximizing these objectives encourages $\{ \bm{x}_{\phi,1},...,\bm{x}_{\phi,T} \}$ to track subsets of $\bm{X}_{\text{grid}}$.
The intuition is two-fold:
(i) we can view them as entropy objectives regularized by an additional search space indicator, or
(ii) we can view them as imitation objectives because a subset of grid points, if happens to have small joint likelihood or large joint entropy, maximizes the objective.

Note that, to keep the policy from overfitting those sparse grid samples, which are not necessarily optimal points, we re-sample $\bm{X}_{\text{grid}}$ in each training step.
We sample $\bm{X}_{\text{grid}}$ with $\text{Beta}(0.5, 0.5)$, but a uniform distribution can also be used and we did not see obvious difference.
Empirically, training with the two mutual information objectives are easier to converge than the entropy objectives.
Entropy objectives often stick in border-only patterns, particularly when $D \geq 2$.

One drawback of such objectives is that we lack insight into the appropriate number of the grid points.
When the dimension of $\mathcal{X}$ grows, $N_{\text{grid}}$ might need to be increased, which can make our mutual information objectives computationally impossible.
We provide our exact $N_{\text{grid}}$ in~\cref{tableS-batch_sizes}.

Our ablation study in~\cref{appendix-ablation} demonstrate the results of $\mathcal{I}_{\text{mean}}$.

Note that this regularization is meaningless to safe AL, because the border of $\mathcal{X}$ is typically less safe and the safe AL objectives already reflect this~\crefp{figureS-safe_acq}

\subsection{Objectives: DAD Baseline}\label{appendix-objective_details-DAD}

\begin{algorithm}[H]
\captionof{algorithm}{DAD Training with GP Prior}
\label{alg-dad_training}
\begin{algorithmic}[1]
\Require  \cref{assump-gp_prior}, $T$, $N_{\text{init}}$
\State draw a batch of $(\theta, \sigma^2)$
\For{each $(\theta, \sigma^2)$}
\State draw $f_{0,\text{raw}} \sim \mathcal{GP}(0, k_\theta)$
\State $f_{0}(\cdot)=f_{0,\text{raw}}(\cdot) - \mathbb{E}_{\bm{x}\in\mathcal{X}}[f_{0,\text{raw}}(\bm{x})]$
\State draw $\bm{X}_{\text{init}} \sim \text{Uniform}[\mathcal{X}], |\bm{X}_{\text{init}}|=N_{\text{init}}$
\State draw $\mathcal{E}_{\text{init}} \sim \mathcal{N}(0, \sigma^2 I), Y_{\text{init}}=f_{0}(\bm{X}_{\text{init}})+\mathcal{E}_{\text{init}}$
\State $\mathcal{D}_0 = \{ \bm{X}_{\text{init}}, Y_{\text{init}} \}$
\For{$t=1, ..., T$}
\State $\bm{x}_{\phi,t} = \phi(\mathcal{D}_{t-1})$
\State draw $\epsilon_{t} \sim \mathcal{N}(0, \sigma^2), y_{\phi,t}=f_{0}(\bm{x}_{\phi,t}) + \epsilon_t$
\State $\mathcal{D}_{t} \gets \mathcal{D}_{t-1} \cup \{ \bm{x}_{\phi,t}, y_{\phi,t} \}$
\EndFor
\State draw $f_{l,\text{raw}} \sim \mathcal{GP}(0, k_\theta), l=1,...,N_{f,q}$
\State $f_{l}(\cdot)=f_{l,\text{raw}}(\cdot) - \mathbb{E}_{\bm{x}\in\mathcal{X}}[f_{l,\text{raw}}(\bm{x})], l=1,...,N_{f,q}$
\EndFor
\State compute DAD loss~\crefp{eq-dad_objective}, update $\phi$
\end{algorithmic}
\end{algorithm}

We describe the DAD baseline in our GP context~\citep{foster_dad_2021}.
DAD is a scheme which samples tasks and learns to perform Bayesian optimal experimental design originally aiming at parametric models.
Bayesian optimal experimental design aims to query data to learn about a posterior Bayesian model~\citep{Rainforth2024_bed_overview}.
In a GP learning problem, this means collecting $\bm{x}_{1:T}, y_{1:T}$ to make inference $p(f(\cdot) | \bm{X}_{\text{init}}, Y_{\text{init}}, \bm{x}_{1:T}, y_{1:T})$.
DAD follows similar training procedure:
we (i) sample tasks from a prior, and (ii) learn a data querying policy by optimizing the DAD objective.
To train with DAD, one may use our main training algorithm because we generate more quantities than needed by the DAD objective.
For clarity, however, we write down a clean version of DAD training in~\cref{alg-dad_training}, which is the Algorithm 1 of~\citet{foster_dad_2021} combined with our GP prior.
We summarize the comparison between DAD and our main safe AL training~\crefp{alg-asal_training}:
(i) safety measurements are not present in DAD, (ii) the policy of DAD does not take the budget variable while $T_{\text{sim}}=T$ is fixed during the training, and (iii) functions $f_0, ..., f_{N_{f,q}}$ are sampled but the DAD objective simulates the observations $Y_{\text{init}}, y_{1:T}$ on $f_0$ only while $f_1, ..., f_{N_{f,q}}$ are contrastive samples for noise contrastive estimations.
Note that in this paper, $N_{f,q}$ denotes the number of functions $|\{(f,q)\}|$ per set of GP hyperparameters.
Here, we use $N_{f,q}$ to denote the number of contrastive functions.

The objective is
\begin{align}\label{eq-dad_objective}
DAD(\phi)=
\mathbb{E}_{\theta, \sigma^2}\left[
\mathbb{E}_{f_0, ...,f_{N_{f,q}}, \mathcal{E}_{\text{init}}, \epsilon_{1:T}}\left[
\log\frac{
    p(Y_{\text{init}}, y_{\phi,1:T} | f_0)
}{
    1/(N_{f,q}+1)\sum_{l=0}^{N_{f,q}}p(Y_{\text{init}}, y_{\phi,1:T} | f_l)
} 
\right]
\right].
\end{align}

This objective is meant to be maximized.

This objective was derived for parametric models, i.e. $p(\text{model parameters} | \text{data})$ in contrast to our $p(f | \text{data})$. 
Please see~\citet{foster_dad_2021} for details.

Notice that a further work,~\citet{ivanova_idad_2021}, extended DAD to more general scenarios, for example, allowing intractable $p(y | f)$.
However, their proposed learning objectives require another NN mapping, $\{(\mathcal{D}, f)\} \rightarrow \mathbb{R}$ if described in our notation, to help estimate the involved intractable distributions.
Such a mapping is not applicable for nonparametric $f$, and the objectives in~\citet{ivanova_idad_2021} were again not designed for AL of nonparametric functions.
This is a different direction to what we need in this paper.

\subsection{Objectives: Safe AL}\label{appendix-objective_details-safe_objectives}

\begin{figure}[t]
	\vskip 0.2in
	\begin{center}
		\centerline{\includegraphics[width=0.8\linewidth]{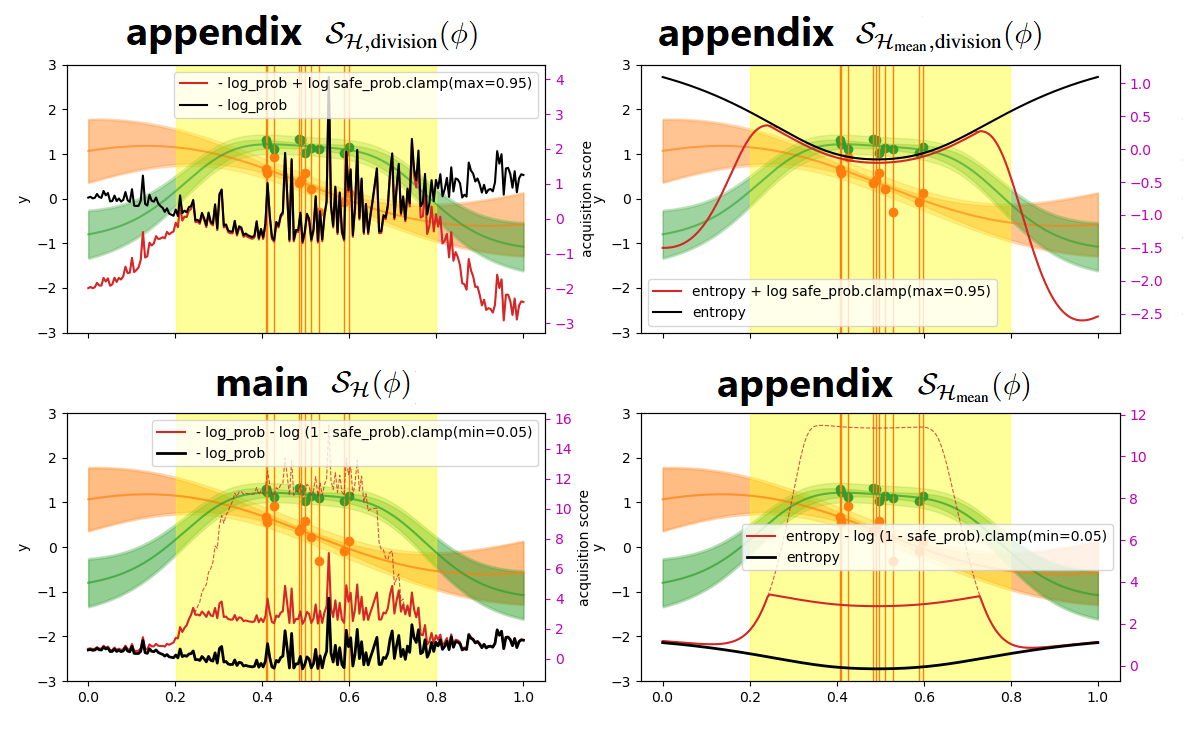}}
		\caption{
\textbf{Simulated safe AL objectives.}
$T_{\text{sim}}=1$.
As described in~\cref{figureS-acq_safeprob}, orange and green are $y,z$ and the corresponding GP predictions, while black curves are unconstrained objectives $\mathcal{H}, \mathcal{H}_{\text{mean}}$.
Our safe AL objectives (red) decorate the unconstrained objectives with our main min unsafe log probability (bottom, $\mathcal{S}_{\cdot}$,~\cref{eq-safe_gp_logprob_objective}) or with the appendix max safe log probability (top, $\mathcal{S}_{\cdot, \text{division}}$,~\cref{eq-safe_gp_logprob_division_objective}).
The objectives illustrated are meant to be maximized.
Red dashed lines show the objectives if the safety probability is not clamped by $\gamma$.
Note again that $y,z$ are available because our training scheme simulates AL before utilizing the evaluations for objective computations.
}
		\label{figureS-safe_acq}
	\end{center}
	\vskip -0.2in
\end{figure}

We illustrate our main safe AL objective, and we introduce more objectives. 

In our main paper, we propose a safe AL objective~\crefp{eq-safe_gp_logprob_objective}
\begin{align*}
\mathcal{S_{H}}(\phi)
&=
\mathbb{E}
\left[
\frac{
-\log p(y_{\phi,1}, ..., y_{\phi,T_{\text{sim}}} | Y_{\text{init}})
-\sum_{t=0}^{T_{\text{sim}}-1} \log \ \text{max}(\gamma, p(z(\bm{x}_{\phi,t+1}) < 0 | z_{\phi, 1:t}, Z_{\text{init}}) )
}{N_{\text{init}}+T_{\text{sim}}}
\right].
\end{align*}
This objective is a combination of an unconstrained exploration objective and a safety regularization.
We are able to exchange the exploration term $-\log p(y_{\phi,1}, ..., y_{\phi,T_{\text{sim}}} | Y_{\text{init}})$ by any other objectives, e.g. $\mathcal{S}_{\mathcal{H}_{\text{mean}}}(\phi)$ corresponds to safety decorated $\mathcal{H}_{\text{mean}}(\phi)$.

If we take a closer look into the safety term, the objective is maximized when $\log p(z(\bm{x}_{\phi,t+1}) < 0 | z_{\phi, 1:t}, Z_{\text{init}})$ is small.
Ideally, the probability of being unsafe should be small (or equivalently the probability of being safe is large,~\cref{figureS-acq_safeprob}).
Minimizing the unsafe probability nevertheless makes the $\log$ value explode to negative infinity (if not clamped by a non-zero $\gamma$).
Numerically, we suggest to add a small number to the probability to stabilize the computations; e.g., we add $10^{-5}$ for clamped and non-clamped versions for stability.

Nevertheless, another approach we consider is to change the safety term, so that we directly maximize the safety probability
\begin{align*}
\begin{split}
&-\log p(y_{\phi,1}, ..., y_{\phi,T_{\text{sim}}} | Y_{\text{init}})
+\sum_{t=0}^{T_{\text{sim}}-1} \log p(z(\bm{x}_{\phi,t+1}) \geq 0 | z_{\phi, 1:t}, Z_{\text{init}})\\
=
&-\sum_{t=0}^{T_{\text{sim}}-1}
\log\frac{
p(y_{\phi,t+1} | y_{\phi,1:t}, Y_{\text{init}})
}{
p(z(\bm{x}_{\phi,t+1}) \geq 0 | z_{\phi, 1:t}, Z_{\text{init}})
},
\end{split}
\end{align*}
which corresponds to
\begin{align}\label{eq-safe_gp_logprob_division_objective}
\mathcal{S}_{\mathcal{H},\text{division}}(\phi)
&=
\mathbb{E}
\left[
\frac{
-\log p(y_{\phi,1}, ..., y_{\phi,T_{\text{sim}}} | Y_{\text{init}})
+\sum_{t=0}^{T_{\text{sim}}-1} \log p(z(\bm{x}_{\phi,t+1}) \geq 0 | z_{\phi, 1:t}, Z_{\text{init}})
}{N_{\text{init}}+T_{\text{sim}}}
\right].
\end{align}
Note that $z(\bm{x}_{\phi,t+1})$ is a prediction, not the evaluated $z_{\phi,t+1}$.
Similarly, we may exchange the base exploration term to get e.g. $\mathcal{S}_{\mathcal{H}_{\text{mean}},\text{division}}(\phi)$. 
With this objective, we do not clamp the safety probability with $\gamma$ because this has no visible effect (see~\cref{figureS-safe_acq}).
The reason is, if we would clamp the likelihood, i.e. $\text{min}(1-\gamma, p(z(\bm{x}_{\phi,t+1}) \geq 0 | z_{\phi, 1:t}, Z_{\text{init}})$ (maximizing the safety likelihood until $1-\gamma$), then the clamped term is optimized towards $p(z(\bm{x}_{\phi,t+1}) \geq 0 | z_{\phi, 1:t}, Z_{\text{init}}) \geq 1-\gamma$ but for small $\gamma$, $\log (1-\gamma) \approx \log 1$.

Please see~\cref{figureS-safe_acq} for an illustration.
Our main $\mathcal{S}_{\cdot}$ is more conservative at the safe set border while the appendix $\mathcal{S}_{\cdot, \text{division}}$ is more prone to space exploration.

In this paper, we avoid coupling $\mathcal{I}, \mathcal{I}_{\text{mean}}$ with our safety terms because
(i) it is rather tricky to sample $\{ \bm{X}_{\text{grid}}, Y_{\text{grid}}, Z_{\text{grid}} \}$~\crefp{eq-gp_entropy_reduction_objective,eq-gp_logprob_reduction_objective} when a safety constraint is present,
(ii) we do not see an obvious intuition behind such safe AL objectives,
and (iii) we do not see empirical benefit in our preliminary experiments (not shown in the paper).

We illustrate the safe AL objectives in~\cref{figureS-safe_acq} for $T_{\text{sim}}=1$.

\section{Training: GP Function Sampling}\label{appendix-training_samplers}

This section outlines the mathematical details of our GP function sampling, i.e.~\cref{alg-initial_sample}.

We first introduce our GP kernel.
In our paper, we always use an RBF kernel, including $k_{\theta}, k_{q}$ in~\cref{alg-initial_sample} and the GP kernel in benchmark experiments.
An RBF kernel has $D+1$ variables: the variance $v$ and a $D$ dimension lengthscale vector $\bm{l}=(l_1, ..., l_D)$
\begin{align}\label{eqS-rbf_kernel}
k(\bm{x}, \bm{x}') = 
v \exp
\left(
-1/2 \sum_{d=1}^{D}
\left(
\frac{
[\bm{x} - \bm{x}']_d
}{
l_d
}
\right)^2
\right).
\end{align}

\subsection{Fourier Feature Functions}\label{appendix-training_samplers-fff}

In~\cref{alg-initial_sample}, the GP functions $f_{\text{raw}}\sim \mathcal{GP}(0, k_{\theta}), q_{\text{raw}} \sim \mathcal{GP}(0, k_{q})$ are approximated by Fourier features, which means each function sample is a linear combination of cosine functions~\citep{Rahimi2007_rff,Wilson2020_icml_rff_gp_post}:
\begin{align*}
\text{e.g. } f_{\text{raw}}(\bm{x}) = \sum_{i=1}^{L} \omega_{i} \sqrt{2/L} \cos\left( \bm{a}_i^T\bm{x} + b_i \right),
\end{align*}
$\omega_i \sim \mathcal{N}(0, \sqrt{v}^2), \bm{a}_i \sim \mathcal{N}(0, \text{diag}\{\bm{l}\}^{-1})$ are kernel dependent and $b_i\sim \text{Uniform}(0, 2\pi)$.
Each function has $L*(D+2)$ parameters.
Larger $L$ lead to better approximations.
An error bound w.r.t. $D$ and $L$ can be seen in~\citet{Rahimi2007_rff}.
We set $L=100$.

\begin{remark}[running out of symbols]
We highlight that the parameters of our Fourier feature functions, $\omega_i, \bm{a}_i$ and $b_i$, are independent of our acquisition function notation $a(\cdot)$ or other constants.
The Fourier feature function parameters (except for $L$, the number of features) are used only for this subsection and will not be referred in any other part of the paper.
\end{remark}

In our main paper, the analytical mean of window $\mathcal{X}=[0, 1]^D$ is computed such that all functions can be shifted to zero mean in the particular domain. 
The analytical mean is the integral of $f_{\text{raw}}(\bm{x}), q_{\text{raw}}(\bm{x})$ divided by volume of $[0, 1]^D$.
We give examples for the one and two dimensional cases:
\begin{align*}
\sum_{i=1}^{L} \frac{1}{a_i} \omega_{i} \sqrt{2/L} &\sin\left( a_i x + b_i \right) \vert_{x=0}^1 \text{, for } D=1,\\
\sum_{i=1}^{L} \frac{-1}{[\bm{a}_i]_1[\bm{a}_i]_2} \omega_{i} \sqrt{2/L} &\cos\left( [\bm{a}_i]_1 x_1 + [\bm{a}_i]_2 x_2 + b_i \right) \vert_{x_1=0}^1\vert_{x_2=0}^1 \text{, for } D=2,\\
&\text{ higher dimensions are analogous.}
\end{align*}
It may happen that at least one component of $\bm{a}_i$ is zero or is close to zero, which causes a problem in the division.
In this case, we replace $\vert 1 / \prod_{d=1}^D [\bm{a}_i]_d \vert$ by $100000$.
The error is negligible, i.e. much smaller than noise level.
We can replace $[0, 1]^D$ by any windows of our interest.

The complexity of computing such mean is $\mathcal{O}\left( L2^D \right)$, and this can be distributed to time or space.
With common GP and RBF kernel dimensions, e.g. $D \leq 5$ or $10$, this complexity is negligible.

\subsection{Prior Mean of Safety Functions}\label{appendix-training_samplers-safety_prior_mean}

\begin{figure}[t]
\vskip 0.2in
\begin{center}
\centerline{\includegraphics[width=\linewidth]{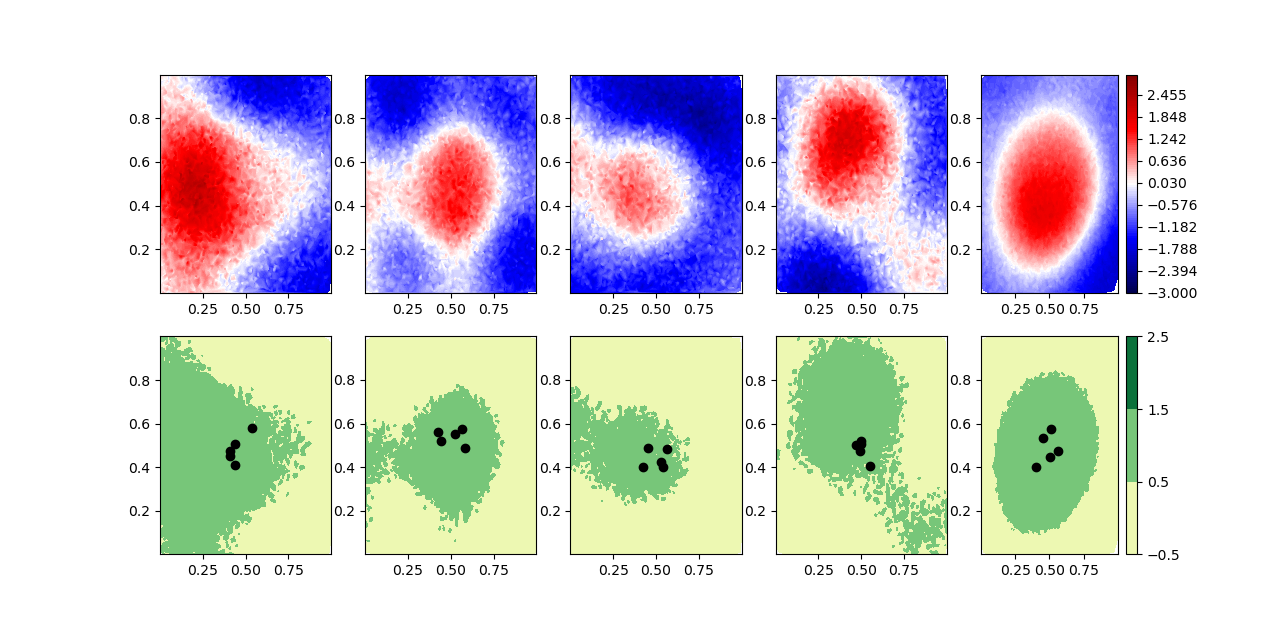}
}
\caption{
\textbf{Examples of sampled} $q \sim \mathcal{GP}\left( \mu_{q}, k_{q} \right)$.
We run~\cref{alg-initial_sample} to sample $f$ (not shown) and $q$.
The observations are blurred with noises.
The top are five samples of $z\left( [0, 1]^2 \right)$, the bottom safe and unsafe classifications $z\left( [0, 1]^2 \right) \geq 0$.
Black dots are the sampled initial data.
The parameters of $\mu_q, k_{q}$ leverage samplers listed in~\cref{tableS-training_samplers}.
Our algorithm samples diverse safety patterns while guaranteeing stable initial safe data sampling.
}
\label{figureS-q_examples}
\end{center}
\vskip -0.2in
\end{figure}

As described in our main~\cref{section-method-function_sampling}, we use a GP mean function $\mu_{q}$ to help us control the safety function sampling.

The overall goal is to sample functions $q \sim \mathcal{GP}_{\theta_{q}}(\mu_{q}, k_{q})$, while guaranteeing a high probability of the existence of central safe data.
We design $\mu_q$ such that it is safe at the center of the domain $\mathcal{X}$.
We use the hyperbolic secant function
\begin{align}\label{eqS-sech_mean}
\begin{split}
\mu_{q}(\bm{x}) &=
3.2c \left(
-0.47 + \text{sech}\left(
\frac{1}{D}
\sum_{d=1}^{D}
w_d [\bm{u}]_d^2
\right)
\right),\\
\bm{u} &=
Q^T \left(
\bm{x} - 
\begin{pmatrix}
    0.5 \\ \vdots \\ 0.5
\end{pmatrix}
\right),
\end{split}
\end{align}
$Q \in \mathbb{R}^{D \times D}$ is an orthogonal matrix where columns of $Q$ are orthonormal, each $w_d > 0$ is a shape parameter.
We provide our design logic and the chosen parameters of this function step by step:
\begin{enumerate}
    \item Consider $Q=I_{D}$, then $\frac{1}{D} \sum_{d=1}^{D} w_d [\bm{u}]_d^2 = \frac{1}{D} \sum_{d=1}^{D} w_d ([\bm{x}]_d - 0.5)^2$ is an ellipsoid centering around $(0.5, ..., 0.5)\in \mathbb{R}^D$.
    \item We can see that $\mu_q$ has the center area being a safe ellipsoid as long as $c>0$, with shape and size controlled by $w_d$, and the orthogonal matrix $Q$ allows us to rotate the ellipsoid around the center $(0.5, ..., 0.5)$.
    The orthogonal matrix $Q$ is obtained by performing a QR-decomposition of a sampled $A \in \mathbb{R}^{D \times D}$ (each entity from $\text{Uniform}[-1, 1]$).
    \item The above steps describe variables $w_d, c, Q$, and we then describe the constants.
    \item If we consider $c=1, w_d/D=10, \forall d \leq D$ (e.g. $w_d = 20, D=2$) and $Q=I_D$, then the central safe area is a ball and it takes about half of the space, i.e. the mean function $\mu_{q}$ brings half of the space safe and half unsafe.
    We will later sample the shape and the half-safe space is only for an initial design.
    \item With the same $c, w_d, Q$, the constants $3.2$ and $-0.47$ ensure zero mean and unit variance of this $\mu_q$ function, which aligns with our setup that the deployment problems are normalized, and this provides us an estimated variance of $\mu_{q} \approx c^2$.
\end{enumerate}

When we sample $q=\mu_{q} + q_{\text{raw}} - \mathbb{E}_{\bm{x}\in\mathcal{X}}[q_{\text{raw}}(\bm{x})], q_{\text{raw}}\sim \mathcal{GP}(0,k_{q})$ (see~\cref{alg-initial_sample}), we aim for mean $\mathbb{E}[q]$ around $0$ and variance of $q$ around $1$, as our deployment problems.
We distribute the variance by making sure $c^2 + v + \sigma_q^2$ is around $1$ (kernel variance $v$, observation noise level $\sigma_q$), in particular, we set $c^2 = 0.5$ while the kernel and the noise take the remaining half.
The kernel and the noise are the main sources of functional stochasticity.
The assumption of having additive variances is in fact not true for our $\mu_{q}$, but this setting is enough for the NN to learn.
One may also consider the overall function variance as an amplitude of the function.
The shape parameters $\bm{w}=(w_1,...,w_D)$ are sampled around value $20$, such that $\mu_{q}$ has around 10\% to 100\% of the space above safety threshold $0$. 
We will summarize the sampling setting later in~\cref{appendix-experiment_details}.

We illustrate examples of sampled $q$ in~\cref{figureS-q_examples}.

\section{Training Complexity}\label{appendix-training_complexity}

\begin{table}[t]
\scriptsize
\begin{center}
\caption{
\textbf{Batch sizes in training.} 
}\label{tableS-batch_sizes}
\begin{tabular}{c|cc|c}
\toprule
\textbf{loss functions}
&$\mathcal{I}$
&DAD &$\mathcal{S}_{\mathcal{H}}$\\
\midrule
$N_k=|\{ (\theta, \theta_{q}) \}|$
& 10 & 10 & 10\\
$N_{f,q}=|\{ (f, q) \}|$
& 5 & 200 & 5\\
$B=|\{ (\mathcal{E}_{\text{init}}, \epsilon_{1:T}, \mathcal{E}_{q, \text{init}}, \epsilon_{q,1:T}) \}|$
& 10 & 10 & 1\\
$L=$num of fourier features & 100 & 100 & 100\\
$N_{\text{grid}}$ (number of $\bm{X}_{\text{grid}}$)
& 100 (for NN $D\leq 2$) & N/A & N/A\\
& 500 (for NN $D =5$) & & \\
\bottomrule
\end{tabular}
\end{center}
\end{table}

\begin{table}[t]
\scriptsize
\begin{center}
\caption{
\textbf{Loss function complexities.}
The batch sizes $B, N_{k}, N_{f,q}$ contribute linearly in time or space (but not both), and this is up to the implementation.
We derive the worst case $T_{\text{sim}}=T$.
}
\label{tableS-training_complexities}
\begin{tabular}{c|cc}
\toprule
& time & space \\
\midrule
$p( \cdot | Y_{\text{init}})$ (\cref{eqS-gp_posterior}, plug $y_{\phi,1:T}$ in later) & {\color{blue}$\mathcal{O}\left( N_{\text{init}}^3 \right)$} & {\color{blue}$\mathcal{O}\left( N_{\text{init}}^2 \right)$}
\\
$p(\cdot | Y_{\text{init}}, Y_{\text{grid}})$~\crefp{eqS-gp_posterior} & {\color{teal}$\mathcal{O}\left( (N_{\text{init}} + N_{\text{grid}})^3 \right)$} & {\color{teal}$\mathcal{O}\left( (N_{\text{init}} + N_{\text{grid}})^2 \right)$}
\\
$\log p(z(\bm{x}_{\phi,T}) < 0 | z_{\phi,1:T-1}, Z_{\text{init}})$ & {\color{purple}$\mathcal{O}\left( (N_{\text{init}} + T)^3 \right)$} & {\color{purple}$\mathcal{O}\left( (N_{\text{init}} + T)^2 \right)$}
\\
\crefp{eqS-gp_posterior,eqS-gp_posterior_cdf} & & \\
\hline
$\mathcal{H}$~\crefp{eq-gp_logprob_objective,eq-gp_logprob_objective_inner} &  {\color{blue}$\mathcal{O}\left( N_{\text{init}}^3 \right)$} $+$ $\mathcal{O}\left( T^3 \right)$ & {\color{blue}$\mathcal{O}\left( N_{\text{init}}^2 \right)$} $+$ $\mathcal{O}\left( T^2 \right)$
\\
\hline
$\mathcal{I}$~\crefp{eq-gp_logprob_reduction_objective,eq-gp_logprob_reduction_objective_inner} &  {\color{teal}$\mathcal{O}\left( (N_{\text{init}} + N_{\text{grid}})^3 \right)$} $+$ $\mathcal{O}\left( T^3 \right)$ & {\color{teal}$\mathcal{O}\left( (N_{\text{init}} + N_{\text{grid}})^2 \right)$} $+$ $\mathcal{O}\left( T^2 \right)$
\\
\hline
$\mathcal{S}_{\mathcal{H}}$~\crefp{eq-safe_gp_logprob_objective} & {\color{blue}$\mathcal{O}\left( N_{\text{init}}^3 \right)$} $+$ $\mathcal{O}\left( T^3 \right)$ & {\color{blue}$\mathcal{O}\left( N_{\text{init}}^2 \right)$} $+$ $\mathcal{O}\left( T^2 \right)$
\\ & $+$ {\color{purple}$\mathcal{O}\left( (N_{\text{init}} + T)^3 \right)$} & $+$ {\color{purple}$\mathcal{O}\left( (N_{\text{init}} + T)^2 \right)$}
\\
\hline 
DAD baseline~\crefp{eq-dad_objective} & \multicolumn{2}{c}{$\mathcal{O}\left( N_{\text{init}}+T \right)$ in time or space} 
\\
\bottomrule
\end{tabular}
\end{center}
\end{table}

Our loss functions take expectation over GP hyperparameters and functions.
We summarize the batch sizes in~\cref{tableS-batch_sizes}.
Note that for each set of GP hyperparameters, the output realizations consist of noise-free $f, q$ and noise realizations, denoted individually.
The number of noise realizations, $B$, can be considered as the repetitions of simulated safe AL per function pair $(f,q)$.

The training complexities are dominated by the NN forward passes and the loss computations.

The NN forward passes takes $\mathcal{O}\left( \sum_{t=1}^{T}(N_{\text{init}}+ t - 1)^2 \right)$ in time, as self attention has square complexity (\cref{figureS-nn},~\cite{Vaswani2017_attention}).
The space complexity depends on the number of NN parameters.

\cref{tableS-training_complexities} summarizes the complexities of computing our loss functions.
Note that the conditional probability are expressed by the posterior Gaussian mean and covariance~\crefp{eqS-gp_posterior}.
Then we use the posterior to compute the log likelihood.
We use different colors for the GP posteriors to indicate the sources of the complexities.
Our appendix objectives ($\mathcal{H}_{\text{mean}}, \mathcal{I}_{\text{mean}}$) have the same complexities as the main objectives ($\mathcal{H}, \mathcal{I}$).

The safe AL objective $\mathcal{S}_{\mathcal{H}}$ is a combination of $\mathcal{H}$ and safety score.
Note that $\log p(z(\bm{x}_{\phi,t+1}) < 0 | z_{\phi,1:t}, Z_{\text{init}})$ are all intermediate results of $\log p(z(\bm{x}_{\phi,T}) < 0 | z_{\phi,1:T-1}, Z_{\text{init}})$, which creates negligible additional complexities.
The appendix safety score has the same complexity as the main, i.e. $\mathcal{S}_{\mathcal{H},\text{division}}$ and $\mathcal{S}_{\mathcal{H}}$ have the same complexities.

\section{Experiment Details}\label{appendix-experiment_details}

\subsection{Offline Training \& Training Time}\label{appendix-experiment_details-training_numerical}

\begin{table}[t]
\scriptsize
\begin{center}
\caption{\textbf{Training samplers.}}\label{tableS-training_samplers}
\begin{tabular}{c|l}
\toprule
\textbf{samples}&\textbf{distributions}\\
\midrule
$f \sim \mathcal{GP}(0, k_{\theta})$
& 
$k_{\theta}:$ RBF kernel,~\cref{eqS-rbf_kernel} (parameters $\theta=(v, \bm{l})$, $\bm{l}=(l_1,...,l_D)$)
\\ \cline{2-2} 
& 
$v\sim \text{Uniform}[0.9616, 1.0]$
\\ & 
$\forall d \leq D, (l_{d}-0.2)\sim \text{Gamma}(shape=1, rate=10)$
\\ \cline{2-2} 
$\epsilon \sim \mathcal{N}(0, \sigma^2)$
& 
$\sigma^2 = 1.0001 - v$, i.e. $0.01 \leq \sigma \leq 0.2$, signal-to-noise ratio $ \frac{\sqrt{v}}{\sigma} \geq 5$
\\ \hline
$q \sim \mathcal{GP}_{\theta_{q}}(\mu_{q}, k_{q})$
& 
$\mu_{q}$: sech prior mean,~\cref{eqS-sech_mean} (parameters $c, \bm{w}, Q$, $\bm{w}=(w_1,...,w_D)$)
\\ 
$\theta_{q}=(c, \bm{w}, Q, v, \bm{l})$
&
$k_{q}$: RBF kernel,~\cref{eqS-rbf_kernel} (parameters $v, \bm{l}$, $\bm{l}=(l_1,...,l_D)$)
\\ \cline{2-2} 
& 
$Q$: QR-decomposition of $A$, $[A]_{i,j} \sim \text{Uniform}[-1, 1], \forall i,j = 1,...,D$
\\ & 
$\forall d \leq D, w_d \sim \text{Uniform}[5, 40]$
\\ & 
$c=\sqrt{0.5}$
\\ & 
$v\sim \text{Uniform}[0.9616-c^2, 1.0-c^2]$
\\ & 
$\forall d \leq D, (l_{d}-0.2)\sim \text{Gamma}(shape=1, rate=10)$
\\ \cline{2-2} 
$\epsilon_{q} \sim \mathcal{N}(0, \sigma^2_{q})$
& $\sigma_{q}^2 = 1.0001 - c^2 - v$, i.e. $0.01 \leq \sigma_{q} \leq 0.2$, $\frac{\sqrt{c^2 + v}}{\sigma_{q}} \geq 5$
\\ \hline
$\mathcal{I}$~\crefp{eq-gp_logprob_reduction_objective,eq-gp_logprob_objective_inner}
& $\bm{X}_{\text{grid}} \sim \text{Beta}(0.5, 0.5)$, $Y_{\text{grid}}=f(\bm{X}_{\text{grid}}) + \text{noise}, \text{noise} \sim \mathcal{N}(0, \sigma^2)$
\\ \hline
Fluid System
& kernels: $\bm{l}$ of $k_{\theta}, k_{q}$ sampled as~\cref{eqS-fluid_system_prior}\\
& safety mean shape: $\bm{w}$ sampled as~\cref{eqS-fluid_system_prior}
\\
\bottomrule
\end{tabular}
\end{center}
\end{table}

\begin{table}[t]
\scriptsize
\begin{center}
\caption{
\textbf{Policy training time.} All training jobs were on a single Intel i5-12600K CPU and an NVIDIA RTX 3080 GPU.
A combined comparison of training and runtime against baselines is provided in our ablation study~\crefp{appendix-ablation,tableS-trained_run_thresholds}.
}\label{tableS-training_time}
\begin{tabular}{l|ccc|c}
\textbf{loss functions}
&$\mathcal{I}$
&DAD baseline
&ALINE baseline
&$\mathcal{S}_{\mathcal{H}}$
\\
training steps
&$10k$
&$20k$
&$40k$
&$10k$
\\ \midrule
NN 1D, 2D
& $\sim$ 1 hours & & & \\
$N_{\text{init}}=1, T_{\text{sim}}\leq T=30$
& ($N_{\text{grid}}=100$) & & &
\\ \hline
NN 5D
& $\sim$ 2 hours & & & \\
$N_{\text{init}}=20, T_{\text{sim}}\leq T=40$
& ($N_{\text{grid}}=500$) & & &
\\ \hline
NN 1D
& & $\sim$ [2, 2.5] hours & $\sim$ [6, 64] hours & \\
$N_{\text{init}}=1, T = [10, 20]$
& & & &
\\ \hline
NN 2D
& & $\sim$ [2, 2.5, 3.5] hours & $\sim$ [6, 77, 93.5] hours & \\
$N_{\text{init}}=1, T = [10, 20, 30]$
& & & &
\\ \hline
NN 5D
& & $\sim$ [3, 4, 6, 9] hours & $\sim$ [14, 72, 200, 350] hours & \\
$N_{\text{init}}= 20, T = [10, 20, 30, 40]$
& & \cref{appendix-experiment_details-training_numerical} explains why slow & &
\\ \hline
NN 2D
& & & & $\sim$ 2.5 hours \\
$N_{\text{init}}=5, T_{\text{sim}}\leq T=40$
& & & &
\\ \hline
NN 3D
& & & & $\sim$ 4 hours \\
$N_{\text{init}}=5, T_{\text{sim}}\leq T=60$
& & & &
\\ \hline
NN 7D
& & & & $\sim$ 12 hours \\
$N_{\text{init}}=10, T_{\text{sim}}\leq T=100$
& & & &
\\ \bottomrule
\end{tabular}
\end{center}
\end{table}

In our current implementation, the data dimension and input bound need to be pre-defined.
We fix $\mathcal{X}=[0, 1]^D$, and we map all test problems to this domain.
For each number of dimensions, $D$, we train one policy to deploy on various AL problems.
We summarize all of our sampling distributions in~\cref{tableS-training_samplers}.
For GP function sampling, see~\cref{appendix-training_samplers} for mathematical details.
Our numerical setting utilizes the assumption that deployment problems are normalized to zero mean and unit variance.
An implicit assumption we make here is that the functions do have information to be extracted, not just noises (so we set $\sigma \leq 0.2, \sigma_q \leq 0.2$).
The numerical setting can be adapted according to different applications.
Our Fluid System is a $D=7$ task, beyond typical deployment dimension of conventional methods ($D \geq 3$ or $4$ is challenging for safe AL/BO methods,~\citealt{Kirschner2019lineBO,alessandro2022safe}).
We tailor the kernel lengthscale and mean shape parameters to a broad range of plausible operating values:

\begin{align}
\begin{split}\label{eqS-fluid_system_prior}
    l_1 \sim \text{Uniform}\left[0.24, 0.72\right]&,
    w_1 \sim \text{Uniform}\left[10.42, 41.66\right],\\
    l_2 \sim \text{Uniform}\left[1.59, 4.77\right]&,
    w_2 \sim \text{Uniform}\left[1.57, 6.29\right],\\
    l_3 \sim \text{Uniform}\left[0.415, 1.245\right]&,
    w_3 \sim \text{Uniform}\left[6.02, 24.1\right],\\
    l_4 \sim \text{Uniform}\left[0.3, 0.9\right]&,
    w_4 \sim \text{Uniform}\left[8.33, 33.34\right],\\
    l_5 \sim \text{Uniform}\left[0.15, 0.45\right]&,
    w_5 \sim \text{Uniform}\left[16.66, 66.67\right],\\
    l_6 \sim \text{Uniform}\left[1.7, 5.1\right]&,
    w_6 \sim \text{Uniform}\left[1.47, 5.89\right],\\
    l_7 \sim \text{Uniform}\left[0.22, 0.66\right]&,
    w_7 \sim \text{Uniform}\left[11.36, 45.45\right].\\
\end{split}
\end{align}

The gradient of all our loss functions can be computed by PyTorch autograd.
This is because the GP i.i.d. noise assumption enables all our expectations to separate $f,q$ and noises, which means a derivative w.r.t. $y_{\phi,1:T}, z_{\phi,1:T}$ propagates automatically through $f, q$ to $\bm{x}_{\phi,1:T}$ which are direct outputs of the NN policies.

The batch sizes we set are listed in~\cref{tableS-batch_sizes}.
For each training pipeline, we train with a few different seeds.
The optimizer is RAdam~\citep{Liu2020radam}.
We set a lr scheduler to discount the lr by $2\%$ every $50$ training steps.
We usually train with $200*50=10000$ steps, but we give the DAD baseline more steps to converge (still not competitive on GP functions when $D \geq 2$).
The ALINE baseline utilizes even more steps as it learns posterior estimates and AL decisions together~\citep{huang2025aline}.
The exact traininig steps and times are summarized in~\cref{tableS-training_time}.
All training jobs were on a single Intel i5-12600K CPU and an NVIDIA RTX 3080 GPU.

\paragraph{NN Architectural Hyperparameters}

The NN configurations are taken largely from~\citet{ivanova_idad_2021}:
The transformer block is taken as it is;
the number of hidden neurons in the MLPs (Data Embed, Budget Feed Forward, and Decision Feed Forward) are all set to 512.
The remaining parameter is the embedding dimension (the dimension of $e_t, E_t, ES_t, EB$, and the input dimension of the Decision Feed Forward), which determines the size of the transformer layers and plays a crucial role in the NN size.
In our main paper, we set this dimension to 128, but we provide an ablation study on this dimension in~\cref{appendix-ablation}.

In our unconstrained AL experiments, our policy NNs were trained without safety measurements:
an NN does not have a safety history encoder~\crefp{figureS-nn}, and the training algorithm has neither safety measurements $q$ and $z$ nor safety center $\mathcal{C}$, i.e. the training~\cref{alg-asal_training} is reduced to~\cref{alg-aal_training}.

\paragraph{Training of Baseline Approaches}
For the DAD baseline (described in~\cref{appendix-objective_details-DAD},~\cref{alg-dad_training}), we take the code from~\citet{foster_dad_2021,ivanova_idad_2021}.
The DAD implementation utilizes Pyro~\citep{bingham2018pyro} to condition the same sequence of queries on different functions, which however involve CPUs for each loss computation.
In comparison, our loss functions are computed completely with PyTorch which requires little CPU-GPU interaction.
This is why the training time per step of DAD is not necessarily faster than our objectives~\crefp{tableS-training_time}, despite DAD time complexity much smaller~\crefp{tableS-training_complexities}.

For ALINE~\citep{huang2025aline}, we use the original code to train the method.
As ALINE optimizes the AL decisions for a predefined $T$, we train individual ALINE for different $T$.
Similar to our NN policies~\crefp{figureS-nn}, we use 2 layers of transformer attender.

For PFN baselines, we train PFN models~\citep{mueller2022pfn} on GP data~\crefp{tableS-training_samplers}.
The training was adapted from the implementation of~\cite{chang2025ace}.
PFNs are amortized Bayesian models which can condition on observed data and output predictive distributions of unseen data.
We use 2 layers of transformer attender to ensure the models have similar sizes (around 300K parameters) to our NN policies~\crefp{figureS-nn}.
Each PFN is limited to the specified dimension $D$; we train individual PFNs for each dimension of problems.
Each training uses 1.28M GP functions and around 3 hours (320k steps).

\subsection{Online Policy Deployment}\label{appendix-experiment_details-deployment_algs_ours}

We deploy our Amortized AL (AAL) policy with~\cref{algS-policy_al} and Amortized safe AL (ASAL) policy with~\cref{alg-policy_sal}.

\begin{algorithm}[H]
\captionof{algorithm}{Unconstrained AL with NN Policy}
\label{algS-policy_al}
\begin{algorithmic}[1]
\Require $\mathcal{D}_0\subseteq \mathcal{X}\times\mathcal{Y}$, AL policy $\phi$, $T$
\For{$t=1, ..., T$}
\State $\bm{x}_{t} = \phi(T-t+1, \mathcal{D}_{t-1})$ \textbf{if} $\phi$ is budget aware
\textbf{else} $\bm{x}_{t} = \phi(\mathcal{D}_{t-1})$
\State Query at $\bm{x}_{t}$ to get  $y_{t}$
\State $\mathcal{D}_{t} \gets \mathcal{D}_{t-1} \cup \{ \bm{x}_{t}, y_{t} \}$
\EndFor
\State \textbf{return} $\mathcal{D}_{T}$ suitable to model $f$
\end{algorithmic}
\end{algorithm}

\begin{algorithm}[H]
\captionof{algorithm}{Safe AL with NN Policy}
\label{alg-policy_sal}
\begin{algorithmic}[1]
\Require $\mathcal{D}_0\subseteq \mathcal{X}\times\mathcal{Y}\times\mathcal{Z}$, Safe AL policy $\phi$, $T$
\For{$t=1, ..., T$}
\State $\bm{x}_{t} = \phi(T-t+1, \mathcal{D}_{t-1})$
\State Query at $\bm{x}_{t}$ to get  $y_{t}, z_{t}$
\State $\mathcal{D}_{t} \gets \mathcal{D}_{t-1} \cup \{ \bm{x}_{t}, y_{t}, z_{t} \}$
\EndFor
\State \textbf{return} $\mathcal{D}_{T}$ suitable to model $f$
\end{algorithmic}
\end{algorithm}

\subsection{Baseline Methods}\label{appendix-experiment_details-deployment_algs_baselines}

The DAD baseline has to be trained up-front~\crefp{appendix-objective_details-DAD,appendix-experiment_details-training_numerical}.
In this case, we further take the budget encoder out of the NN structure~\crefp{figureS-nn}, as this is useless for DAD (and $T_{\text{sim}}=T$ is fixed).
The DAD policies are deployed with~\cref{algS-policy_al}.

The ALINE baseline is as it is in~\citet{huang2025aline}, except that (i) the input domain is adapted to our $\mathcal{X}$ and (ii) the GP hyperparameters for training are as~\cref{tableS-training_samplers}.

\cref{algS-convention_al} is the conventional unconstrained AL~\citep{settles2010_al,KumarGupta2020,tharwat2023_al}.


\begin{algorithm}[H]
\captionof{algorithm}{Conventional AL}
\label{algS-convention_al}
\begin{algorithmic}[1]
\Require $\mathcal{D}_0 \subseteq \mathcal{X}\times\mathcal{Y}$, acquisition function $a$, $T$
\For{$t=1, ..., T$}
\State Model $\mathcal{M}_{t-1}$ with $\mathcal{D}_{t-1}$
\State $\bm{x}_t = \text{argmax}_{\bm{x}\in\mathcal{X}} a(x | \mathcal{M}_{t-1}, \mathcal{D}_{t-1})$
\State Query at $\bm{x}_{t}$ to get  $y_{t}$
\State $\mathcal{D}_{t} \gets \mathcal{D}_{t-1} \cup \{ \bm{x}_t, y_t \}$
\EndFor
\State \textbf{return} $\mathcal{D}_{T}$
\end{algorithmic}
\end{algorithm}

Then we write down the safe AL~\citep{Schreiter2015,ZimmerNEURIPS2018_b197ffde,cyli2022}.
\begin{algorithm}[H]
\captionof{algorithm}{Conventional Safe AL}
\label{algS-convention_sal}
\begin{algorithmic}[1]
\Require $\mathcal{D}_0 \subseteq \mathcal{X}\times\mathcal{Y}\times\mathcal{Z}$, acquisition function $a$, safety threshold $0$, confidence tolerance $\gamma$, $T$
\For{$t=1, ..., T$}
\State Model $\mathcal{M}_{t-1}$, $\mathcal{M}_{\text{safety}, t-1}$ with $\mathcal{D}_{t-1}$
\State $\bm{x}_t = \text{argmax}_{\bm{x}\in\mathcal{X}} a(x | \mathcal{M}_{t-1}, \mathcal{D}_{t-1})$  subject to $p(z(\bm{x}) \geq 0 | \mathcal{M}_{\text{safety}, t-1}, \mathcal{D}_{t-1}) \geq 1-\gamma $
\State Query at $\bm{x}_{t}$ to get $y_t, z_t$
\State $\mathcal{D}_{t} \gets \mathcal{D}_{t-1} \cup \{ \bm{x}_t, y_t, z_t \}$
\EndFor
\State \textbf{return} $\mathcal{D}_{T}$
\end{algorithmic}
\end{algorithm}

We fix a safety tolerance of $\gamma = 0.05$, if not particularly described.

The base acquisition function $a$ is the predictive entropy $\mathbb{H}(y(\bm{x}) \vert \mathcal{D}_{t-1})$~\crefp{eqS-gp_posterior_entropy}.
The models $\mathcal{M}_{t-1}, \mathcal{M}_{\text{safety}, t-1}$ are GPs or PFNs in this paper.
The models are described in detail later in~\cref{appendix-experiment_details-gp_modeling}.

The benchmark problems consist of continuous functions and discrete datasets.

For the datasets, the (constrained) acquisition optimizations are solved by an exhaustive search, as done in safe learning literature~\citep{sui15safeopt,berkenkamp2020bayesian,cyli2022}.
In other words, we compute the acquisition scores and the safety distributions on the entire pool of unseen data, and we use the values to solve the (constrained) acquisition optimization.

For function problems, since a fine discretization with exhaustive search is computationally possible, we discretize the space densely by randomly sampling $5000$ input points.
Then we solve the (constrained) acquisition optimization problems as if these are pools.
Please be aware that the discretization is inherited from conventional safe learning methods~\citep{sui15safeopt,berkenkamp2020bayesian,cyli2022}) to solve the constrained acquisition optimization problem.
In the main paper, our policies propose queries on continuous space, which requires no discretization.

\subsection{Online Evaluations and Posterior Estimates}\label{appendix-experiment_details-gp_modeling}
In our experiments, we compute posteriors for either RMSE evaluations (using posterior means) or model-based AL and safe AL baselines.

\paragraph{GP Posteriors}
The first estimation approach is via GPs, leveraged in the following scenarios.
We always use a GP of zero mean and RBF kernel.
\begin{enumerate}
    \item Our \textbf{AAL}, our \textbf{ASAL}, \textbf{DAD}, Random: we deploy our amortized (safe) AL~\crefp{algS-policy_al,alg-policy_sal} or the DAD, Random baselines to collect data.
    After the specified $T$ data points are collected, we use the initial and queried data to fit a vanilla GP model with Type II maximum likelihood (optimization: L-BFGS-B algorithm).
    \item \textbf{GP AL}, \textbf{Safe GP AL}, \textbf{Safe Random}: We deploy conventional AL and safe AL~\crefp{algS-convention_al,algS-convention_sal} with vanilla GPs. Each iteration updates GPs with Type II maximum likelihood (optimization: L-BFGS-B algorithm).
    
    \item \textbf{AGP AL}, \textbf{Safe AGP AL}: We deploy conventional AL and safe AL~\crefp{algS-convention_al,algS-convention_sal} with AGPs. AGP is an amortized GP method developed by~\citet{liu2020amortized_gp,Bitzer2023amortized_gp} (AGP).
    Such a method sampled GP data and trained a transformer model to approximate the Type II maximum likelihood.
    The AGP is a model with a transformer module.
    Whenever an observation dataset is given, the transformer module infers the kernel structure (which we fix to an RBF) together with the kernel hyperparameters.
    Afterward, one can apply~\cref{eqS-gp_posterior} for computing the predictive distribution.
    \citet{Bitzer2023amortized_gp} provides a trained model attached to their code.
    
    \item \textbf{SVGP AL}, \textbf{Safe SVGP AL}: We deploy conventional AL and safe AL~\crefp{algS-convention_al,algS-convention_sal} with sparse variational GPs (SVGPs,~\citealt{hensman2013gaussian,hensman2015mcmc}).
    A SVGP uses a set of inducing variables (ivs, pseudo data points) to approximate the distribution conditioned on the full dataset. The number of ivs are smaller than the number of observed data, resulting in a cheaper decomposition of the gram matrix.
    The ivs are selected via k-means and the GP models are optimized via the L-BFGS-B algorithm, as done in the literature.
    We do not perform mini-batching (thus equivalent to~\citealt{pmlr-v5-titsias09a}).
    For Engine and Fluid System problems, we use 20 ivs, and 15 for all other problems.
    The ivs are fixed to the observed data if given less data than the desired number of ivs (i.e., a SVGP reduces to full GP with the variational inference objective).
    
    \item \textbf{MGP AL, Safe MGP AL}: We deploy conventional AL and safe AL~\crefp{algS-convention_al,algS-convention_sal} with mixture of GPs (MGPs,~\citealt{riis2023mgp_bal}), an ensembling GP method fine-tuning multiple vanilla GPs for a fixed number of steps, each with an individual set of initial model hyperparameters.
    We use 30 GPs and fine-tune each with Adam optimizer (learning rate $0.01$) for 30 steps~\citep{riis2023mgp_bal}.
\end{enumerate}

We want to point out that the cubic complexity is inevitable as long as we compute a GP distribution.
The difference of GP and AGP is that training a GP computes the gram matrix multiple times while AGP computes the gram matrix only once.
Our Amortized (safe) AL methods take GP computation completely out of the deployment cycle.

\paragraph{Other Posterior Estimates}
In addition, the following baselines have posterior estimates available as the queries are acquired.

\begin{enumerate}
    \item \textbf{ALINE}~\citep{huang2025aline}: the method uses NN forward passes for AL decisions and posterior modeling jointly. The posterior estimates predictive density conditioned on previous observations.

    \item \textbf{PFN/TabPFN AL, Safe PFN/TabPFN AL}: PFNs and TabPFN are amortized Bayesian models which can condition on observed data and output predictive distributions of unseen data. PFNs are trained on our GP data while TabPFN is a foundation model published by~\cite{hollmann2025tabpfn} (we use the TabPFNv2 regressor). The output posterior estimates can be used to compute the entropy~\citep{viering2025pfn} and safety likelihood, which are then used for (constrained) acquisition optimizations.
    In other words, we deploy~\cref{algS-convention_al,algS-convention_sal} by using a PFN or TabPFN model.

\end{enumerate}

\subsection{Deployment Hardware \& Complexities}\label{appendix-experiment_details-deploy_complexity}

\begin{table}[t]
\scriptsize
\begin{center}
\caption{
\textbf{Deployment complexities.}
We deploy (safe) AL for $t=1, ..., T$, and provide the complexities for each $t$.
Details are given in~\cref{appendix-experiment_details-deploy_complexity}. 
}
\label{tableS-deploy_complexities}
\begin{tabular}{l|l}
\toprule
method & time complexity \\
\midrule
our AAL~\crefp{algS-policy_al} & \multirow{3}{*}{$\mathcal{O}\left( (N_{\text{init}}+t-1)^2 \right)$} \\
our ASAL~\crefp{alg-policy_sal} & \\
DAD~\crefp{algS-policy_al} & \\
\hline
ALINE~\citep{huang2025aline} & \multirow{3}{*}{$\mathcal{O}\left( N_{\text{pool}} (N_{\text{init}}+t-1)^2 \right)$} \\
PFN AL, TabPFN AL~\crefp{algS-convention_al} & \\
Safe PFN AL, Safe TabPFN AL~\crefp{algS-convention_sal} & \\
\hline
GP AL, AGP AL~\crefp{algS-convention_al} & \multirow{3}{*}{$\mathcal{O}\left( (N_{\text{init}}+t-1)^3 \right) + \mathcal{O}\left( N_{\text{pool}}(N_{\text{init}}+t-1)^2 \right)$} \\
Safe GP AL, Safe AGP AL~\crefp{algS-convention_sal} & \\
Safe Random~\crefp{algS-convention_sal} & \\
\hline
SVGP AL~\crefp{algS-convention_al} & \multirow{2}{*}{$\mathcal{O}\left( (N_{\text{init}}+t-1)N_{\text{iv}}^2 \right) + \mathcal{O}\left( N_{\text{pool}}N_{\text{iv}}^2 \right)$} \\
Safe SVGP AL~\crefp{algS-convention_sal} & \\
\hline
MGP AL & $\mathcal{O}\left( N_{\text{ensemble}}(N_{\text{init}}+t-1)^3 \right) + \mathcal{O}\left( N_{\text{pool}}N_{\text{ensemble}}(N_{\text{init}}+t-1)^2 \right)$ \\
\bottomrule
\end{tabular}
\end{center}
\end{table}

All of our (safe) AL deployments are run on the same personal computer without a GPU.

The (safe) AL deployment complexities are summarized in~\cref{tableS-deploy_complexities} and detailed below.

\begin{itemize}
    \item 
Our \textbf{AAL}, our \textbf{ASAL}, \textbf{DAD}~\crefp{alg-policy_sal,algS-policy_al}:
NN forward passes take $\mathcal{O}\left( (N_{\text{init}}+t-1)^2 \right)$ at each $t=1, ..., T$, dominated by the transformer attender layers (\cref{figureS-nn},~\cite{Vaswani2017_attention}).

    \item \textbf{ALINE}~\citep{huang2025aline}: this method uses NN forward passes for AL decisions and posterior modeling jointly. The complexity is $\mathcal{O}\left( N_{\text{pool}}(N_{\text{init}}+t-1)^2 \right)$ as the method uses a transformer to process the pool as well as the observed data.

    \item \textbf{PFN/TabPFN AL, Safe PFN/TabPFN AL}~\crefp{algS-convention_al,algS-convention_sal}: PFN and TabPFN use transformers to process the observed data and the target pool. The observed data are self attended while the target pool cross attends to the observed data. The complexity is $\mathcal{O}\left( (N_{\text{init}}+t-1)^2 \right)$ to produce the posterior estimates of each data point. A linear burden is necessary for the (constrained) acquisition optimization, resulting in the final complexity $\mathcal{O}\left( N_{\text{pool}}(N_{\text{init}}+t-1)^2 \right)$.

    \item
\textbf{(A)GP AL}, \textbf{Safe (A)GP AL}~\crefp{algS-convention_al,algS-convention_sal}:
\begin{enumerate}
    \item Model fitting: at each $t=1, ..., T$, GP modeling takes $\mathcal{O}\left( (N_{\text{init}}+t-1)^3 \right)$ in time, exact factor depends on GP training methods; note that AGP methods have much smaller factor because the GP hyperparameters are inferred and the gram matrix is computed only once.
    \item Acquisition optimization: at each $t=1, ..., T$, (constrained) acquisition optimization via exhaustive search takes $\mathcal{O}\left( N_{\text{pool}}(N_{\text{init}}+t-1)^2 \right)$ due to GP inferences, where $N_{\text{pool}}$ is the size of the search pool.
\end{enumerate}

    \item
\textbf{Safe Random}:
A vanilla GP is used to model the safety confidence, resulting in the same complexities as $\textbf{Safe (A)GP AL}$ (up to a constant scaling).

    \item
\textbf{SVGP AL}, \textbf{Safe SVGP AL}~\crefp{algS-convention_al,algS-convention_sal}:
\begin{enumerate}
    \item Model fitting: at each $t=1, ..., T$, GP modeling takes $\mathcal{O}\left( (N_{\text{init}}+t-1)N_{\text{iv}}^2 \right)$ in time, $N_{\text{iv}}$ is the number of inducing variables; the time of selecting the inducing variables via k-means is negligible; in comparison to vanilla GPs (GP AL, Safe GP AL), we observe more training iterations needed while the complexity reduction of each gram matrix decomposition is minor in our data set sizes, resulting in a worse empirical time performance.
    \item Acquisition optimization: at each $t=1, ..., T$, (constrained) acquisition optimization via exhaustive search takes $\mathcal{O}\left( N_{\text{pool}}N_{\text{iv}}^2 \right)$ due to GP inferences, where $N_{\text{pool}}$ is the size of the search pool.
\end{enumerate}

    \item
\textbf{MGP AL}~\crefp{algS-convention_al}:
\begin{enumerate}
    \item Model fitting: at each $t=1, ..., T$, GP modeling takes $\mathcal{O}\left( N_{\text{ensemble}}(N_{\text{init}}+t-1)^3 \right)$ in time, $N_{\text{ensemble}}$ is the number of GPs.
    \item Acquisition optimization: at each $t=1, ..., T$, (constrained) acquisition optimization via exhaustive search takes $\mathcal{O}\left( N_{\text{pool}}N_{\text{ensemble}}(N_{\text{init}}+t-1)^2 \right)$ due to GP inferences, where $N_{\text{pool}}$ is the size of the search pool.
\end{enumerate}

\end{itemize}

\subsection{Benchmark Problems}\label{appendix-experiment_details-datasets}

\begin{table}[t]
\scriptsize
\begin{center}
\caption{
\textbf{Benchmark dimension in the literature.}
}\label{tableS-benchmark_dimension}
\begin{tabular}{l|ll}
\toprule
& dimension $D$ & area of impact \\
\midrule
\citet{riis2022bayesian} & $D \leq 6$ & GP AL on benchmark problems\\
\hline
\citet{pu2025weighted} & $D \leq 3$, $C \leq 4$, & GP modeling, BO on hyperparameter tuning \\
& $C$: number of categorical variables & \\
\hline
DAD & $D \leq 2$ & Amortized BED \\
\citep{foster_dad_2021,ivanova_idad_2021} & & on scientific experiments \\
\hline
\citet{ZimmerNEURIPS2018_b197ffde} & time series of 2 free variables & Safe AL on engineering problems \\
\hline
Stage Opt~\citep{yanan_sui_stagewise_2018} & $D \leq 2$ & Safe BO on clinical experiments \\
\hline
Safe Opt~\citep{berkenkamp2020bayesian} & $D \leq 2$ & Safe BO on robotics \\
\hline
\citet{alessandro2022safe} & $D \leq 5$ & Safe exploration on benchmark problems\\
\bottomrule
\end{tabular}
\end{center}
\end{table}

We run \textit{AL} and \textit{safe AL} experiments over the following benchmark problems.
All problems are mapped to $\mathcal{X}=[0, 1]^D$ if they are not originally defined on such domain.

We select benchmark tasks in line with the GP, AL, and safe-AL literature, focusing on low- to moderate-dimensional regimes where these methods are proven most competitive (summarized in~\cref{tableS-benchmark_dimension}).
Importantly, our final Fluid System scales to 7D--exceeding the usual $D \leq 3 \text{ or } 4$ in safe AL/BO--indicating that our approach remains effective on conventionally challenging problems.

\paragraph{AL, continuous - sin function:}
This is a one dimension problem $x \in [0, 1]$,
\begin{align*}
	f(x)=\sin(20x).
\end{align*}
In the experiments, we sample Gaussian noise $\epsilon \sim \mathcal{N}\left(0, 0.1^2\right)$.

We randomly sample 50 test points to evaluate the modeling RMSE.

\paragraph{AL, continuous - branin function:}
This function is defined over $(x_1, x_2) \in [-5, 10] \times [0, 15]$,
\begin{align*}
f_{a, b, c, r, s, t}\left((x_1, x_2)\right) &= a(x_2 - bx_1^2 + cx_1 -r) + s(1-t) \cos(x_1) + s,
\end{align*}
where $(a, b, c, r, s, t)=(1, \frac{5.1}{4 \pi^2}, \frac{5}{\pi}, 6, 10, \frac{1}{8 \pi})$ are constants.
We sample noise free data points and use the samples to normalize our output
\begin{align*}
f_{a, b, c, r, s, t}\left((x_1, x_2)\right)_{\text{normalize}}
&=\frac{
	f_{a, b, c, r, s, t}\left((x_1, x_2)\right) - mean(f_{a, b, c, r, s, t})
}{
	std(f_{a, b, c, r, s, t})
}.
\end{align*}
$std$ is a standard deviation.
In the experiments, we sample Gaussian noise $\epsilon \sim \mathcal{N}\left(0, 0.1^2\right)$.

We randomly sample 200 test points to evaluate the modeling RMSE.

\paragraph{AL, pool - airline passenger dataset:}
This is a publically available time series dataset~\footnote{https://github.com/jbrownlee/Datasets/blob/master/airline-passengers.csv}.
Each data point has a date input (year and month) and a number of passengers as output.
We convert the input into real number as $year + (month-1)/12$, and then rescale the entire input space to $[0, 1]$ (the earliest date becomes $0$ while the latest becomes $1$).
The output data are again normalized to zero mean and unit variance.

This is a dataset of $144$ measurements.
Before the experiments, we randomly pick $50$ points as test data to evaluate RMSE, $N_{\text{init}}$ initial data, and the remaining forms the pool.

\paragraph{AL, pool - Langley Glide-Back Booster (LGBB) dataset:}
This is a two dimension dataset described in~\citet{LGBB}\footnote{https://bobby.gramacy.com/surrogates/lgbb.tar.gz, lgbb\_original.txt}.
The dataset has multiple outputs and we take the "lift" to run our experiments (after normalized to zero mean and unit variance).
The inputs are $x_1$ (mach) and $x_2$ (alpha),
 which are normalized by
\begin{align*}
	x_1 &= mach / 6,\\
	x_2 &= (alpha + 5) / 35.
\end{align*}
After doing this, the input space is $[0, 1]^2$.

This is a dataset of around $850$ measurements.
Before the experiments, we randomly pick $200$ points as test data to evaluate RMSE, $N_{\text{init}}$ initial data, and the remaining forms the pool.

\paragraph{AL, pool - Airfoil dataset:}
This is a five dimension NASA dataset available on UCI datasets\footnote{https://archive.ics.uci.edu/dataset/291/airfoil+self+noise}.
The first five channels are taken as inputs, after mapped to $[0, 1]^5$.
The last channel is an output which needs to be normalized to zero mean and unit variance.
This is a dataset of around $1500$ measurements.
Before the experiments, we randomly pick $500$ points as test data to evaluate RMSE, $N_{\text{init}}$ initial data, and the remaining forms the pool.

\paragraph{Safe AL, continuous - Simionescu function:}
This is a constrained problem~\citep{Simionescu_function} defined over $(x_1, x_2) \in [-1.25, 1.25]^2$.
The main function is
\begin{align*}
	f(x_1, x_2)=0.1 x_1 x_2
\end{align*}
We sample noise free data points and use the samples to normalize our output
\begin{align*}
f\left((x_1, x_2)\right)_{\text{normalize}}
&=\frac{
	f\left((x_1, x_2)\right) - mean(f)
}{
	std(f)
}.
\end{align*}
$std$ is a standard deviation.
In the experiments, we sample Gaussian noise $\epsilon \sim \mathcal{N}\left(0, 0.1^2\right)$.

The task is subject to a constraint function:
\begin{align*}
q(x_1, x_2)&= \left[ 1 + 0.2 \cos\left( 8 \arctan(x_1/x_2) \right) \right]^2 - x_1^2 - x_2^2,\\
q\left((x_1, x_2)\right)_{\text{normalize}}
&=\frac{
	q\left((x_1, x_2)\right)
}{
	std(q)
}.
\end{align*}
$std$ is a standard deviation.
The constraint is $q\geq 0$, and we normalize only the standard deviation to ensure the constraint level remains the same.
In the experiments, we sample Gaussian noise $\epsilon_q \sim \mathcal{N}\left(0, 0.1^2\right)$.

We randomly sample 200 safe test points to evaluate the modeling RMSE.
The initial data points of each deployment are sampled at a central area $\mathcal{C} = [0.4, 0.6]^2 \subseteq \mathcal{X}$ (standardized $\mathcal{X}=[0, 1]^2$), under constraint $z \geq 0$.

\paragraph{Safe AL, continuous - Townsend function:}
This is a constrained problem~\citep{Townsend_function}~\footnote{https://www.chebfun.org/examples/opt/ConstrainedOptimization.html} defined over $(x_1, x_2) \in [-2.25, 2.25]\times[-2.5, 1.75]$.
The main function is
\begin{align*}
f(x_1, x_2)= - \left[ \cos((x_1-0.1)x_2) \right]^2 -x_1 \sin(3x_1 + x_2).
\end{align*}
We sample noise free data points and use the samples to normalize our output
\begin{align*}
f\left((x_1, x_2)\right)_{\text{normalize}}
&=\frac{
	f\left((x_1, x_2)\right) - mean(f)
}{
	std(f)
}.
\end{align*}
$std$ is a standard deviation.
In the experiments, we sample Gaussian noise $\epsilon \sim \mathcal{N}\left(0, 0.1^2\right)$.

The task is subject to a constraint function:
\begin{align*}
q(x_1, x_2)&=
\left(
2\cos(b)
- \frac{1}{2}\cos(2b)
- \frac{1}{4}\cos(3b)
- \frac{1}{8}\cos(4b)
\right)^2
+ \left( 2 \sin(b) \right)^2
- x_1^2 - x_2^2,\\
b &= \text{arctan2}(x1, x2)\\
&=\begin{cases}
\arctan( x_1 / x_2 ) &\text{, if } x_2 > 0\\
\arctan( x_1 / x_2 ) + \text{sign}(x_1) \pi &\text{, if } x_2 < 0, \text{ say }\text{sign}(0)=1\\
\text{sign}(x_1) \pi/2 &\text{, if } x_1 \neq 0, x_2 = 0\\
0 &\text{, if } x_1 = 0, x_2 = 0
\end{cases},\\
&q\left((x_1, x_2)\right)_{\text{normalize}}
=\frac{
	q\left((x_1, x_2)\right)
}{
	std(q)
}.
\end{align*}
$std$ is a standard deviation.
The constraint is $q\geq 0$, and we normalize only the standard deviation to ensure the constraint level remains the same.
In the experiments, we sample Gaussian noise $\epsilon_q \sim \mathcal{N}\left(0, 0.1^2\right)$.

We randomly sample 200 safe test points to evaluate the modeling RMSE.
The initial data points of each deployment are sampled at a central area $\mathcal{C} = [0.4, 0.6]^2 \subseteq \mathcal{X}$ (standardized $\mathcal{X}=[0, 1]^2$), under constraint $z \geq 0$.

\paragraph{Safe AL, pool - Langley Glide-Back Booster (LGBB) dataset:}
As described above for unconstrained AL, this dataset has multiple outputs and "lift" is taken as $y$.
We additionally take "pitch" as $z$ (pitching moment coefficient, see~\cite{lgbb_2004}).
We take a threshold $z \geq 0$, corresponding to around 60\% of the space.

The pitching moment coefficient is a quantity in aerodynamics, which is not necessarily safety critical but is important for the stability.
Collecting data under the constraint $z \geq 0$ means we model more carefully for larger and positive pitching moment coefficient.

This is a dataset of around $850$ measurements.
Before the experiments, we randomly pick $200$ safe points as test data to evaluate RMSE, $N_{\text{init}}$ initial data, and the remaining forms the pool.
The initial data points of each deployment are sampled at a central area $\mathcal{C} = [0.4, 0.6]^2 \subseteq \mathcal{X}$, under constraint $z \geq 0$.

\paragraph{Safe AL, pool - Engine dataset:}
This is a dataset published by Bosch\footnote{https://github.com/boschresearch/Bosch-Engine-Datasets/tree/master/pengines}.
We use the second file from the link (engine2).
We take engine\_speed, engine\_load, air\_fuel\_ratio as inputs, engine\_roughness\_s as output $y$, and temperature\_exhaust\_manifold as constraint $-z$.
The negative sign ($-z$) is added because we want a small temperature, and thus a large negative temperature $z$.
The inputs need to be mapped to $[0, 1]^3$, $y$ and $z$ are already normalized by Bosch.
We shift the constraint by $0.2$ so that $z - 0.2 \geq 0$ is around half of the space.

This is a dataset of around $800$ measurements.
Before the experiments, we randomly pick $200$ safe points as test data to evaluate RMSE, $N_{\text{init}}$ initial data, and the remaining forms the pool.
The initial data points of each deployment are sampled at a central area $\mathcal{C} = [0.4, 0.6]^3 \subseteq \mathcal{X}$, under constraint $z - 0.2 \geq 0$.

\paragraph{Safe AL, semi-continuous - high-pressure Fluid System:}
The high-pressure Fluid System is a nonlinear dynamical system, where we aim to model the rail pressure (our $y$)~\citep{ZimmerNEURIPS2018_b197ffde}.
The input has two free variables, actuation signal $v_k\in[0, 60]$ and speed of an external engine $n_k\in[1000, 4000]$, $k\in \mathbb{Z}^+$ is a time stamp.
Originally, this is a time series problem: $y_k$ is a function of $\bm{x}_k = (n_{k}, n_{k-1}, n_{k-2}, n_{k-3}, v_{k}, v_{k-1}, v_{k-3})$.

We sample $10^6$ random trajectories with a bounded step size (0.3 of the domain) and treat the trajectories as a 7D dataset. This renders the Fluid System into a pool-based 7D safe AL problem.
Note that our training does not incorporate time series nature of this task~\crefp{appendix-experiment_details-training_numerical}.
Sampling a pool with controlled maximum reachable step ensures the trajectories are realistic (e.g. as system's max acceleration is constrained); $10^6$ is dense enough yet not too expensive for baseline methods.

Concretely, each $\bm{x}_k$ is sampled by
\begin{align*}
    n_{k-3} &\sim \text{Uniform}[1000, 4000], \\
    n_{i} &\sim \text{Uniform}[\max(1000, n_{i-1}-\Delta n_{\text{max}}), \min(n_{i-1}+\Delta n_{\text{max}}, 4000)], \text{ for } i=k-2, k-1, k \\
    \Delta n_{\text{max}} &= 900 = 0.3*(4000-1000)\\
    v_{k-3} &\sim \text{Uniform}[0, 60], \\
    v_{k-1} &\sim \text{Uniform}[\max(0, v_{k-3}-2*\Delta v_{\text{max}}), \min(v_{k-3}+2*\Delta v_{\text{max}}, 60)], \\
    v_{k} &\sim \text{Uniform}[\max(0, v_{k-1}-\Delta v_{\text{max}}), \min(v_{k-1}+\Delta v_{\text{max}}, 60)],\\
    \Delta v_{\text{max}} &= 18 = 0.3*60.
\end{align*}

To obtain the data $y, z$, we take the implementation from~\citet{ZimmerNEURIPS2018_b197ffde} with minor adaptation:
\begin{enumerate}
    \item The noise-free observation is normalized $f_{\text{normalize}} = (f - 18) / 15$.
    \item The final observation is $y = f_{\text{normalize}} + \epsilon, \epsilon \sim \mathcal{N}\left(0, 0.1^2\right)$.
    \item The safety constraint $f \leq 18$ ($f_{\text{normalize}} \leq 0$) is treated as $q_{\text{normalize}} \geq 0$ where $q_{\text{normalize}} = - f_{\text{normalize}}$.
\end{enumerate}


The input domain $[1000, 4000]^4 \times [0, 60]^3$ is mapped to $\mathcal{X}=[0, 1]^7$.
We randomly sample 10000 safe test points to evaluate the modeling RMSE.
The initial data points are sampled at a central area $\mathcal{C} = [0.4, 0.6]^7 \subseteq \mathcal{X}$.

This 7D Fluid System exceeds the usual $D \leq 3 \text{ or } 4$ in safe AL/BO.
Conventional methods struggle as solving constrained acquisition optimization on such dimension is known to be difficult due to the naturally large pool and the subsequent high complexity (\citealt{alessandro2022safe,Kirschner2019lineBO}; see~\cref{appendix-experiment_details-deploy_complexity},~\cref{tableS-deploy_complexities} for the complexity).

\section{Ablation Studies}\label{appendix-ablation}

\subsection{Unconstrained AL Objectives, Train and Run Time Thresholds}
\label{appendix-ablation-amortization_threshold}

\begin{table}[t]
\scriptsize
\begin{center}
\caption{\textbf{Offline vs Online Time Thresholds}. The number of deployments required for the cumulative time savings of our trained policies to offset their initial pretraining time~\crefp{tableS-training_time}.
Thresholds are calculated relative to GP and MGP reference baselines~\crefp{figure-result_al,figure-result_sal}.
}
\label{tableS-trained_run_thresholds}
\begin{tabular}{rc|cccc}
\toprule
& $(D, N_{\text{init}}, T)$ & Our AAL & DAD & ALINE & Our ASAL \\
\midrule
\multirow{3}{*}{GP\_AL} & $(1, 1, 20)$ & $\sim$ 17x-18x & $\sim$ 21x-22x & $\sim$ 275x-342x &  \\
& $(2, 1, 30)$ & $\sim$ 10x-12x & $\sim$ 18x-21x & $\sim$ 281x-302x &  \\
& $(5, 20, 40)$ & $\sim$ 17x & $\sim$ 38x & $\sim$ 773x &  \\
\hline
\multirow{3}{*}{Safe\_GP\_AL} & $(2, 5, 30)$ &  &  &  & $\sim$ 12x-15x \\
& $(3, 5, 60)$ &  &  &  & $\sim$ 10x \\
& $(7, 10, 100)$ &  &  &  & $\sim$ 9x \\
\hline
\multirow{3}{*}{MGP\_AL} & $(1, 1, 20)$ & $\sim$ 1x & $\sim$ 1x & $\sim$ 10x-13x &  \\
& $(2, 1, 30)$ & $\sim$ 1x & $\sim$ 1x & $\sim$ 10x-11x &  \\
& $(5, 20, 40)$ & $\sim$ 1x & $\sim$ 1x & $\sim$ 28x &  \\
\hline
\multirow{3}{*}{Safe\_MGP\_AL} & $(2, 5, 30)$ &  &  &  & $\sim$ 1x \\
& $(3, 5, 60)$ &  &  &  & $\sim$ 1x \\
& $(7, 10, 100)$ &  &  &  & $\sim$ 1x \\
\bottomrule
\end{tabular}
\end{center}
\end{table}

We analyze the number of deployments required for our reusable trained policies to offset their initial pretraining time.
This metric can help inform decisions on whether training an amortized policy is computationally beneficial overall.

Before presenting these thresholds, we emphasize that our work caters primarily to scenarios with strictly constrained deployment times.
In such settings, deployment time is the primary bottleneck; if an experiment takes too long at runtime, it cannot be executed regardless of the pretraining cost.

However, if no such constraint is present, we can use the timing results~\crefp{tableS-training_time,figure-result_al,figure-result_sal} to calculate this threshold as follows:
\begin{align*}
    \text{NumDeployTrials} =
    \frac{\text{PolicyTrainingTime}}{5 \times \left(\text{BaselineDeployTime} - \text{PolicyDeployTime}\right)},
\end{align*}
the denominator includes a coefficient of $5$ to reflect that each task is deployed for 5 repetitions for statistical robustness;
additionally, for a fair comparison, the policy training times of the baselines are normalized to match our 10k training steps: DAD is divided by $2$ (from 20k steps) and ALINE is divided by $4$ (from 40k steps).

Because our policy deployment time is significantly shorter than that of conventional model-based baselines, running a baseline beyond this threshold results in a higher total time than offline training and subsequently deploying our policy.
Note that our AAL and ASAL methods generalize to various $T$, making the initial training effort more reusable than methods such as ALINE and DAD.

We compute these thresholds for GP AL, Safe GP AL, MGP AL, and Safe MGP AL under the same configurations as~\crefp{figure-result_al,figure-result_sal} (e.g., deployed $T$ settings).
GP is a standard baseline in the literature, while MGP serves as a strongly performing alternative, making both reasonable benchmarks for this comparison.

The results are detailed in~\cref{tableS-trained_run_thresholds}.
In summary, our methods (AAL and ASAL) offset their pretraining costs relatively quickly, requiring only a modest number of deployments. Notably, when compared to the computationally-intensive MGP baseline, our methods achieve amortization almost immediately. In contrast, ALINE requires hundreds of deployments (compared to baselines based on vanilla GPs) to justify its extensive training time.

\subsection{Unconstrained AL Objectives, Main vs Appendix}
\label{appendix-ablation-aal_objectives}

\begin{figure}[t]
\vskip 0.2in
\begin{center}
\textbf{1D appendix} $-\mathcal{I}_{\text{mean}}(\phi)$\\
\centerline{\includegraphics[width=\linewidth]{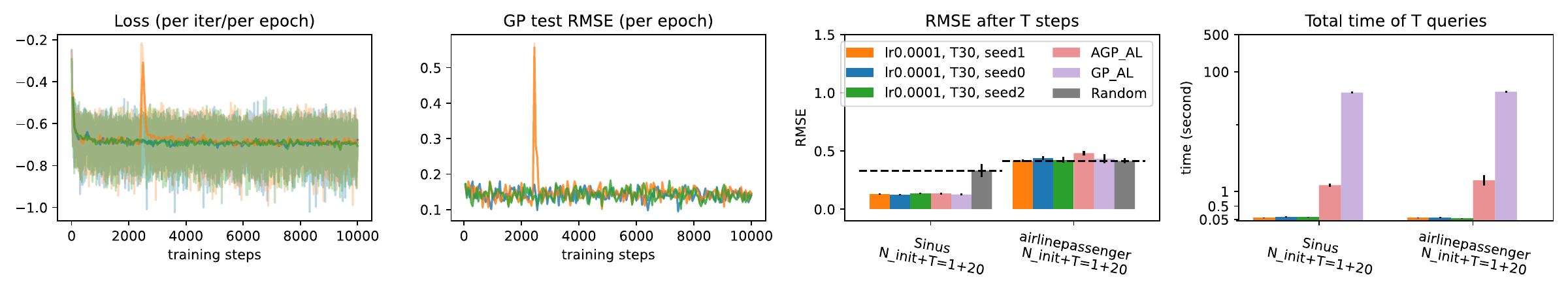}}
\textbf{1D main} $-\mathcal{I}(\phi)$\\
\centerline{\includegraphics[width=\linewidth]{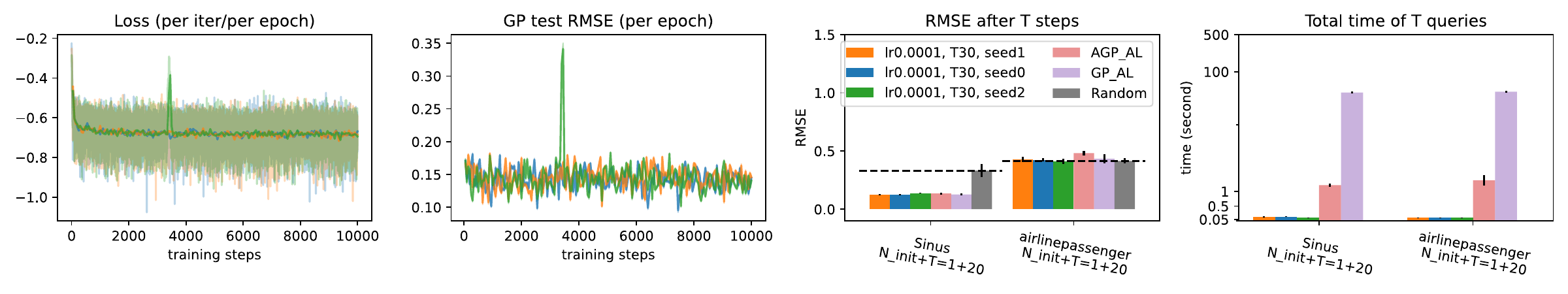}}
\textbf{2D appendix} $-\mathcal{I}_{\text{mean}}(\phi)$\\
\centerline{\includegraphics[width=\linewidth]{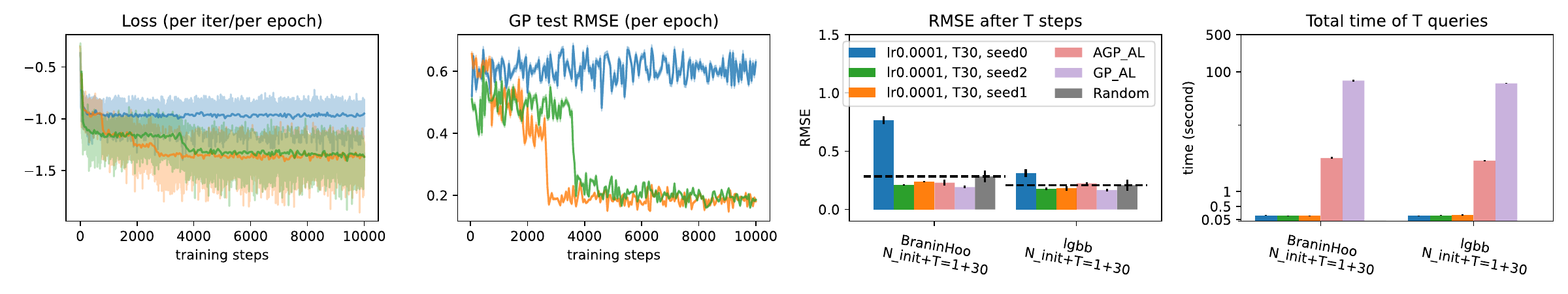}}
\textbf{2D main} $-\mathcal{I}(\phi)$\\
\centerline{\includegraphics[width=\linewidth]{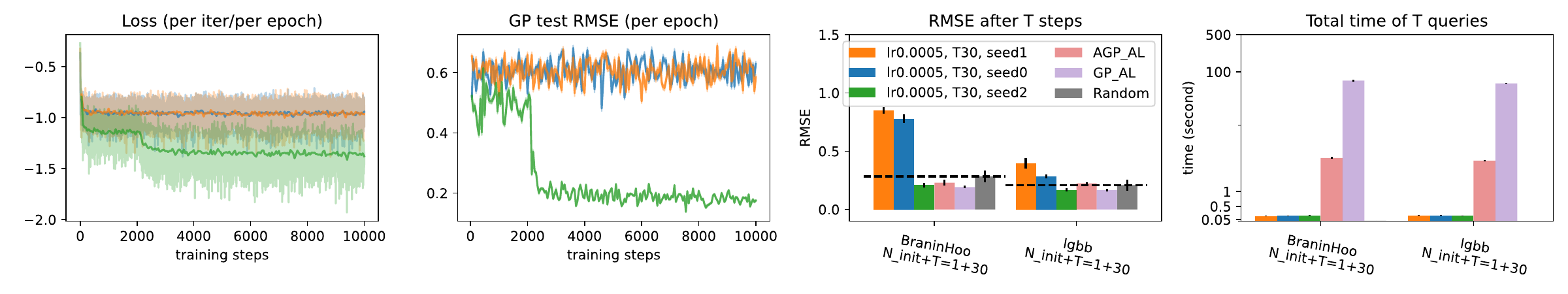}}
\captionof{figure}{
\textbf{Unconstrained AL results of objectives main $\mathcal{I}$~\crefp{eq-gp_logprob_reduction_objective} and appendix $\mathcal{I}_{\text{mean}}$~\crefp{eq-gp_entropy_reduction_objective}.}
The numerical congifurations (e.g. $D, T$) are identical for both objectives~\crefp{tableS-batch_sizes,tableS-training_samplers,tableS-training_complexities}.We deploy the policy with $T=20$ for $1D$ and $T=30$ for $2D$ problems, and further results on $5D$ problems are shown in~\cref{figureS-ablation_al2}.
The first column shows the training losses $-\mathcal{I}$ or $-\mathcal{I}_{\text{mean}}$ per training step and per epoch mean (mean of 50 steps).
At the end of each epoch, we sample a batch of GP functions, not for training, but we deploy the policy for $T$ steps (e.g. train $T_{\text{sim}}\leq T= 30$, then deploy $T=30$) and evaluate the GP RMSE against the ground truth.
The GP RMSEs per training epoch are shown in the second column.
After the training, we deploy each policy on benchmark problems and the results are the third and fourth columns (time illustrated in seconds).
}
\label{figureS-ablation_al}
\end{center}
\vskip -0.2in
\end{figure}

\begin{figure}[t]
\vskip 0.2in
\begin{center}
\textbf{5D appendix} $-\mathcal{I}_{\text{mean}}(\phi)$\\
\centerline{\includegraphics[width=\linewidth]{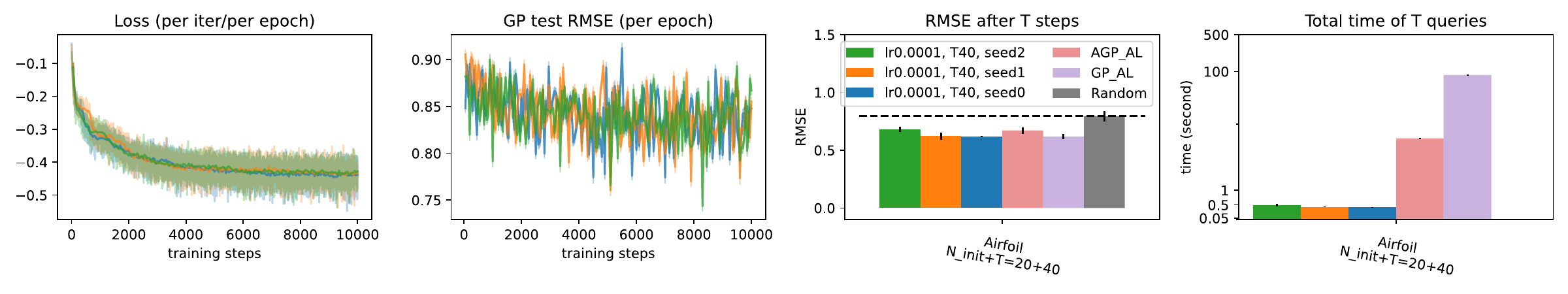}}
\textbf{5D main} $-\mathcal{I}(\phi)$\\
\centerline{\includegraphics[width=\linewidth]{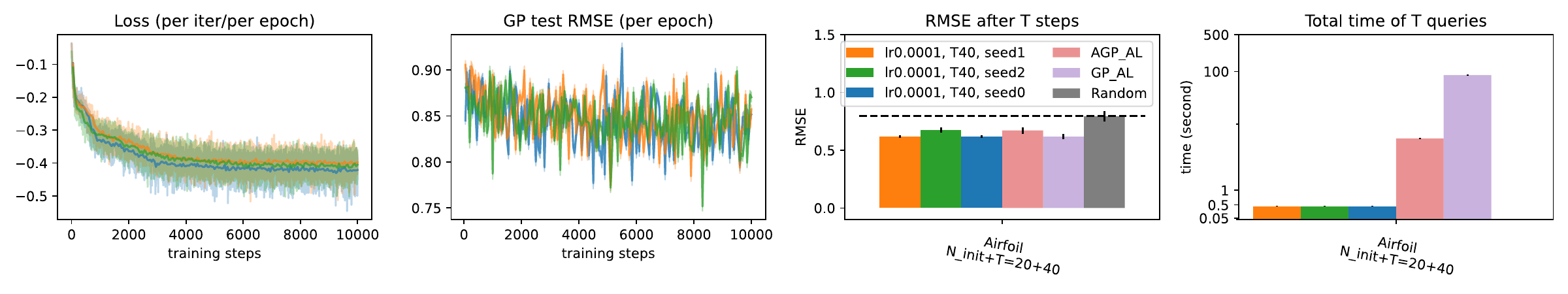}}
\captionof{figure}{
\textbf{Unconstrained 5D AL results of objectives main $\mathcal{I}$~\crefp{eq-gp_logprob_reduction_objective} and appendix $\mathcal{I}_{\text{mean}}$~\crefp{eq-gp_entropy_reduction_objective}.}
We deploy the policy with $T=40$ for $5D$ problems.
The columns are illustrate as~\cref{figureS-ablation_al}.
}
\label{figureS-ablation_al2}
\end{center}
\vskip -0.2in
\end{figure}

This experiment provide two pieces of information: (i) how the training is monitored, and (ii) comparison of the main objective and appendix objective.

We compare the two objectives $\mathcal{I}, \mathcal{I}_{\text{mean}}$~\crefp{eq-gp_logprob_reduction_objective,eq-gp_entropy_reduction_objective} under the same numerical setup as described in~\cref{tableS-batch_sizes,tableS-training_samplers}.
These objectives have the same complexity~\crefp{tableS-training_complexities}.

At the end of each training epoch, we sample GP functions, deploy the policy for $T$ steps, where this $T$ is the training max sequence length, and then we use the ground truth GP and the $T$ queries to evaluate the modeling RMSE.
We call this the GP test RMSE.
Please be aware of the difference that we deploy policies on benchmarks problems \textit{after} training, and such deployment RMSE requires GP fitting as ground truth is not available.

In~\cref{figureS-ablation_al}, we demonstrate the training losses (negative training objectives) as well as the deployment results.
The behavior of the two objectives do not seem to have obvious differences.
Note that this paper use RBF kernels, which means the functions are assumed stationary.
To avoid confusion, we perform only Type II maximum likelihood (no AGP) for the deployment GP modeling.

The training loss and GP test RMSE appear to be good indicators of the training performance, as the $1D$ and $2D$ examples show good deployments for policies of good training loss or GP test RMSE~\crefp{figureS-ablation_al}. 
The $2D$ policies sometimes get stuck in patterns where queries are put only at the border of $\mathcal{X}$, namely those with worse training losses and GP test RMSEs.
This pattern happens way more often if we train with the entropy losses $\mathcal{H}, \mathcal{H}_{\text{mean}}$ (results not shown).

The policies shown in our main paper are the one here with the best average training loss in the last 10 epochs (last 500 training steps).

\subsection{Unconstrained AL of Varied NN Sizes}
\label{appendix-ablation-nn_sizes}

\begin{table}[t]
\scriptsize
\begin{center}
\caption{\textbf{Policies of varied sizes, training metrics}. Shown is the average training loss (negative $\mathcal{I}$) and GP test RMSE (last 500 training steps).
Both metrics are meant to be minimized, giving first insight which trained policy to select.
}
\label{tableS-trained_varied_policy_size}
\begin{tabular}{l|ccc}
\toprule
& \multicolumn{3}{c}{loss / GP test RMSE}\\
& AAL ed 128 & AAL ed 64 & AAL ed 32\\
\midrule
1D & -0.6844 / 0.137 & -0.6841 / 0.141 & -0.6835 / 0.141 \\
2D & -1.3582 / 0.177 & -1.3292 / 0.201 & -1.2761 / 0.245 \\
5D & -0.4216 / 0.834 & -0.4044 / 0.845 & -0.3870 / 0.850 \\
\bottomrule
\end{tabular}
\end{center}
\end{table}

\begin{table}[t]
\scriptsize
\begin{center}
\caption{\textbf{Policies of varied sizes, model performance after deploying}. Shown is the RMSE on the test set. Smaller NNs (smaller ed) seem to result in noticeable performance deterioration.}
\label{tableS-result_al_varied_policy_size}
\begin{tabular}{c|cccccc}
\toprule
& AAL ed 128 & AAL ed 64 & AAL ed 32 & GP AL & Random\\
\midrule
Sinus
& $ 0.14 \pm 0.004 $ & $ 0.13 \pm 0.005 $ & $ 0.12 \pm 0.005 $ & $ 0.13 \pm 0.009 $ & $ 0.33 \pm 0.055 $\\
$N_{\text{init}}+T=1+20$ &&&&& \\
Airline
& $ 0.41 \pm 0.022 $ & $ 0.46 \pm 0.027 $ & $ 0.42 \pm 0.023 $ & $ 0.43 \pm 0.038 $ & $ 0.41 \pm 0.023 $\\
$N_{\text{init}}+T=1+20$ &&&&& \\
Branin
& $ 0.21 \pm 0.020 $ & $ 0.19 \pm 0.013 $ & $ 0.35 \pm 0.032 $ & $ 0.19 \pm 0.011 $ & $ 0.28 \pm 0.052 $\\
 $N_{\text{init}}+T=1+30$ &&&&& \\
LGBB
& $ 0.17 \pm 0.013 $ & $ 0.22 \pm 0.016 $ & $ 0.20 \pm 0.014 $ & $ 0.17 \pm 0.010 $ & $ 0.21 \pm 0.050 $\\
$N_{\text{init}}+T=1+30$ &&&&& \\
Airfoil
& $ 0.62 \pm 0.012 $ & $ 0.70 \pm 0.017 $ & $ 0.70 \pm 0.031 $ & $ 0.62 \pm 0.021 $ & $ 0.79 \pm 0.047 $ \\
 $N_{\text{init}}+T=20+40$ &&&&& \\
\bottomrule
\end{tabular}
\end{center}
\end{table}

As described in~\cref{appendix-experiment_details-training_numerical}, the embedding dimension plays a crucial role in the NN size.
Here, we would like to study whether we can make the NN even smaller.

We set the embedding dimension, denoted by ed, to 32 or 64, in comparison to our main 128, and we train on the unconstrained AL problems.
Similar to the main~\cref{table-result_al}, we report the test RMSE after AL deployment.
The results are shown in~\cref{tableS-result_al_varied_policy_size}.
It seems that ed 32 can be too small on 2D and 5D problems, ed 64 performs worse than ed 128 on 1D Airline, 2D LGBB, 5D Airfoil but comparable on 1D Sinus and 2D Branin problems.
This ablation study shows that our current setting, ed 128, seems to give an effective yet small NNs.

\subsection{Unconstrained AL of Smaller $T$}
\label{appendix-ablation-aal_varied_T}

\begin{table*}[t]
\scriptsize
\begin{center}
\caption{\textbf{RMSE on standard AL tasks of smaller $T$} ($T=10$ for Sinus, Airline and $T=20$ for Branin, LGBB). The results are attached to the main~\cref{table-result_al}, which demonstrates results of $T=20, 20, 30, 30$ for Sinus, Airline, Branin, LGBB, respectively. On Sin function, $T=10$ is usually too few for convergence.
}
\label{tableS-result_al_varied_T}
\begin{tabular}{l|cccc}
\toprule
Method & Sinus ($1+10$) & Airline ($1+10$) & Branin ($1+20$) & LGBB ($1+20$) \\
\midrule
Our AAL    & $0.64 \pm 0.056$ & $0.50 \pm 0.032$ & $0.33 \pm 0.008$ & $0.23 \pm 0.005$ \\
ALINE      & $0.56 \pm 0.022$ & $0.47 \pm 0.022$ & $0.37 \pm 0.024$ & $0.18 \pm 0.007$ \\
DAD        & $0.84 \pm 0.119$ & $0.45 \pm 0.014$ & $0.81 \pm 0.040$ & $0.32 \pm 0.012$ \\
PFN\_AL    & $1.16 \pm 0.037$ & $0.52 \pm 0.053$ & $0.43 \pm 0.036$ & $0.20 \pm 0.003$ \\
TabPFN\_AL & $1.19 \pm 0.026$ & $0.44 \pm 0.025$ & $0.34 \pm 0.028$ & $0.21 \pm 0.012$ \\
AGP\_AL    & $0.28 \pm 0.016$ & $0.53 \pm 0.028$ & $0.35 \pm 0.031$ & $0.23 \pm 0.010$ \\
GP\_AL     & $0.54 \pm 0.046$ & $0.46 \pm 0.033$ & $0.36 \pm 0.014$ & $0.18 \pm 0.014$ \\
SVGP\_AL   & $0.52 \pm 0.013$ & $0.50 \pm 0.060$ & $0.38 \pm 0.024$ & $0.20 \pm 0.013$ \\
MGP\_AL    & $0.22 \pm 0.040$ & $0.47 \pm 0.024$ & $0.30 \pm 0.023$ & $0.19 \pm 0.015$ \\
Random     & $0.51 \pm 0.070$ & $0.53 \pm 0.088$ & $0.45 \pm 0.068$ & $0.20 \pm 0.009$ \\
\bottomrule
\end{tabular}
\end{center}
\vspace{-10pt}
\end{table*}

We demonstrate in~\cref{tableS-result_al_varied_T} the learning performance on unconstrained AL tasks of smaller $T$, as attached to~\cref{table-result_al}.
The results show similar conclusion that our amortized approach (AAL) performs comparable learning outcomes.
ALINE and DAD require retraining for different $T$ and are not shown.

\subsection{Main Safe AL Objective, $\gamma=0.05$ vs $\gamma=0$}
\label{appendix-ablation-safe_AL_gamma}

\begin{figure}[t]
\vskip 0.2in
\begin{center}
\centerline{\includegraphics[width=\linewidth]{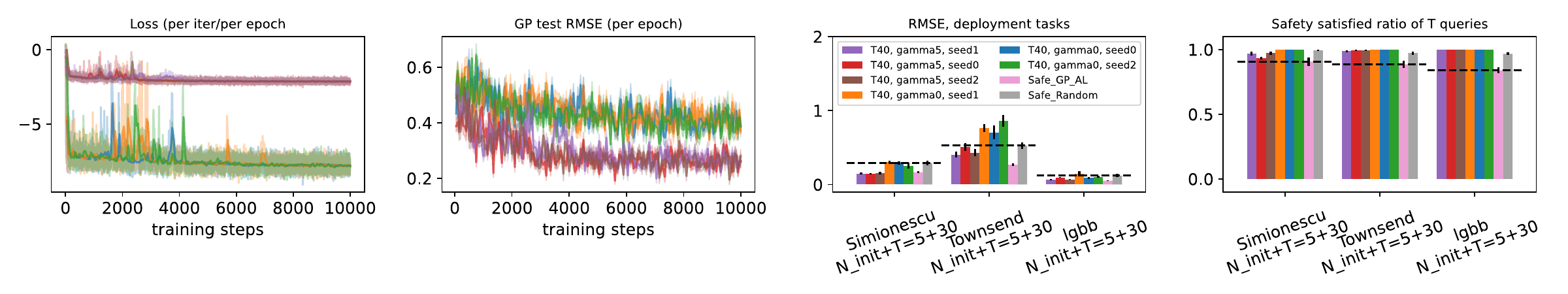}}
\captionof{figure}{
\textbf{The main safe AL objective $\mathcal{S}_{\mathcal{H}}$ with $\gamma=0.05$ vs $\gamma=0$, on $2D$ problems.}
The training loss values $-\mathcal{S}_{\mathcal{H}}$ are dominated by the minimized likelihood of unsafe queries if $\gamma=0$ (first column), and safety is thus prioritized more than the AL exploration in the training, as indicated by the RMSEs against ground truth GPs (second column).
After the training, the policies are deployed on three problems, and the results show similar exploration and safety trade-off as observed during the training (third and fourth columns).
The deployment colored bars are sorted according to the averaged training losses of the last 10 epochs (left to right: highest to lowest losses).
The baseline methods Safe GP AL and Safe Random are deployed with $\gamma=0.05$.
When $\gamma=0$, a conventional safe AL cannot operates because the entire input space will be identified unsafe.
}
\label{figureS-ablation_minunsafe_alpha}
\end{center}
\vskip -0.2in
\end{figure}

We train on $2D$ with $\mathcal{S}_{\mathcal{H}}, \gamma=0.05$ or $\gamma=0$~\crefp{eq-safe_gp_logprob_objective}.
Note that we add $10^{-5}$ to the unsafe likelihood for numerical stability, i.e. the safety term of $\mathcal{S}_{\mathcal{H}}$ has a bound $\log \text{max}( \gamma, p(z(\bm{x}_{\phi,t+1}) < 0 | z_{\phi, 1:t}, Z_{\text{init}}) ) \geq \log\left(10^{-5}\right)$.
To avoid confusion, we perform only Type II maximum likelihood for GP modeling.
We see from~\cref{figureS-ablation_minunsafe_alpha} that $\gamma=0$ makes the safety scores dominate the loss values.
As a result, the deployment is safer but the collected data lead to worse modeling performance.

Please be aware that our policy still does not need GP to deploy, which means we are faster than all the GP-based baselines, including Safe Random.
Overall, we should specify $\gamma$ depending on the safety criticality.

We train with a few different random seed values.
In our main paper, we again select the policy based on the average loss values of the last 10 epochs (last 500 training steps).

\subsection{ASAL Objectives Main vs Appendix, Safe AL of Minimum Unsafe Criterion~\cref{eq-acq_minunsafe}}\label{appendix-ablation-safe_al_criteria}

\begin{figure}[t]
\vskip 0.2in
\begin{center}
\textbf{appendix} $\mathcal{S}_{\mathcal{H},\text{division}}(\phi), \mathcal{S}_{\mathcal{H}_{\text{mean}},\text{division}}(\phi)$\\
\centerline{\includegraphics[width=\linewidth]{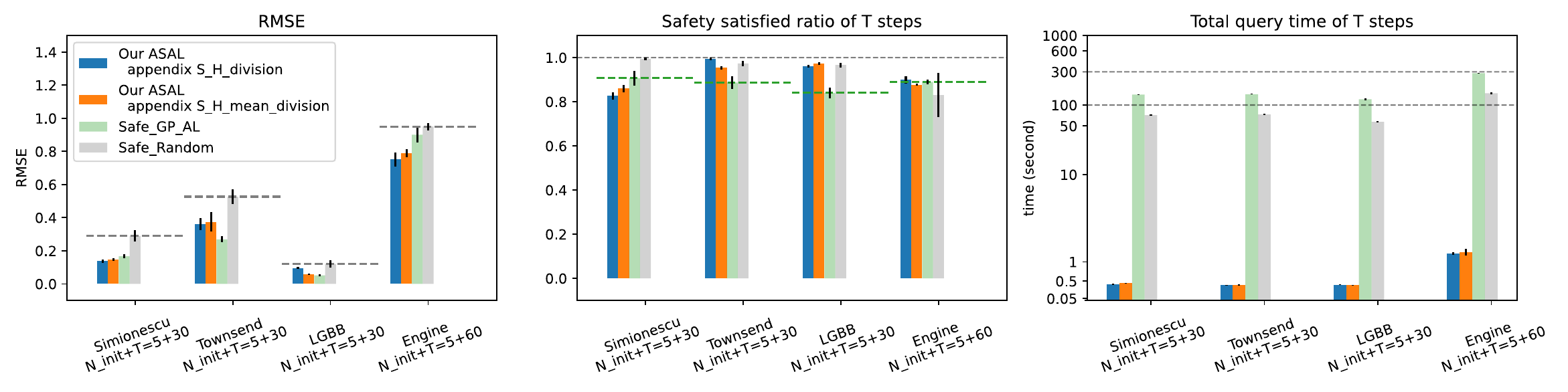}}
\textbf{main} $\mathcal{S}_{\mathcal{H}}(\phi)$, \textbf{appendix} $\mathcal{S}_{\mathcal{H}_{\text{mean}}}(\phi)$, \textbf{MinUnsafe GP AL}
\centerline{\includegraphics[width=\linewidth]{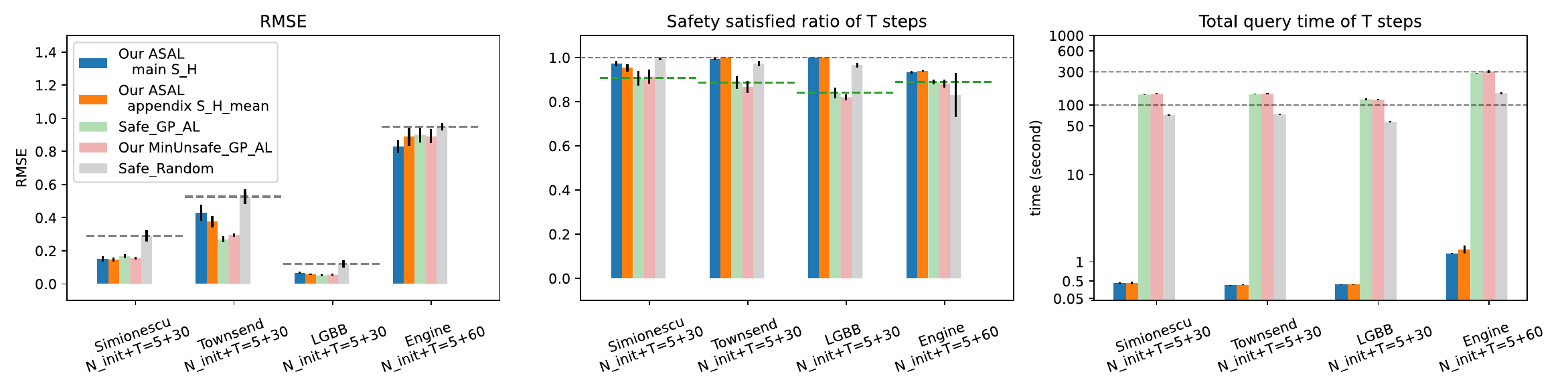}}
\captionof{figure}{\textbf{Results of different amortized safe AL objectives and the safe AL of MinUnsafe criterion.}
Our safe AL policies are trained on four objectives: main $\mathcal{S}_{\mathcal{H}}, \gamma=0.05$, appendix $\mathcal{S}_{\mathcal{H}_{\text{mean}}}, \gamma=0.05$, and appendix $\mathcal{S}_{\mathcal{H}, \text{division}}, \mathcal{S}_{\mathcal{H}_{\text{mean}}, \text{division}}$ which have no $\gamma$ clamping.
As ordered in~\cref{figureS-safe_acq}, $\mathcal{S}_{\mathcal{H}, \text{division}}, \mathcal{S}_{\mathcal{H}_{\text{mean}}, \text{division}}$ are on the top and $\mathcal{S}_{\mathcal{H}}, \mathcal{S}_{\mathcal{H}_{\text{mean}}}$ are at the bottom.
The MinUnsafe acquisition function corresponds to $\mathcal{S}_{\mathcal{H}}, \mathcal{S}_{\mathcal{H}_{\text{mean}}}$ objectives and is thus shown at the bottom.
}
\label{figureS-ablation_safe_al_losses}
\end{center}
\vskip -0.2in
\end{figure}

We compare a couple of methods:
\begin{enumerate}
    \item safe AL of policy trained on our main $\mathcal{S}_{\mathcal{H}}$, $\gamma=0.05$~\crefp{eq-safe_gp_logprob_objective},
    \item safe AL of policy trained on our appendix $\mathcal{S}_{\mathcal{H}_{\text{mean}}}$, $\gamma=0.05$ (unconstrained $\mathcal{H}_{\text{mean}}$ decorated with our main min unsafe likelihood, see~\cref{figureS-safe_acq}),
    \item safe AL of policy trained on our appendix $\mathcal{S}_{\mathcal{H}, \text{division}}$ (unconstrained $\mathcal{H}$ decorated with our appendix max safe likelihood, see~\cref{eq-safe_gp_logprob_division_objective,figureS-safe_acq}),
    \item safe AL of policy trained on our appendix $\mathcal{S}_{\mathcal{H}_{\text{mean}}, \text{division}}$ (unconstrained $\mathcal{H}_{\text{mean}}$ decorated with our appendix max safe likelihood, see~\cref{figureS-safe_acq}), and
    \item conventional GP based safe AL~\crefp{algS-convention_sal} but we add the unconstrained safety-aware acquisition criterion~\crefp{eq-acq_minunsafe}, named MinUnsafe GP AL: $
    \bm{x}_{t} = \text{argmax}
    \{\mathbb{H}[y(\bm{x}) | y_{1:t-1}, Y_{\text{init}}]
    - \log \text{max}(\gamma, p(z(\bm{x}) < 0 | z_{1:t-1}, Z_{\text{init}}) ) \}
    $ ($\gamma=0.05$, this is the same as~\cref{eq-acq_minunsafe} if we take expectation over the forecasted $y(\bm{x})$, and this corresponds to objectives $\mathcal{S}_{\mathcal{H}},\mathcal{S}_{\mathcal{H}_{\text{mean}}}$, see the paragraph of~\cref{eq-acq_minunsafe,eq-safe_gp_logprob_objective}).
\end{enumerate}

All objectives are leveraged under the same numerical setup as described in~\cref{tableS-batch_sizes,tableS-training_samplers}.
These objectives have the same complexity~\crefp{tableS-training_complexities}.

To avoid confusion, we perform only Type II maximum likelihood for GP modeling.
We demonstrate the result on all benchmark problems and datasets~\crefp{figureS-ablation_safe_al_losses}.
Firstly, the performance of Safe GP AL (constrained predictive entropy acquisition) does not seem to have obvious empirical differences from MinUnsafe GP AL.
This shows an advantage of MinUnsafe GP AL because its acquisition criterion is safety-aware but unconstrained, and an unconstrained optimization is much easier to be solved.
Note that, for a fair comparison, our paper solves the acquisition optimization of MinUnsafe GP AL on discretized set.

Our ASAL policies trained with different objectives all seem to query reasonable data.
In average, minimizing unsafe likelihood ($\mathcal{S}_{\mathcal{H}},\mathcal{S}_{\mathcal{H}_{\text{mean}}}$) prioritizes safety over AL exploration.
This is consistent to what we observe from the objective values~\crefp{figureS-safe_acq}.


\end{document}